\documentclass{article}

\usepackage{microtype}
\usepackage{graphicx}
\usepackage{subcaption}
\usepackage{booktabs} 

\usepackage{hyperref}
\usepackage{multirow}
\usepackage{array}
\usepackage{adjustbox}
\usepackage{graphicx}   

\usepackage[preprint]{icml2026}

\usepackage{subcaption}  
\usepackage{amsmath}
\usepackage{amssymb}
\usepackage{mathtools}
\usepackage{amsthm}

\usepackage[capitalize,noabbrev]{cleveref}

\theoremstyle{plain}

\theoremstyle{definition}

\theoremstyle{remark}

\usepackage{algorithm}
\usepackage{algorithmic}
\usepackage{amsmath} 
\usepackage{amssymb}
\usepackage{booktabs}
\usepackage{multirow}
\usepackage{xcolor}
\usepackage{svg}
\usepackage[textsize=tiny]{todonotes}
\usepackage[table]{xcolor}

\icmltitlerunning{Rethinking Input Domains in Physics-Informed Neural Networks via Geometric Compactification Mappings}

\begin{document}

\twocolumn[
  \icmltitle{Rethinking Input Domains in Physics-Informed Neural Networks\\ via Geometric Compactification Mappings}



  \icmlsetsymbol{equal}{*}

  \begin{icmlauthorlist}
    \icmlauthor{Zhenzhen Huang}{1}
    \icmlauthor{Haoyu Bian}{2}
    \icmlauthor{Jiaquan Zhang}{1}
    \icmlauthor{Yibei Liu}{1}
    \icmlauthor{Kuien Liu}{3}
    \icmlauthor{Caiyan Qin}{4}
    \icmlauthor{Guoqing Wang}{2}
    \icmlauthor{Yang Yang}{2}
    \icmlauthor{Chaoning Zhang}{2}
  \end{icmlauthorlist}

  \icmlaffiliation{1}{School of Information and Software Engineering, University of Electronic Science and Technology of China, Chengdu, China}
  \icmlaffiliation{2}{Computer Science and Engineering, University of Electronic Science and Technology of China, Chengdu, China}
  \icmlaffiliation{3}{Institute of Software Chinese Academy of Sciences, Beijing, China}
  \icmlaffiliation{4}{School of Robotics and Advanced Manufacture, Harbin Institute of Technology, Shenzhen, China}

  \icmlcorrespondingauthor{Chaoning Zhang}{chaoningzhang@uestc.edu.cn}

  \icmlkeywords{Machine Learning, Geometric Compactification Mappings, Physics-Informed Neural Networks}

  \vskip 0.3in
]



\printAffiliationsAndNotice{}  

\begin{abstract}
Several complex physical systems are governed by multi-scale partial differential equations (PDEs) that exhibit both smooth low-frequency components and localized high-frequency structures. Existing physics-informed neural network (PINN) methods typically train with fixed coordinate system inputs, where geometric misalignment with these structures induces gradient stiffness and ill-conditioning that hinder convergence. To address this issue, we introduce a mapping paradigm that reshapes the input coordinates through differentiable geometric compactification mappings and couples the geometric structure of PDEs with the spectral properties of residual operators. Based on this paradigm, we propose Geometric Compactification (GC)-PINN, a framework that introduces three mapping strategies for periodic boundaries, far-field scale expansion, and localized singular structures in the input domain without modifying the underlying PINN architecture. Extensive empirical evaluation demonstrates that this approach yields more uniform residual distributions and higher solution accuracy on representative 1D and 2D PDEs, while improving training stability and convergence speed.
\end{abstract}

\section{Introduction}
\label{sec:1}
Real-world physical systems exhibit complex dynamical behaviors, and accurate modeling of such systems is essential for understanding causal mechanisms in natural phenomena \cite{wang2025physics}. To describe the continuous spatiotemporal evolution, partial differential equations (PDEs) provide a mathematical framework in fields such as fluid mechanics \cite{flandoli2023stochastic}, electromagnetism \cite{khan2022physics}, and materials science \cite{zhang2022analyses}. As research moves to realistic settings, the resulting solutions often exhibit rich multiscale structures. In particular, smooth large-scale variations frequently coexist with localized, rapidly varying features, posing challenges for numerical modeling and learning-based solvers. Such multiscale behaviors arise in applications including vortex-dominated fluid flows \cite{ali2025machine} and wave propagation in heterogeneous media \cite{piao2024domain}.

\begin{figure}[t]
  \centering
  \includegraphics[width=\linewidth]{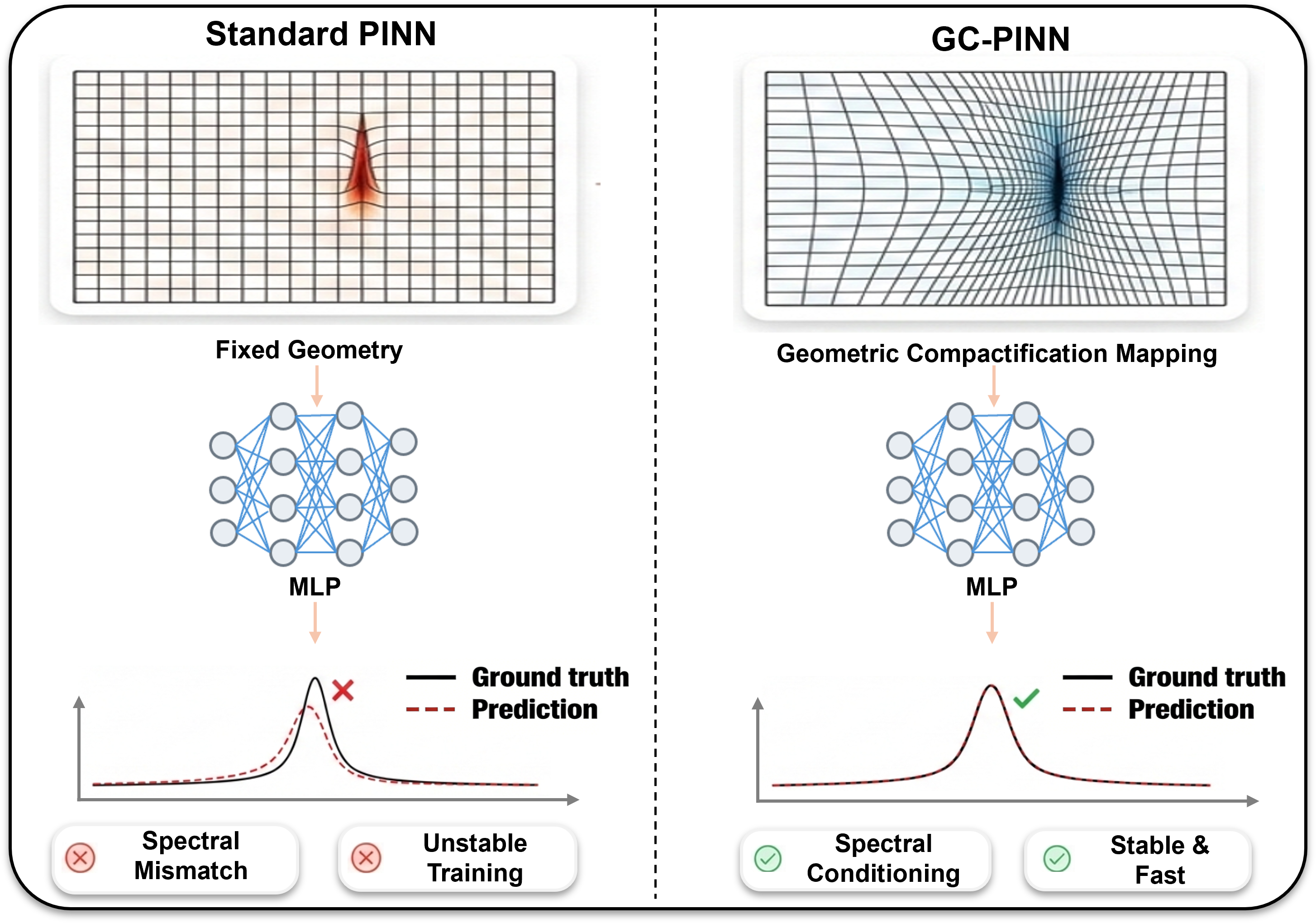}
  \caption{Conceptual comparison of standard PINNs and the proposed GC-PINN. Existing PINNs operate on the original Euclidean geometry, leading to unstable spectral conditioning and training. GC-PINN introduces a geometric compactification mapping paradigm that transforms the input domain, resulting in stable spectral properties and accelerated convergence while preserving predictive accuracy.}
  \label{fig:first}
\end{figure}

Physics-informed neural networks (PINNs) incorporate PDEs into the learning process and are widely used to model complex physical systems \cite{luo2025physics}. In recent years, PINNs have been extended to practical problems exhibiting pronounced multiscale characteristics \cite{kim2024review}. One class of methods targets multiscale solutions that simultaneously contain smooth regions and localized sharp variations. Such problems result in the development of multi-region or multi-network PINN frameworks, including XPINN \cite{jagtap2020extended} and hp-VPINN \cite{kharazmi2021hp}, which enhance local solution accuracy via domain decomposition. Another line of research emphasizes multiscale behavior in the spectral domain, such as learnable Fourier features in FF-PINN \cite{zhang2024application} and sinusoidal activation functions in SIREN \cite{sitzmann2020implicit}, to improve representation of high-frequency structures. Additional studies consider dynamical systems with pronounced temporal scale separation and adopt staged training or hierarchical modeling to capture interactions between slow and fast variables, as in T-PINN \cite{manikkan2023transfer}. Related approaches include adaptive sampling strategies \cite{torres2025adaptive} and frequency-weighted loss formulations, such as SA-PINN \cite{mcclenny2023self}, designed to improve training efficiency for multiscale PDEs.


However, most existing PINN variants optimize in input spaces fixed in Euclidean coordinate systems. Multiscale effects are typically addressed through posterior mechanisms such as network architecture, loss formulation, or sampling strategies, which allow neural networks to represent multiscale physical structures \cite{lawal2022physics}. Alignment between input geometry and the spectral properties of residual operators is generally not considered. As a result, optimization occurs in a coordinate system that does not reflect the geometric structure of the physical solution. This misalignment biases the residual spectrum, allowing high-frequency components to dominate backpropagation and significantly deteriorate the optimization condition number. As a result, the loss landscape becomes highly anisotropic across parameter directions \cite{rathore2024challenges}, which may lead to gradient stiffness, residual oscillations, and degraded training stability and convergence.

To overcome this limitation, this work reexamines PINN training stability from a geometric perspective and introduces a complementary mapping paradigm based on geometric compactification. The paradigm operates directly on the input domain and provides a lightweight, model-agnostic mechanism for geometric regulation.  Specifically, differentiable geometric compactification mappings renormalize local scales and topological structures of input coordinates in the physical domain. These mappings are equivalent to an effective modulation of the spectral condition numbers of residual operators, with their effects characterized by local variations in mapping derivatives. A set of mappings with complementary inductive biases is constructed, including Torus Mapping, Radial Mapping, and Local Stretching, which address representative geometric challenges in PDEs such as periodic topological discontinuities, far-field scale compression, and localized multiscale structures. Based on this paradigm, we propose GC-PINN, which incorporates the mappings into the loss formulation, redistributes highly concentrated residual energy in the spectral domain, and alleviates gradient stiffness to achieve spectral stability. Experimental results on representative multiscale PDE benchmarks show that the method improves convergence speed, enhances residual consistency, and yields more stable and accurate solutions in scenarios involving shocks, boundary layers, and singular structures.

    
    

Our contributions are summarized as follows:
\begin{itemize}
    \item \textbf{Proposing a geometric compactification mapping paradigm.}  
    We introduce a geometric compactification paradigm, applying differentiable input-domain reparameterizations to mitigate condition number deterioration and training instability in multiscale PDEs.
    
    \item \textbf{Developing a PINN framework.}  
    Based on the paradigm, we construct GC-PINN by explicitly incorporating a family of mappings into the loss formulation to regulate representative geometric challenges in PDEs.
    
    \item \textbf{Theoretical and empirical analysis.}  
    We examine the effects of geometric mappings on residual operators and evaluate the method on multiscale PDEs with localized singular structures in terms of training stability and solution accuracy.
\end{itemize}

\section{Related Work}
\label{sec:2}
\subsection{Existing Developments in PINNs}
\label{sec:2.1}
PINNs adopt a mesh-free formulation that naturally accommodates multi-scale physical phenomena and preserves predictive accuracy in data-scarce regimes, motivating extensive efforts to improve training stability and expressive capacity \cite{sun2024physics,wang2024practical,gnanasambandam2023self}. Since their introduction in 2019 \cite{raissi2019physics}, PINNs have evolved from solving simple forward PDE problems to handling complex inverse problems across diverse domains. For instance, Kissas et al. employed PINNs to model cardiovascular flows and predict arterial blood pressure using MRI data \cite{kissas2020machine}, while Cao et al. applied PINNs to turbulent airflow modeling in enclosed spaces, achieving substantial improvements in prediction accuracy compared to classical methods \cite{cao2026physics}. To enhance expressiveness, researchers have developed specialized architectures such as Fourier feature embeddings for high-frequency problems \cite{wang2021eigenvector}, conservative PINNs (cPINNs) \cite{quita2023conservative} for discrete domains, and variational PINNs \cite{kharazmi2019variational} for handling complex boundary conditions.

Recent studies extend PINNs to multiscale analysis through several representative approaches. Beyond explicitly embedding high-frequency bases or applying domain decomposition strategies, NH-PINN \cite{leung2022nh} employs neural homogenization techniques to solve representative cell problems for complex heterogeneous media. PINNs are also combined with classical numerical solvers such as finite element methods to preserve geometric and mechanical accuracy \cite{wu2024pinn,liu2026fem}. Han and Lee propose a neural network for multiscale homogenization trained with derivative-free loss \cite{han2023neural} , while Hall et al. propose graph-informed neural networks combining deep learning with probabilistic graphical models \cite{hall2021ginns} . However, these mechanisms are applied mainly at the level of model architecture or training dynamics and still neglect the role of the input domain.

\subsection{Geometric Methods in Artificial Intelligence}
\label{sec:2.2}

Geometric methods are widely adopted in artificial intelligence to improve model generalization and training stability through explicit incorporation of geometric priors \cite{rath2024boosting}. By transcending Euclidean assumptions, these methods demonstrate effectiveness in graph learning, generative modeling, and physical simulation \cite{gerken2023geometric}. In generative modeling, Ross et al. analyze memory behavior in GANs and VAEs from an optimization-geometric perspective, revealing the role of Hessian curvature in noise suppression and training stability \cite{ross2024geometric}. Lobashev et al. regulate latent-space curvature to enhance sampling and generalization in molecular generation \cite{lobashev2025hessian}. Beyond generative modeling, geometric deep learning systematically develop for representation learning. Gerken et al. review geometric deep learning and equivariant neural networks, focusing on group and gauge symmetries in constructing symmetry-aware models \cite{gerken2023geometric} . Papillon et al. survey topological deep learning and topological neural networks, highlighting the use of topological priors and invariants for representing complex structural relationships \cite{papillon2304architectures}. 
Recent advances further demonstrate that global structure awareness enhances reasoning coherence in large language models \cite{zhang2026ghs} and improves text summarization through holistic content organization \cite{zhang2026text}, suggesting the broader applicability of geometric and structural priors beyond physical systems. Geometric methods have also been introduced into PINNs. GAPINN constructs surrogate models for non-geometric boundaries using purely physics-driven, data-free training, enabling efficient physical simulation \cite{oldenburg2022geometry}. Costabal et al. propose a positional encoding based on Laplace–Beltrami eigenfunctions, providing neural networks with an input space that explicitly represents domain geometry \cite{costabal2024delta} . These studies demonstrate the potential of geometric approaches to enhance PINN modeling capability and stability.

\section{Preliminaries and Motivation}
\label{sec:3}
We analyze the intrinsic link between gradient stiffness in multiscale PDEs and the Euclidean geometry of the input space, which motivates the design of our method. Additional analysis is deferred to Appendix \ref{app:3.1} and \ref{app:4}.
\subsection{Gradient Stiffness in Multiscale PDEs}
\label{sec:3.1}

In a PINN framework, a neural network $u_\theta$ approximates the solution $u$ of a PDE defined on a spatiotemporal domain $\Omega \subset \mathbb{R}^d$ over time interval $t \in [0,T]$ by minimizing the residual-based loss
\begin{equation}
\mathcal{L}(\theta) = \frac{1}{N}\sum_{i=1}^N \|\mathcal{R}(\mathbf{x}_i, t_i; \theta)\|^2,
\quad \mathcal{R} = \mathcal{N}[u_\theta],
\end{equation}
where $\mathcal{N}$ denotes the governing differential operator.

For multiscale PDEs, optimization under a fixed Euclidean coordinate system gives rise to three intertwined challenges:

\begin{itemize}
    \item \textbf{Operator-scale amplification.} Differential operators amplify high-frequency components in regions with small characteristic scales, increasing residual magnitude and making the PDE constraint ill-conditioned.
    
    \item \textbf{Hessian condition number growth.} The Hessian spectrum becomes highly dispersed, causing gradient updates to be dominated by singular regions while suppressing contributions from smooth regions.
    
    \item \textbf{Imbalanced residual distribution.} Small singular regions contribute disproportionately to the total residual, biasing optimization toward localized structures and impeding learning of global behavior.
\end{itemize}

\subsection{Motivation for Geometric Compactification}
\label{sec:3.2}

In regions with characteristic length scale $\epsilon\ll1$, the residual magnitude of a $k$-th order differential operator $\mathcal{N}$ scales as
\begin{equation}
\|\mathcal{N}[u]\| = O(\epsilon^{-k}),
\end{equation}
arising from the natural amplification of high-frequency components in localized structures.

To counteract this effect, we introduce a geometric compactification mapping $\boldsymbol{\xi} = \phi(\mathbf{x}, t)$, where $\mathbf{x}\in\mathbb{R}^d$. Under the reparameterization with $\tilde{u}(\boldsymbol{\xi}) = u(\phi^{-1}(\boldsymbol{\xi}))$, the chain rule yields
\begin{equation}
\mathcal{N}[u](\mathbf{x}, t) = \tilde{\mathcal{N}}_\phi[\tilde{u}](\boldsymbol{\xi}),
\end{equation}
where $\tilde{\mathcal{N}}_\phi$ denotes the transformed operator with metric tensor induced by $\phi$. When $\phi$ adapts locally to scale $\epsilon$ with $\|\nabla\phi\| = O(\epsilon)$ and $\|\nabla^k\phi\|$ bounded for orders up to $k$, the amplification $\epsilon^{-k}$ is compensated by derivative scaling, yielding
\begin{equation}
\|\tilde{\mathcal{N}}_\phi[\tilde{u}]\| = O(1).
\end{equation}
This spectral normalization redistributes residual energy across scales, alleviating gradient stiffness and improving training stability.

\section{Method}
\label{sec:4}
In this section, we introduce the geometric compactification mapping paradigm and the resulting GC-PINN framework.

\begin{figure}[h]
  \centering
  \includegraphics[width=\linewidth]{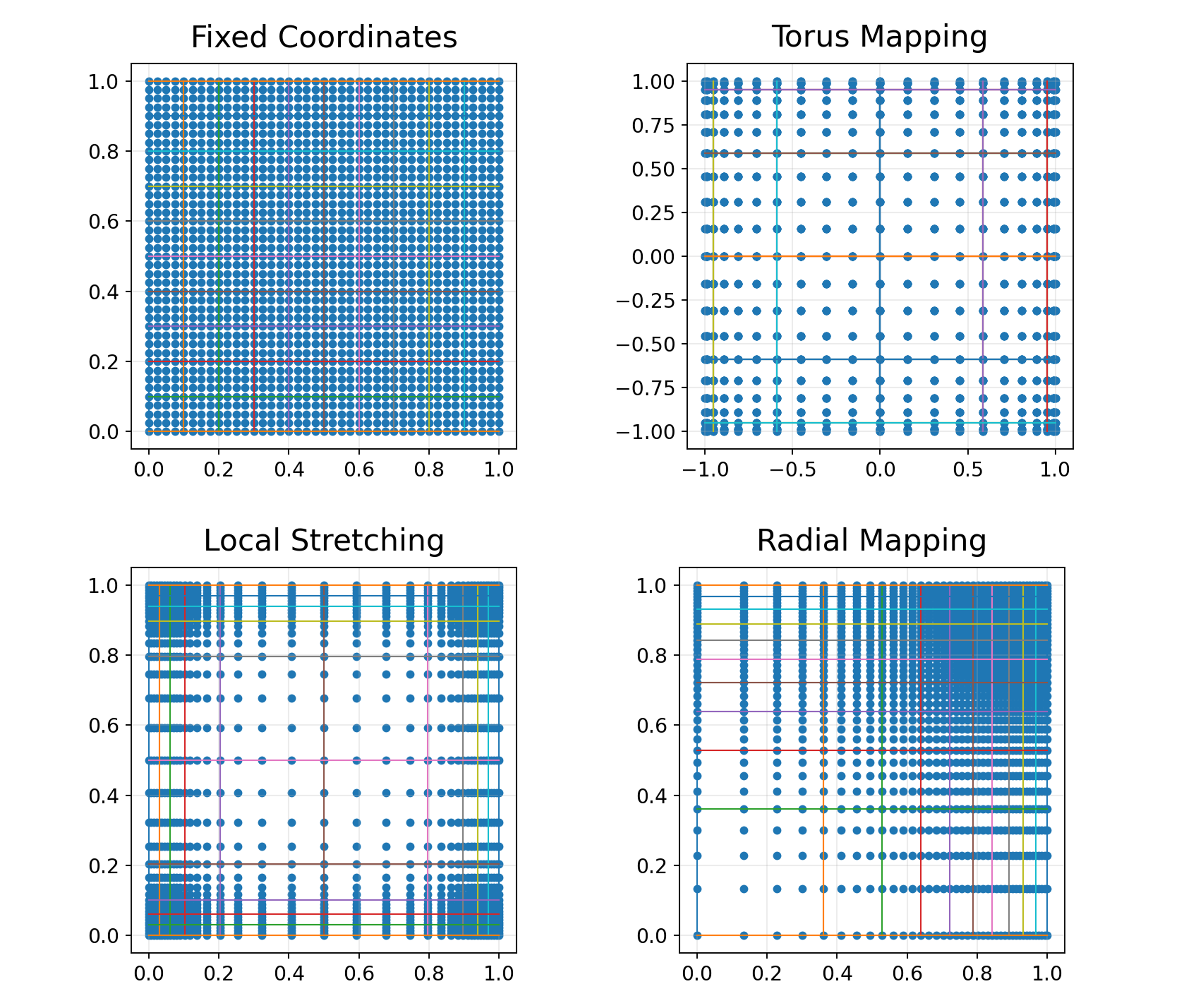}
  \caption{Coordinate distributions under the three mappings.}
  \label{fig:map}
\end{figure}



\begin{figure*}[h]
  \centering
  \includegraphics[width=\linewidth]{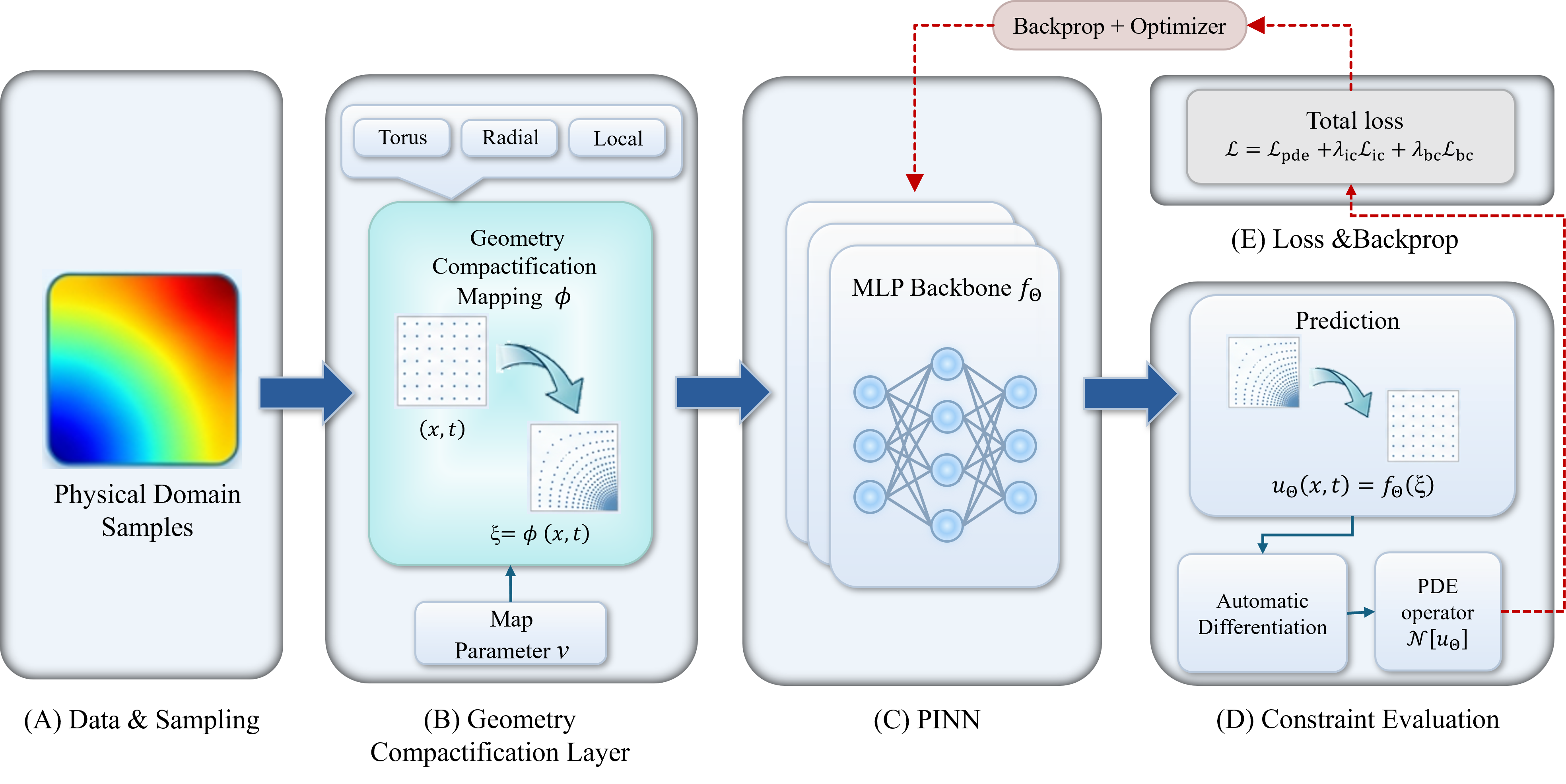}
  \caption{Overview of the GC-PINN pipeline. Physical domain coordinates are transformed via a learnable geometric compactification mapping and subsequently fed into an MLP backbone for solution prediction. The PDE residuals are computed through automatic differentiation, enabling end-to-end training of both the mapping parameters and the network weights.}
  \label{fig:main}
\end{figure*}

\subsection{Geometric Compactification Mappings}
\label{sec:4.1}
We regulate the input space of PINNs by introducing a differentiable mapping that compactifies the original physical domain into a reparameterized and bounded domain. Specifically, we define a geometric mapping
\begin{equation}
\phi \colon \Omega \subset \mathbb{R}^d\times\mathbb{R} \rightarrow \tilde{\Omega},
\qquad
(\mathbf{x},t)\mapsto \boldsymbol{\xi}=\phi(\mathbf{x},t),
\end{equation}
which replaces the original spatiotemporal coordinates $(\mathbf{x},t)$ with transformed coordinates $\boldsymbol{\xi}$ that reside in a bounded, controlled domain. Under this reparameterization, the solution is expressed as $\tilde{u}(\boldsymbol{\xi}, t)$, and spatial derivatives with respect to $\mathbf{x}$ are transformed according to the chain rule:
\begin{equation}
\begin{aligned}
\nabla_{\!\mathbf{x}} u &= \mathbf{J}_{\phi}^{\top}\nabla_{\!\boldsymbol{\xi}}\tilde{u}, \\
\nabla_{\!\mathbf{x}}^{2} u &= \mathbf{J}_{\phi}^{\top}\nabla_{\!\boldsymbol{\xi}}^{2}\tilde{u}\,\mathbf{J}_{\phi} + \nabla_{\!\boldsymbol{\xi}}\tilde{u}\cdot\nabla_{\!\mathbf{x}}^{2}\phi,
\end{aligned}
\end{equation}
where $\mathbf{J}_{\phi}=\bigl[\partial_{x_i}\phi_j\bigr]_{i,j=1}^{d}\in\mathbb{R}^{d\times d}$ is the Jacobian of $\phi$ and $\nabla_{\!\mathbf{x}}^{2}\phi$ denotes the Hessian tensor with entries $\partial_{x_i x_j}\phi_k$. Substituting these relations into the original PDE operator yields an equivalent operator defined on the compactified coordinates. This compactification explicitly bounds potentially large or unbounded physical coordinates, rescaling local derivatives and suppressing excessive contributions from far-field or high-frequency regions, which leads to more balanced training dynamics. Based on this formulation, we design three classes of mappings, each targeting geometric challenges commonly encountered in PDEs. The detailed analysis is provided in the Appendix~\ref{app:3.2}, \ref{app3.3} and \ref{app3.4}.

\textbf{Torus Mapping.} For PDEs with periodic boundary conditions or periodic dynamics, we introduce the mapping that embeds an interval $[x_{\min,i}, x_{\max,i}]$ into a toroidal topology, eliminating geometric discontinuities at periodic boundaries. For $d$-dimensional torus domains $\mathbb{T}^d$, we define
\begin{equation}
\phi_{\mathrm{tor}}(\mathbf{x},t) = \Biggl(\frac{1}{2\pi}\operatorname{atan2}\Bigl(\sin(2\pi \hat{\mathbf{x}}), \cos(2\pi \hat{\mathbf{x}})\Bigr),\; t \Biggr),
\end{equation}
where $\hat{\mathbf{x}} = (\hat{x}_1, \dots, \hat{x}_d)^\top$, and $\hat{x}_i$ denotes the normalized coordinate $i$. This mapping treats the torus domain as a compact topology, which helps suppress high-frequency modes generated by boundary discontinuities.

\textbf{Radial Mapping} is designed for problems with radial symmetry or far-field decay. Nonlinearly compresses far-field coordinates while preserving resolution near the center, thereby reducing extreme spatial variations in derivative scales. We employ a log-radial scaling transformation in the radial coordinate $r = \|\mathbf{x}\|$:
\begin{equation}
\label{eq:1}
\phi_{\mathrm{rad}}(\mathbf{x},t) = \left(\frac{\mathbf{x}}{r}\cdot\frac{\log(1+\alpha r)}{\log(1+\alpha)},\;t\right),
\end{equation}
where $\alpha>0$ controls the compression strength. This mapping compresses potentially large or unbounded coordinates into a finite domain, reducing the influence of far-field regions and enabling more balanced optimization.

\textbf{Local Stretching} is designed for PDEs with shocks or localized high-gradient structures. It adaptively stretches local coordinates to increase effective resolution near singular regions and alleviate gradient concentration caused by high-frequency features. We adopt a radial basis stretching defined as
\begin{equation}
\label{eq:2}
\phi_{\mathrm{loc}}(\mathbf{x},t) = \left((1-w(\mathbf{x}))\mathbf{x} + w(\mathbf{x})\tanh(\beta\mathbf{x}),\;t\right),
\end{equation}
where $w(\mathbf{x}) = \exp(-\beta|\mathbf{x}|^2)$, which improves the representation of sharp local features during training, and $\beta>0$ controls both the stretching width and the decay of $w(\mathbf{x})$. Examples of these mappings are provided in Figure \ref{fig:map}.

\subsection{GC-PINN Architecture}
\label{sec:4.2}
GC-PINN embeds a differentiable geometric compactification mapping at the input of a standard PINN:
\begin{equation}
\boldsymbol{\xi} = \boldsymbol{\phi}(\mathbf{x}, t), \qquad
u_\Theta(\mathbf{x}, t) = f_\Theta(\boldsymbol{\xi}),
\end{equation}
where $\mathbf{x} \in \mathbb{R}^d$ denotes spatial coordinates of arbitrary dimension and $f_\Theta$ is a conventional multilayer perceptron (MLP) \cite{rumelhart1986learning}. The training objective remains the classical PINN loss:
\begin{equation}
\begin{split}
\mathcal{L}_{\mathrm{PINN}} =
\mathbb{E}\|\mathcal{N}[u_\Theta]\|^2
+ \lambda_{\mathrm{IC}} \mathbb{E}\|u_\Theta(\cdot, t_0) - u_0\|^2\\
+ \lambda_{\mathrm{BC}} \mathbb{E}\|\mathcal{B}[u_\Theta]\|^2 .
\end{split}
\end{equation}
Through the chain rule, the Jacobian $\mathbf{J}_{\boldsymbol{\phi}}$ and higher-order Hessian tensors automatically rescale and redistribute derivatives of different orders, thereby alleviating spectral imbalance in multiscale residuals. The mapping parameters are optimized jointly with network weights, with regularization applied only to prevent degenerate mappings. The coordinate mapping is fully coupled with MLP, and all derivatives are calculated on the fly via automatic differentiation \cite{paszke2017automatic}. This paradigm performs geometric reparameterization exclusively at the input layer, leaving the network architecture, loss formulation, and training procedure unchanged. Consequently, GC-PINN integrates seamlessly into existing PINNs, providing a unified and lightweight solution for multiscale PDEs.

\section{Evaluations}
\label{sec:5}
In this section, we evaluate the performance of GC-PINN on benchmark PDEs that exhibit multiscale characteristics. We first describe the experimental setup and then provide a comprehensive assessment demonstrating the effectiveness of the proposed mapping paradigm. Due to space constraints, detailed experimental settings are provided in Appendix~\ref{app:2}, and additional results are provided in Appendix~\ref{app:5}.
\subsection{Experimental Setup}
\label{sec:5.1}

\textbf{Benchmarks.} We consider 1D Burgers equation, 1/2D convection–diffusion equation, 1D Helmholtz equation, and 2D Navier–Stokes equations. To explicitly test multiscale scenarios, high-frequency analytical terms are incorporated directly into the manufactured solutions, thereby introducing multiscale features. All problems use analytically manufactured solutions, where explicit expressions are substituted into the PDEs to compute the source terms, with detailed designs provided in Appendix~\ref{app:1}.

\textbf{Baseline Methods.} GC-PINN is compared with five baselines. The original PINN is implemented with a standard MLP architecture  \cite{raissi2019physics}. PINN+RAR implements the residual-based adaptive refinement method \cite{lu2021deepxde} proposed by \cite{wu2023comprehensive}. Fourier feature PINN, referred to as FF-PINN for convenience, embeds Fourier features into PINN following \cite{wang2021eigenvector}. SA-PINN \cite{mcclenny2023self} and gPINN \cite{yu2022gradient} enhance the baseline PINN through adaptive loss weighting and gradient-based augmentation, respectively. These baselines represent existing posterior optimization strategies for solving multiscale PDEs.

\begin{table*}[t]
\centering
\caption{Performance comparison across benchmark PDEs. 
(GC-* denote GC-PINN with Torus Mapping, Radial Mapping, and Local Stretching;  RAR, FF, SA denote PINN+RAR, FF-PINN, and SA-PINN. PDE names are also abbreviated.)}
\label{tab:results}
\small
\begin{tabular}{llcccccccc}
\toprule
PDE & Metric & GC-Torus & GC-Radial & GC-Local & gPINN & RAR & FF & SA & PINN \\
\midrule
\multirow{3}{*}{1D Burgers} 
& MSE & 3.25E-10 & \cellcolor{gray!25}\textbf{8.69E-11} & 1.69E-10 & 1.73E-10 & 1.96E-10 & 2.65E-09 & 5.13E-10 & 3.14E-09 \\
& $\text{Rel}_{L^2}$ & 2.44E-05 & \cellcolor{gray!25}\textbf{1.26E-05} & 1.76E-05 & 1.78E-05 & 1.90E-05 & 6.97E-05 & 3.07E-05 & 7.59E-05 \\
& $\text{Rel}_{H^1}$ & \cellcolor{gray!25}\textbf{3.97E-05} & 1.18E-04 & 9.15E-05 & 4.22E-05 & 6.16E-05 & 8.07E-05 & 8.51E-05 & 3.91E-04 \\
\midrule
\multirow{3}{*}{1D Conv-Diff} 
& MSE & 3.99E-06 & 6.00E-06 & \cellcolor{gray!25}\textbf{9.32E-08} & 1.96E-04 & 2.27E-06 & 1.39E-06 & 7.91E-07 & 2.43E-04 \\
& $\text{Rel}_{L^2}$ & 2.00E-03 & 2.45E-03 & \cellcolor{gray!25}\textbf{3.06E-04} & 1.40E-02 & 1.51E-03 & 1.18E-03 & 8.90E-04 & 1.56E-02 \\
& $\text{Rel}_{H^1}$ & 2.16E-03 & 3.55E-03 & 6.66E-04 & 1.57E-02 & 1.06E-03 & \cellcolor{gray!25}\textbf{5.15E-04} & 1.08E-03 & 1.57E-02 \\
\midrule
\multirow{3}{*}{1D Helmholtz} 
& MSE & \cellcolor{gray!25}\textbf{1.52E-08} & 1.42E-05 & 1.42E-05 & 2.01E-06 & 1.86E-05 & 5.05E-04 & 1.11E-06 & 5.17E-04 \\
& $\text{Rel}_{L^2}$ & \cellcolor{gray!25}\textbf{1.74E-04} & 5.33E-03 & 5.33E-03 & 2.01E-03 & 6.11E-03 & 3.18E-02 & 1.49E-03 & 3.22E-02 \\
& $\text{Rel}_{H^1}$ & \cellcolor{gray!25}\textbf{5.79E-05} & 1.87E-03 & 1.77E-03 & 6.71E-04 & 2.06E-03 & 1.05E-02 & 4.70E-04 & 1.07E-02 \\
\midrule
\multirow{3}{*}{2D Conv-Diff} 
& MSE & 7.19E-08 & 3.88E-07 & \cellcolor{gray!25}\textbf{3.26E-08} & 3.04E-07 & 6.50E-07 & 9.07E-07 & 1.40E-07 & 3.33E-05 \\
& $\text{Rel}_{L^2}$ & 5.33E-04 & 1.24E-03 & \cellcolor{gray!25}\textbf{3.59E-04} & 1.10E-03 & 1.60E-03 & 1.89E-03 & 7.44E-04 & 1.15E-02 \\
& $\text{Rel}_{H^1}$ & 1.09E-03 & 1.76E-03 & \cellcolor{gray!25}\textbf{5.17E-04} & 1.84E-03 & 2.28E-03 & 3.48E-03 & 1.42E-03 & 1.83E-02 \\
\midrule
\multirow{3}{*}{2D NS} 
& MSE & \cellcolor{gray!25}\textbf{6.24E-06} & 9.75E-05 & 2.01E-05 & 1.51E-04 & 2.97E-03 & 3.87E-04 & 2.52E-04 & 1.04E-01 \\
& $\text{Rel}_{L^2}$ & \cellcolor{gray!25}\textbf{2.71E-03} & 1.07E-02 & 4.87E-03 & 1.33E-02 & 5.92E-02 & 2.13E-02 & 1.72E-02 & 3.50E-01 \\
& $\text{Rel}_{H^1}$ & \cellcolor{gray!25}\textbf{1.19E-03} & 4.61E-03 & 1.85E-03 & 3.45E-03 & 1.88E-02 & 6.00E-03 & 4.39E-03 & 1.76E-01 \\
\bottomrule
\end{tabular}
\end{table*}

\textbf{Implementation Details.} All models share the same backbone: a four-layer MLP with 80 hidden units per layer and Tanh activations. Training follows a two-stage optimization protocol with a fixed random seed of 3407. In the first stage, models are pre-trained using Adam \cite{kingma2014adam} with a learning rate of $10^{-3}$ on 3,000 randomly sampled residual points per step for 6,000 epochs, while applying boundary condition losses with weight 100 and any necessary regularization terms. In the second stage, the optimization switches to L-BFGS \cite{wright1999numerical} with a learning rate of 1 using 15,000 residual points for 500 steps, with termination determined by strong Wolfe line search.

\textbf{Metrics.} During testing, the predicted solutions are evaluated against the analytical solutions at 2,000 uniformly sampled points in the domain $[0,1]$, and results are averaged over three random trials. We report the relative $L_2$ error defined as $\text{Rel}_{L_2} = \|\hat{u} - u\|_2 / \|u\|_2$, where the discrete $L_2$ norm is computed as the square root of the sum of squares over the test grid. The relative $H_1$ error is defined as $\text{Rel}_{H_1} = \frac{\sqrt{\|\hat{u} - u\|_2^2 + \|\partial_x \hat{u} - \partial_x u\|_2^2}}{\sqrt{\|u\|_2^2 + \|\partial_x u\|_2^2}}$ and the Mean Squared Error (MSE) is $(1/N) \sum_{i=1}^{N} (\hat{u}_i - u_i)^2$. The derivative $\partial_x \hat{u}$ is obtained via automatic differentiation, and all norms are computed on the discrete test grid. For all reported metrics, smaller values correspond to more accurate solutions.

\subsection{Main Results}
\label{sec:5.2}
Table~\ref{tab:results} reports quantitative results on five benchmark PDEs, including GC-PINN with three geometric mappings (GC-Torus, GC-Radial, and GC-Local). Overall, while FF-PINN achieves the best performance in a few isolated cases, GC-PINN attains the lowest error in most settings. Different mappings exhibit distinct advantages depending on the problem characteristics.

For the 1D Burgers equation, GC-Radial achieves the lowest mean squared error and $\text{Rel}_{L^2}$ error, while GC-Torus yields the best $\text{Rel}_{H^1}$ accuracy, indicating superior gradient fidelity. In the 1D convection--diffusion problem, GC-Local reduces both the mean squared error and $\text{Rel}_{L^2}$ error by approximately three orders of magnitude compared to the standard PINN. For the 1D Helmholtz equation, GC-Torus achieves the lowest errors across all reported metrics. Similar trends are observed in higher-dimensional settings. In particular, for the 2D Navier-Stokes equation, GC-Torus consistently outperforms other methods under all three metrics, reducing $\text{Rel}_{L^2}$ error from $3.50\times10^{-1}$ to $2.71\times10^{-3}$. These results demonstrate that geometric coordinate reparameterization improves PINN accuracy.

\subsection{Visualization Analysis}
\label{sec:5.3}

\begin{figure*}[h!]
  \centering
  \includegraphics[width=0.9\linewidth]{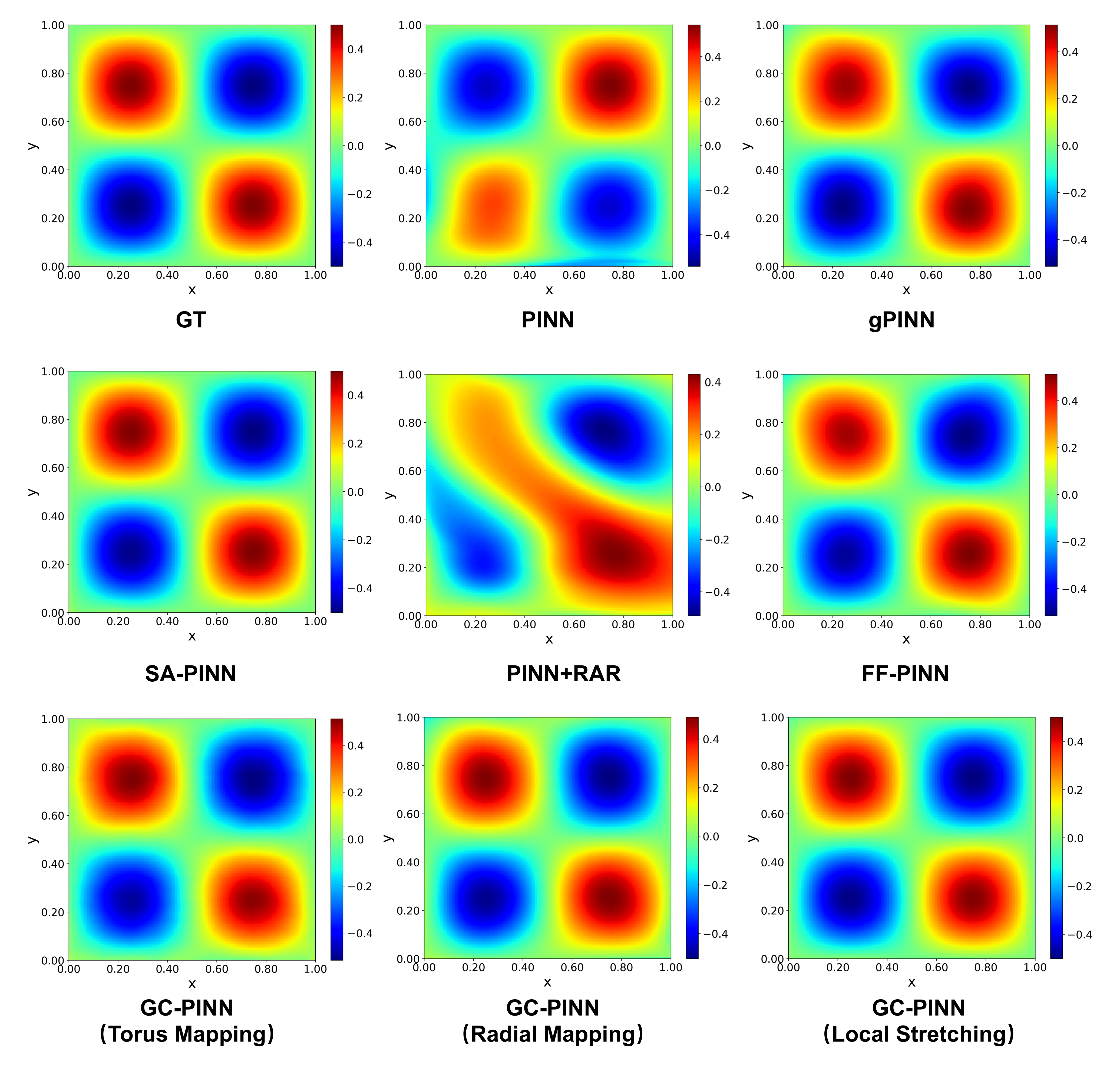}
  \caption{Comparison between predicted and ground-truth solutions for the 2D Navier-Stokes equation.}
  \label{fig:1}
\end{figure*}

\begin{figure*}[t]
  \centering
  \includegraphics[width=\linewidth]{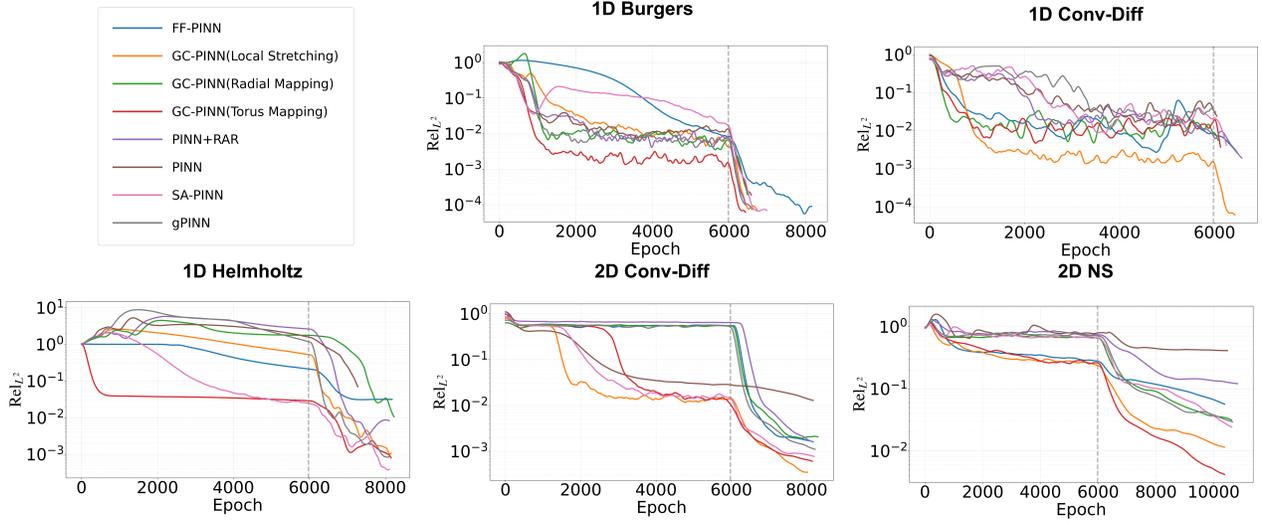}
  \caption{Training convergence on five PDE benchmark tasks, measured by the evolution of $\mathrm{Rel}_{L^2}$..}
  \label{fig:3}
\end{figure*}
\begin{figure}[h!]
  \centering
  \includegraphics[width=\linewidth]{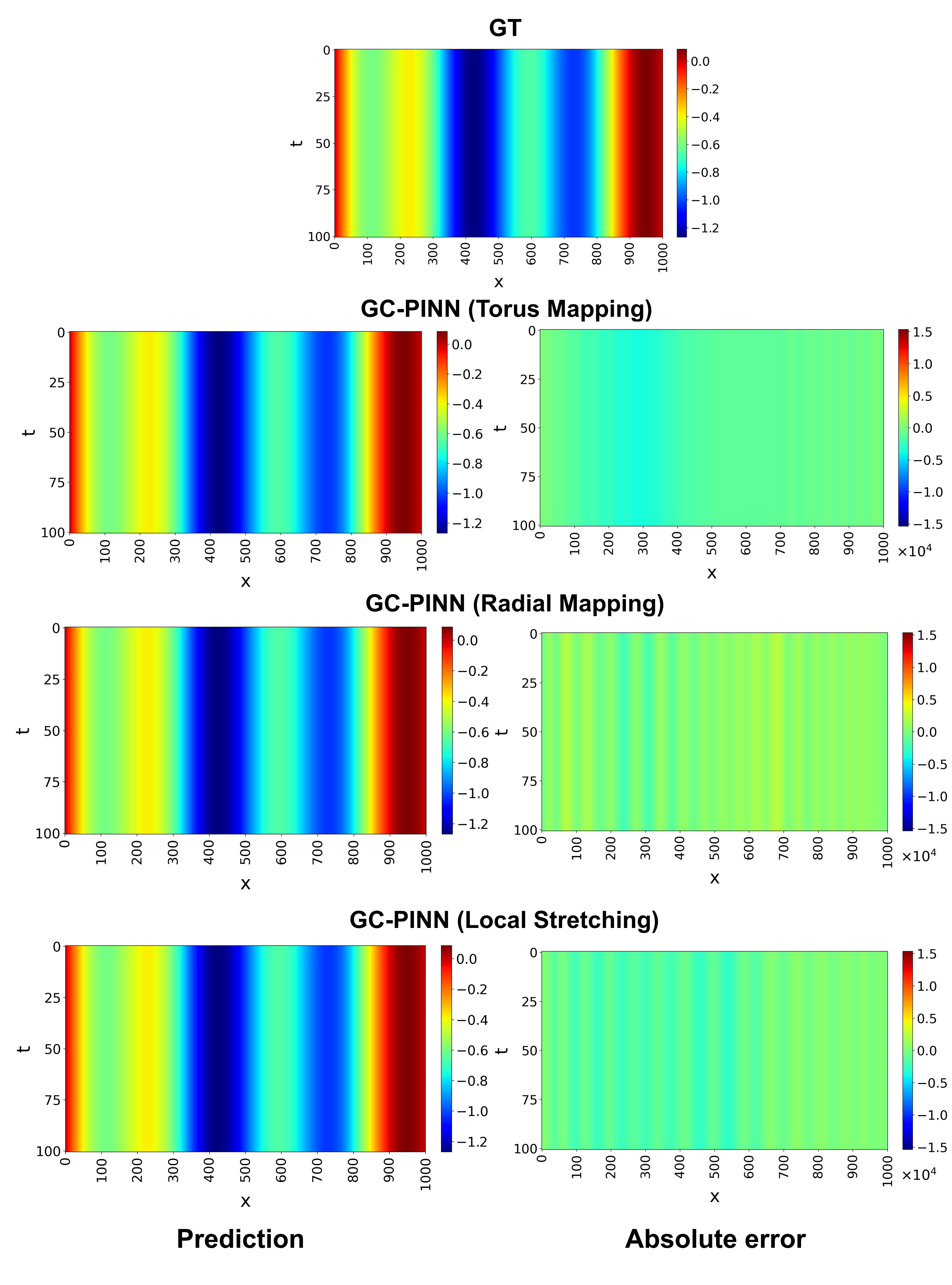}
  \caption{Visualization of the error distribution of predicted solutions for the 1D Burgers equation.}
  \label{fig:2}
\end{figure}

This section presents qualitative visualizations of GC-PINN in practical PDE solvers. Figure \ref{eq:1} reports comparisons between reference solutions and model predictions for the 2D Navier-Stokes equation. The reference solution exhibits a canonical periodic vortex pattern. Predictions produced by PINN and gPINN show noticeable distortions in the vicinity of vortex cores, together with blurred vortex boundaries. PINN+RAR and FF-PINN further exhibit phase shifts and amplitude attenuation, indicating an incomplete recovery of fine-scale flow structures. In contrast, all three GC-PINN variants closely match the reference solution, with vortex intensity and spatial organization accurately preserved. Another Figure \ref{fig:2} shows the spatiotemporal evolution and corresponding error distributions of GC-PINN with three mappings on the 1D Burgers equation. The reference solution forms a smooth traveling wave. All mappings capture the global structure consistently, while their error behaviors differ. Torus Mapping and Radial Mapping yield spatially uniform error distributions with low magnitude over the space-time domain. Local Stretching produces a localized error concentration at early time, reflecting the localized coordinate adaptation induced by the mapping.

\subsection{Parameter Sensitivity Analysis}
\label{sec:5.4}

\begin{table}[h]
\centering
\caption{Parameter sensitivity result for the 1D Burgers equation.}
\label{tab:burgers_sensitivity}
\small
\begin{tabular}{ccc|ccc}
\toprule
\multicolumn{3}{c}{Parameter $\beta$} & \multicolumn{3}{c}{Parameter $\alpha$} \\
\cmidrule(lr){1-3} \cmidrule(lr){4-6}
$\beta$ & MSE & Rel $L^2$ & $\alpha$ & MSE & Rel $L^2$ \\
\midrule
5  & 1.09E-09 & 4.47E-05 & 10 & 1.46E-09 & 5.19E-05 \\
10 & 3.58E-09 & 8.11E-05 & 20 & 8.17E-10 & 3.87E-05 \\
15 & 2.69E-10 & 2.22E-05 & 30 & 4.60E-10 & 2.91E-05 \\
20 & \textbf{1.69E-10} & \textbf{1.76E-05} & 40 & 3.44E-10 & 2.51E-05 \\
25 & 2.08E-10 & 1.95E-05 & 50 & \textbf{8.69E-11} & \textbf{1.26E-05} \\
30 & 3.72E-10 & 2.61E-05 & 60 & 1.28E-10 & 1.53E-05 \\
35 & 4.85E-09 & 9.43E-05 & 70 & 2.96E-10 & 2.33E-05 \\
\bottomrule
\end{tabular}
\end{table}

Table~\ref{tab:burgers_sensitivity} presents a comparison of the parameter sensitivity of GC-PINN with respect to the hyperparameters $\alpha$ and $\beta$ used in the Radial Mapping and Local Stretching, respectively. For Local Stretching, the stretching strength $\beta$ achieves an optimal balance around $\beta \approx 20$. When $\beta$ takes smaller values, the local coordinate stretching is insufficient to resolve boundary layer structures, leading to an increase in error by approximately one order of magnitude. When $\beta$ increases beyond this range, excessive compression introduces geometric distortion, which also degrades accuracy. For the Radial Mapping, the lowest error is attained at $\alpha \approx 50$. Smaller values of $\alpha$ result in insufficient compression of the far-field region, while larger values of $\alpha$ cause over-expansion of the near-field region, which disrupts the local smoothness of the solution. These results indicate that empirically chosen hyperparameters yield stable performance over a moderate range, with limited sensitivity to parameter variations.

\subsection{Convergence Analysis}
Figure~\ref{fig:3} shows the training convergence curves of different models on five benchmark problems, measured on a logarithmic scale of the relative \(L^2\) error. The dashed vertical line indicates the transition from Adam to L-BFGS optimization. For the 1D Burgers equation, GC-Torus reduces the error to the order of \(10^{-3}\) in the early Adam stage and remains stable, while PINN and FF-PINN converge more slowly and reach only the order of \(10^{-1}\) before the optimizer switch at 6000 iterations. For the 1D Helmholtz equation, GC-Torus reduces the error below \(10^{-2}\) within 500 iterations, whereas the errors of PINN and gPINN decrease gradually after approximately 4000 iterations, reflecting the limited adaptability of fixed coordinates to high-frequency problems. For the 2D cases, GC-PINN variants converge faster and achieve lower final errors on both the 2D convection-diffusion and 2D Navier-Stokes equations. In addition, GC-PINN variants exhibit smoother convergence trajectories with smaller fluctuations and limited rebound compared to the baseline methods. These results indicate that geometric compactification alleviates gradient stiffness during the Adam stage and provides improved initialization for L-BFGS optimization, leading to faster convergence across multi-scale PDEs.

\section{Conclusion}
\label{sec:6}
This work addresses the challenge of solving multiscale PDEs in complex physical systems using PINNs. A geometric compactification mapping paradigm is proposed, providing a new prior for PINN optimization. By introducing the mapping in the input domain, the local scales and topological structures of the coordinates are reorganized, forming the basis of the GC-PINN framework. Experiments demonstrate improvements in convergence speed and training stability, as well as more consistent and accurate solutions in multiscale scenarios. Future work will extend this approach to higher-dimensional and more complex geometries, and explore its application within neural operator frameworks.

\section{Impact Statement}  
This work aims to advance the application of PINNs to multiscale PDEs arising in realistic and complex physical systems. The proposed approach highlights the potential contribution of mathematical theory and geometric methods in neural networks, providing a plug-and-play interface and a new paradigm for multiscale problems in engineering and scientific computing. Naturally, this paradigm relies on the user’s mathematical expertise, which may lead to suboptimal mapping selection, the introduction of additional mappings or hyperparameters, increased tuning effort and memory consumption, and, in some cases, even optimization failure.

\bibliography{example_paper}

@book{flandoli2023stochastic,
  title={Stochastic partial differential equations in fluid mechanics},
  author={Flandoli, Franco and Luongo, Eliseo and others},
  volume={2330},
  year={2023},
  publisher={Springer}
}

@article{khan2022physics,
  title={Physics informed neural networks for electromagnetic analysis},
  author={Khan, Arbaaz and Lowther, David A},
  journal={IEEE Transactions on Magnetics},
  volume={58},
  number={9},
  pages={1--4},
  year={2022},
  publisher={IEEE}
}

@article{zhang2022analyses,
  title={Analyses of internal structures and defects in materials using physics-informed neural networks},
  author={Zhang, Enrui and Dao, Ming and Karniadakis, George Em and Suresh, Subra},
  journal={Science advances},
  volume={8},
  number={7},
  pages={eabk0644},
  year={2022},
  publisher={American Association for the Advancement of Science}
}

@article{luo2025physics,
  title={Physics-informed neural networks for PDE problems: a comprehensive review},
  author={Luo, Kuang and Zhao, Jingshang and Wang, Yingping and Li, Jiayao and Wen, Junjie and Liang, Jiong and Soekmadji, Henry and Liao, Shaolin},
  journal={Artificial Intelligence Review},
  volume={58},
  number={10},
  pages={323},
  year={2025},
  publisher={Springer}
}

@article{wang2025physics,
  title={Physics-guided deep learning for dynamical systems: A survey},
  author={Wang, Rui and Yu, Rose},
  journal={ACM Computing Surveys},
  volume={58},
  number={5},
  pages={1--31},
  year={2025},
  publisher={ACM New York, NY}
}

@article{kim2024review,
  title={A review of physics informed neural networks for multiscale analysis and inverse problems},
  author={Kim, Dongjin and Lee, Jaewook},
  journal={Multiscale Science and Engineering},
  volume={6},
  number={1},
  pages={1--11},
  year={2024},
  publisher={Springer}
}

@article{ali2025machine,
  title={Machine learning in fluid dynamics—Physics-informed neural networks (PINNs) using sparse data: A review},
  author={Ali, Mjalled and Miron, Philippe and M{\"o}nnigmann, Martin and Nikolay, Bukharin and others},
  journal={Fluids},
  volume={10},
  number={9},
  pages={226},
  year={2025},
  publisher={MDPI AG}
}

@article{piao2024domain,
  title={A domain-adaptive physics-informed neural network for inverse problems of Maxwell's equations in heterogeneous media},
  author={Piao, Shiyuan and Gu, Hong and Wang, Aina and Qin, Pan},
  journal={IEEE Antennas and Wireless Propagation Letters},
  volume={23},
  number={10},
  pages={2905--2909},
  year={2024},
  publisher={IEEE}
}

@article{jagtap2020extended,
  title={Extended physics-informed neural networks (XPINNs): A generalized space-time domain decomposition based deep learning framework for nonlinear partial differential equations},
  author={Jagtap, Ameya D and Karniadakis, George Em},
  journal={Communications in Computational Physics},
  volume={28},
  number={5},
  year={2020},
  publisher={Brown Univ., Providence, RI (United States)}
}

@article{kharazmi2021hp,
  title={hp-VPINNs: Variational physics-informed neural networks with domain decomposition},
  author={Kharazmi, Ehsan and Zhang, Zhongqiang and Karniadakis, George Em},
  journal={Computer Methods in Applied Mechanics and Engineering},
  volume={374},
  pages={113547},
  year={2021},
  publisher={Elsevier}
}

@article{sitzmann2020implicit,
  title={Implicit neural representations with periodic activation functions},
  author={Sitzmann, Vincent and Martel, Julien and Bergman, Alexander and Lindell, David and Wetzstein, Gordon},
  journal={Advances in neural information processing systems},
  volume={33},
  pages={7462--7473},
  year={2020}
}

@article{manikkan2023transfer,
  title={Transfer physics informed neural network: a new framework for distributed physics informed neural networks via parameter sharing},
  author={Manikkan, Sreehari and Srinivasan, Balaji},
  journal={Engineering with Computers},
  volume={39},
  number={4},
  pages={2961--2988},
  year={2023},
  publisher={Springer}
}

@article{mcclenny2023self,
  title={Self-adaptive physics-informed neural networks},
  author={McClenny, Levi D and Braga-Neto, Ulisses M},
  journal={Journal of Computational Physics},
  volume={474},
  pages={111722},
  year={2023},
  publisher={Elsevier}
}

@article{torres2025adaptive,
  title={Adaptive Physics-informed Neural Networks: A Survey},
  author={Torres, Edgar and Schiefer, Jonathan and Niepert, Mathias},
  journal={arXiv preprint arXiv:2503.18181},
  year={2025}
}

@article{lawal2022physics,
  title={Physics-informed neural network (PINN) evolution and beyond: A systematic literature review and bibliometric analysis},
  author={Lawal, Zaharaddeen Karami and Yassin, Hayati and Lai, Daphne Teck Ching and Che Idris, Azam},
  journal={Big Data and Cognitive Computing},
  volume={6},
  number={4},
  pages={140},
  year={2022},
  publisher={MDPI}
}

@article{rathore2024challenges,
  title={Challenges in training pinns: A loss landscape perspective},
  author={Rathore, Pratik and Lei, Weimu and Frangella, Zachary and Lu, Lu and Udell, Madeleine},
  journal={arXiv preprint arXiv:2402.01868},
  year={2024}
}

@article{sun2024physics,
  title={A physics-informed neural network framework for multi-physics coupling microfluidic problems},
  author={Sun, Runze and Jeong, Hyogu and Zhao, Jiachen and Gou, Yixing and Sauret, Emilie and Li, Zirui and Gu, Yuantong},
  journal={Computers \& Fluids},
  volume={284},
  pages={106421},
  year={2024},
  publisher={Elsevier}
}

@article{wang2024practical,
  title={A practical PINN framework for multi-scale problems with multi-magnitude loss terms},
  author={Wang, Yong and Yao, Yanzhong and Guo, Jiawei and Gao, Zhiming},
  journal={Journal of Computational Physics},
  volume={510},
  pages={113112},
  year={2024},
  publisher={Elsevier}
}

@article{gnanasambandam2023self,
  title={Self-scalable tanh (stan): Multi-scale solutions for physics-informed neural networks},
  author={Gnanasambandam, Raghav and Shen, Bo and Chung, Jihoon and Yue, Xubo and Kong, Zhenyu},
  journal={IEEE Transactions on Pattern Analysis and Machine Intelligence},
  volume={45},
  number={12},
  pages={15588--15603},
  year={2023},
  publisher={IEEE}
}

@article{leung2022nh,
  title={NH-PINN: Neural homogenization-based physics-informed neural network for multiscale problems},
  author={Leung, Wing Tat and Lin, Guang and Zhang, Zecheng},
  journal={Journal of Computational Physics},
  volume={470},
  pages={111539},
  year={2022},
  publisher={Elsevier}
}

@article{wu2024pinn,
  title={PINN enhanced extended multiscale finite element method for fast mechanical analysis of heterogeneous materials},
  author={Wu, Zhetong and Zhang, Hanbo and Ye, Hongfei and Zhang, Hongwu and Zheng, Yonggang and Guo, Xu},
  journal={Acta Mechanica},
  volume={235},
  number={7},
  pages={4895--4913},
  year={2024},
  publisher={Springer}
}

@article{liu2026fem,
  title={A FEM-PINN approach to modelling elastoplastic soil behaviour in boundary value problems},
  author={Liu, Mingpeng and Zhang, Qinghua and Fuentes, Raul},
  journal={Computers and Geotechnics},
  volume={190},
  pages={107761},
  year={2026},
  publisher={Elsevier}
}

@article{hall2021ginns,
  title={Ginns: Graph-informed neural networks for multiscale physics},
  author={Hall, Eric J and Taverniers, S{\o}ren and Katsoulakis, Markos A and Tartakovsky, Daniel M},
  journal={Journal of Computational Physics},
  volume={433},
  pages={110192},
  year={2021},
  publisher={Elsevier}
}

@article{han2023neural,
  title={A neural network approach for homogenization of multiscale problems},
  author={Han, Jihun and Lee, Yoonsang},
  journal={Multiscale Modeling \& Simulation},
  volume={21},
  number={2},
  pages={716--734},
  year={2023},
  publisher={SIAM}
}

@article{rath2024boosting,
  title={Boosting deep neural networks with geometrical prior knowledge: A survey},
  author={Rath, Matthias and Condurache, Alexandru Paul},
  journal={Artificial Intelligence Review},
  volume={57},
  number={4},
  pages={95},
  year={2024},
  publisher={Springer}
}

@article{gerken2023geometric,
  title={Geometric deep learning and equivariant neural networks},
  author={Gerken, Jan E and Aronsson, Jimmy and Carlsson, Oscar and Linander, Hampus and Ohlsson, Fredrik and Petersson, Christoffer and Persson, Daniel},
  journal={Artificial Intelligence Review},
  volume={56},
  number={12},
  pages={14605--14662},
  year={2023},
  publisher={Springer}
}

@article{ross2024geometric,
  title={A geometric framework for understanding memorization in generative models},
  author={Ross, Brendan Leigh and Kamkari, Hamidreza and Wu, Tongzi and Hosseinzadeh, Rasa and Liu, Zhaoyan and Stein, George and Cresswell, Jesse C and Loaiza-Ganem, Gabriel},
  journal={arXiv preprint arXiv:2411.00113},
  year={2024}
}

@article{lobashev2025hessian,
  title={Hessian Geometry of Latent Space in Generative Models},
  author={Lobashev, Alexander and Guskov, Dmitry and Larchenko, Maria and Tamm, Mikhail},
  journal={arXiv preprint arXiv:2506.10632},
  year={2025}
}

@article{papillon2304architectures,
  title={Architectures of Topological Deep Learning: A Survey on Topological Neural Networks, April 2023},
  author={Papillon, Mathilde and Sanborn, Sophia and Hajij, Mustafa and Miolane, Nina},
  journal={arXiv preprint arXiv:2304.10031},
  year={2023}
}

@article{oldenburg2022geometry,
  title={Geometry aware physics informed neural network surrogate for solving Navier--Stokes equation (GAPINN)},
  author={Oldenburg, Jan and Borowski, Finja and {\"O}ner, Alper and Schmitz, Klaus-Peter and Stiehm, Michael},
  journal={Advanced Modeling and Simulation in Engineering Sciences},
  volume={9},
  number={1},
  pages={8},
  year={2022},
  publisher={Springer}
}

@article{raissi2019physics,
  title={Physics-informed neural networks: A deep learning framework for solving forward and inverse problems involving nonlinear partial differential equations},
  author={Raissi, Maziar and Perdikaris, Paris and Karniadakis, George E},
  journal={Journal of Computational physics},
  volume={378},
  pages={686--707},
  year={2019},
  publisher={Elsevier}
}

@article{costabal2024delta,
  title={$\Delta$-PINNs: Physics-informed neural networks on complex geometries},
  author={Costabal, Francisco Sahli and Pezzuto, Simone and Perdikaris, Paris},
  journal={Engineering Applications of Artificial Intelligence},
  volume={127},
  pages={107324},
  year={2024},
  publisher={Elsevier}
}

@article{kissas2020machine,
  title={Machine learning in cardiovascular flows modeling: Predicting arterial blood pressure from non-invasive 4D flow MRI data using physics-informed neural networks},
  author={Kissas, Georgios and Yang, Yibo and Hwuang, Eileen and Witschey, Walter R and Detre, John A and Perdikaris, Paris},
  journal={Computer methods in applied mechanics and engineering},
  volume={358},
  pages={112623},
  year={2020},
  publisher={Elsevier}
}

@article{cao2026physics,
  title={Physics-informed neural networks for modeling turbulent combustion},
  author={Cao, Zhen and Luo, Kun and Liu, Kai and Cheng, Yuzhou and Xing, Jiangkuan and Jiang, Liang and Fan, Jianren},
  journal={Fuel},
  volume={405},
  pages={136475},
  year={2026},
  publisher={Elsevier}
}

@article{quita2023conservative,
  title={Conservative Physics-Informed Neural Networks for Non-Conservative Hyperbolic Conservation Laws Near Critical States},
  author={Quita, Reyna and Chen, Yu-Shuo and Hu, Hsin-Yi Lee Alex C and Hong, John M},
  journal={arXiv preprint arXiv:2305.12817},
  year={2023}
}

@article{kharazmi2019variational,
  title={Variational physics-informed neural networks for solving partial differential equations},
  author={Kharazmi, Ehsan and Zhang, Zhongqiang and Karniadakis, George Em},
  journal={arXiv preprint arXiv:1912.00873},
  year={2019}
}

@article{rumelhart1986learning,
  title={Learning representations by back-propagating errors},
  author={Rumelhart, David E and Hinton, Geoffrey E and Williams, Ronald J},
  journal={nature},
  volume={323},
  number={6088},
  pages={533--536},
  year={1986},
  publisher={Nature Publishing Group UK London}
}

@article{wang2021eigenvector,
  title={On the eigenvector bias of Fourier feature networks: From regression to solving multi-scale PDEs with physics-informed neural networks},
  author={Wang, Sifan and Wang, Hanwen and Perdikaris, Paris},
  journal={Computer Methods in Applied Mechanics and Engineering},
  volume={384},
  pages={113938},
  year={2021},
  publisher={Elsevier}
}

@article{zhang2024application,
  title={Application of Fourier feature physics-information neural network in model of pipeline conveying fluid},
  author={Zhang, Ting and Yan, Rui and Zhang, Siqian and Yang, Dingying and Chen, Anhao},
  journal={Thin-Walled Structures},
  volume={198},
  pages={111693},
  year={2024},
  publisher={Elsevier}
}

@article{yu2022gradient,
  title={Gradient-enhanced physics-informed neural networks for forward and inverse PDE problems},
  author={Yu, Jeremy and Lu, Lu and Meng, Xuhui and Karniadakis, George Em},
  journal={Computer Methods in Applied Mechanics and Engineering},
  volume={393},
  pages={114823},
  year={2022},
  publisher={Elsevier}
}

@article{lu2021deepxde,
  title={DeepXDE: A deep learning library for solving differential equations},
  author={Lu, Lu and Meng, Xuhui and Mao, Zhiping and Karniadakis, George Em},
  journal={SIAM review},
  volume={63},
  number={1},
  pages={208--228},
  year={2021},
  publisher={SIAM}
}

@article{wu2023comprehensive,
  title={A comprehensive study of non-adaptive and residual-based adaptive sampling for physics-informed neural networks},
  author={Wu, Chenxi and Zhu, Min and Tan, Qinyang and Kartha, Yadhu and Lu, Lu},
  journal={Computer Methods in Applied Mechanics and Engineering},
  volume={403},
  pages={115671},
  year={2023},
  publisher={Elsevier}
}

@article{paszke2017automatic,
  title={Automatic differentiation in pytorch},
  author={Paszke, Adam and Gross, Sam and Chintala, Soumith and Chanan, Gregory and Yang, Edward and DeVito, Zachary and Lin, Zeming and Desmaison, Alban and Antiga, Luca and Lerer, Adam},
  year={2017}
}

@inproceedings{yao2023multiadam,
  title={Multiadam: Parameter-wise scale-invariant optimizer for multiscale training of physics-informed neural networks},
  author={Yao, Jiachen and Su, Chang and Hao, Zhongkai and Liu, Songming and Su, Hang and Zhu, Jun},
  booktitle={International Conference on Machine Learning},
  pages={39702--39721},
  year={2023},
  organization={PMLR}
}

@article{wright1999numerical,
  title={Numerical optimization},
  author={Wright, Stephen and Nocedal, Jorge and others},
  journal={Springer Science},
  volume={35},
  number={67-68},
  pages={7},
  year={1999}
}

@article{kingma2014adam,
  title={Adam: A method for stochastic optimization},
  author={Kingma, Diederik P},
  journal={arXiv preprint arXiv:1412.6980},
  year={2014}
}

@article{zhang2026ghs,
  title={GHS-TDA: A Synergistic Reasoning Framework Integrating Global Hypothesis Space with Topological Data Analysis},
  author={Zhang, Jiaquan and Zhang, Chaoning and Chen, Shuxu and Wang, Xudong and Huang, Zhenzhen and Zheng, Pengcheng and Yuan, Shuai and Zheng, Sheng and Sun, Qigan and Zou, Jie and others},
  journal={arXiv preprint arXiv:2602.09794},
  year={2026}
}

@article{zhang2026text,
  title={Text summarization via global structure awareness},
  author={Zhang, Jiaquan and Zhang, Chaoning and Chen, Shuxu and Liu, Yibei and Li, Chenghao and Sun, Qigan and Yuan, Shuai and Puspitasari, Fachrina Dewi and Han, Dongshen and Wang, Guoqing and others},
  journal={arXiv preprint arXiv:2602.09821},
  year={2026}
}
\bibliographystyle{icml2026}

\newpage
\appendix
\onecolumn
\section{Benchmarks}
\label{app:1}
In this section, we present the five PDEs used in our experiments to validate the effectiveness of our method under specific scenarios.

\textbf{1D Steady Burgers Equation.} This canonical benchmark for PINNs takes the form
\begin{equation}
-\nu\,u_{xx} + u\,u_x = f(x),\quad x\in(0,1),
\end{equation}
which contains a diffusion term $-\nu u_{xx}$ and a nonlinear convection term $u u_x$. The diffusion coefficient $\nu$ represents viscosity or diffusion strength, controlling solution smoothness and suppressing gradient blow-up from nonlinear convection. In our experiments, we set $\nu = 0.1$, which avoids sharp shocks while preserving noticeable nonlinearity. This equation demonstrates GC-PINN performance in a general scenario and shows that the method is not overfit to a specific case.

\textbf{1D Convection–Diffusion Equation} tests the method's capability to capture boundary layers under extreme conditions:
\begin{equation}
-\nu\,u_{xx} + a\,u_x = f(x),\quad x\in(0,1),
\end{equation}
with a small diffusion coefficient $\nu = 10^{-3}$ and a convection velocity $a=1.0$ directing information flow from left to right. The dimensionless Péclet number $\mathrm{Pe} = a/\nu = 1000 \gg 1$ indicates convection-dominated behavior. A thin exponential boundary layer forms near $x=0$, with thickness $\nu/a$, presenting steep gradients and multiscale residual spectra challenges for PINNs.

\textbf{1D Helmholtz Equation.} This problem introduces multiscale features to examine high-frequency oscillation capture:
\begin{equation}
u_{xx} + k^2 u = f(x),\quad x\in(0,1),
\end{equation}
with a manufactured solution $u = \sin(2\pi m x)$. The wavenumber $k=10$ controls stiffness, while the frequency $m=5$ sets the oscillation scale, corresponding to a wavelength $\lambda = 1/m$.

\textbf{2D Convection–Diffusion Equation.} To test anisotropic boundary layer handling, we consider
\begin{equation}
-\varepsilon\Delta u + \mathbf{b}\cdot\nabla u = f(x,y),\quad (x,y)\in(0,1)^2,
\end{equation}
with a small diffusion coefficient $\varepsilon=0.01$ and convection velocity $\mathbf{b}=(1.0,1.0)$ along the diagonal. The manufactured solution combines a smooth term $\sin(\pi x)\sin(\pi y)$ and an exponential boundary layer $A\exp(-(1-x)/\varepsilon_{\text{layer}})$ with amplitude $A=0.8$ and thickness $\varepsilon_{\text{layer}}=0.01$, forming a typical boundary layer in $x\in[0.99,1]$.

\textbf{2D Steady Navier–Stokes Equation} evaluates PINN performance in fluid boundary layer problems. The incompressible form is
\begin{equation}
\begin{cases}
\nabla\cdot\mathbf{u} = 0,\\
\mathbf{u}\cdot\nabla\mathbf{u} = -\nabla p + \nu\Delta\mathbf{u} + \mathbf{f},
\end{cases}
\quad (x,y)\in[0,1]^2,
\end{equation}
where the density $\rho=1$ and viscosity $\nu=0.01$, corresponding to moderate Reynolds number, where nonlinear effects are evident but turbulence does not arise. The manufactured velocity field combines a smooth component and an exponential boundary layer:
\begin{equation}
\begin{aligned}
u(x,y) &= \sin(\pi y)\bigl(1 + A e^{-(1-x)/\varepsilon_{\text{layer}}}\bigr),\\
v(x,y) &= \frac{A}{\varepsilon_{\text{layer}}}e^{-(1-x)/\varepsilon_{\text{layer}}}\frac{\cos(\pi y)-1}{\pi}.
\end{aligned}
\end{equation}
And the layer amplitude $A=0.3$ controlling perturbation strength and thickness $\varepsilon_{\text{layer}}=0.01$, producing an exponential velocity boundary layer near $x=1$. The pressure field is
\begin{equation}
p(x,y)=B\sin(2\pi x)\sin(2\pi y),\quad B=0.5,
\end{equation}
regulating the coupling between pressure gradient and velocity, effectively modeling multiscale near-wall flow at moderate to high Reynolds numbers.

\section{More Experimental Setup Details}
\label{app:2}
\subsection{Baselines}
\label{app:2.1}
For the baseline methods used in comparison, all models share the same backbone network, namely a standard PINN based on an MLP. Specific implementation details of each mechanism are described as follows.

\begin{itemize}
\item \textbf{FF-PINN.} Based on Fourier feature embeddings, we adopt a prescribed set of frequencies $(0.5, 1.0, 1.5, 2.0, 2.5, 3.0)$ to cover low-to-mid frequency bands of the solution. The input coordinate $x$ is first projected via matrix multiplication $2\pi x f^{\top}$, and the corresponding sine and cosine components are concatenated, resulting in a $13$-dimensional feature vector that is fed into a standard MLP. The frequency parameters are fixed and not trainable.

\item \textbf{SA-PINN.} We reproduce the adaptive scalar weighting mechanism by introducing two trainable parameters $\log w_{\mathrm{res}}$ and $\log w_{\mathrm{bc}}$, which control the contributions of the residual loss and boundary loss, respectively. During forward propagation, the weights are transformed via the exponential function and clipped to $[10^{-3}, 10^{3}]$ to ensure positivity and boundedness. Before switching to the L-BFGS optimizer, the weights are frozen as constants to prevent instability caused by the sensitivity of second-order optimization to the weight landscape.

\item \textbf{PINN+RAR.} This method implements residual-based adaptive refinement through dynamic sampling. During the Adam pretraining stage, every $500$ iterations, residual magnitudes are evaluated over $8{,}000$ uniformly sampled candidate points, and the top $500$ points with the largest residuals are added to the training pool. The pool capacity is capped at $60{,}000$ to avoid excessive memory usage. Sampling is performed with replacement and combined with random shuffling to maintain i.i.d.\ mini-batches. Once entering the L-BFGS stage, the training pool is fixed, and no new points are introduced, preserving the deterministic closure required by second-order optimization. 

\item \textbf{gPINN.} In addition to the standard PINN loss, gPINN introduces a regularization term $\mathbb{E}\|\partial_x \mathcal{R}\|^2$. In implementation, second-order automatic differentiation is used to compute the spatial derivative of the residual $\partial_x \mathcal{R}$, which is normalized by the energy of the predicted field gradient $\partial_x u_\theta$ to stabilize the scale and prevent numerical blow-up caused by small denominators. This term is enabled only during the Adam stage, with a delayed activation strategy consisting of a $2{,}000$-iteration warm-up followed by a gradual ramp-up over another $2{,}000$ iterations. The base weight is set to $\lambda_g = 5 \times 10^{-6}$ to avoid excessive regularization, and hard clipping up to $100$ is applied to suppress abnormal batches. The gPINN term is completely disabled during the L-BFGS stage to prevent higher-order differential operators from inducing ill-conditioned Hessians.
\end{itemize}

\subsection{Implementation}
\label{app:2.2}
GC-PINN is trained on a machine equipped with AMD EPYC 7763 CPUs and NVIDIA RTX 4090 GPUs with 48 GB memory. The software environment uses Python 3.13.7 and PyTorch 2.5.1, with CUDA support up to version 13.0. In practice, most hyperparameters are determined empirically. The boundary condition loss (ICLoss) is weighted by 100, as PDE residual is integrated over the entire domain and typically has a much larger magnitude than boundary loss evaluated only on the boundary \cite{yao2023multiadam}. This weighting balances gradient contributions and prevents the boundary term from being overwhelmed in the total loss.

Under the PyTorch framework, \emph{$\phi$} is jointly encapsulated with the MLP as an \texttt{nn.Module}, and all derivatives are computed on the fly via \texttt{torch.autograd.grad} through automatic chaining. Jacobians and Hessians are backpropagated automatically without manual derivation, ensuring accurate gradient propagation between the mapping and the PDE residual.

For the remaining hyperparameters, the Radial Mapping uses $\alpha = 20$, and the Local Stretching sets $\beta = 10.0$. In local stretching corresponding formulations, the initialization point $x$ is chosen as the center of physical domain. In addition, a regularization term with coefficient $10^{-6}$ is introduced to prevent degeneracy of the mappings.

\section{Theoretical Validation}
\label{app:3}
In this section, we provide additional proofs and derivations to further substantiate the arguments presented in the main text.

\subsection{Analysis of Challenges in Multiscale PDEs}
\label{app:3.1}
To quantitatively characterize the three challenges discussed in Section~\ref{sec:3.1}, we present the following analysis.

\textbf{Operator-Scale Amplification.} We consider a multiscale solution defined on the interval $(0,1)$,
\begin{equation}
u(x)=\sin(2\pi x)+\sin\!\left(\frac{2\pi x}{\epsilon}\right),
\end{equation}
where $\epsilon \ll 1$ denotes the smallest spatial scale. Within the singular layer, its second derivative satisfies $u_{xx}=O(\epsilon^{-2})$, while in smooth regions it remains $O(1)$. 

For a $k$-th order differential operator $\mathcal N$, the residual magnitude in regions characterized by scale $\epsilon$ satisfies $\|\mathcal N[u]\| = O(\epsilon^{-k})$, where $\|\cdot\|$ denotes a pointwise or local norm of the residual. Let the spatial discretization resolution be $\Delta x$. 

Then, the discrete residual energy scale within the singular region is $(\epsilon^{-k}\Delta x)^2$, whereas in smooth regions it scales as $(\Delta x)^2$. The ratio between the two is therefore $\epsilon^{-2k}$. When $\epsilon \ll 1$, the induced PDE constraint matrix $\mathbf A$ exhibits a condition number satisfying $\kappa(\mathbf A) \sim \epsilon^{-k}$, indicating pronounced ill-conditioning.

\textbf{Hessian Condition Number Growth.}
The residual-based loss function is defined as
\begin{equation}
\mathcal L(\theta)=\frac{1}{N}\sum_{i=1}^{N}\|\mathcal R_i\|^2,
\end{equation}
where $\theta$ denotes the network parameters, $\mathcal R_i$ is the PDE residual evaluated at the $i$-th sample point, and $N$ is the total number of samples. The corresponding Hessian matrix can be written as
\begin{equation}
\mathbf H=\nabla_\theta^2\mathcal L
= \frac{2}{N}\sum_{i=1}^{N}
\left[
\nabla_\theta\mathcal R_i\,\nabla_\theta\mathcal R_i^\top
- \mathcal R_i\,\nabla_\theta^2\mathcal R_i
\right].
\end{equation}
In the vicinity of a singular point $x_s$, the residual and its parameter gradient satisfy
\begin{equation}
\mathcal R_i \sim O(\epsilon^{-k}), \qquad
\nabla_\theta\mathcal R_i \sim O(\epsilon^{-k}),
\end{equation}
which leads to a locally dominant eigenvalue $\lambda_{\max} \sim \epsilon^{-2k}$. In contrast, contributions from smooth regions correspond to a minimal eigenvalue $\lambda_{\min} \sim O(1)$. Consequently, the global condition number of the Hessian satisfies
\begin{equation}
\kappa(\mathbf H)=\frac{\lambda_{\max}}{\lambda_{\min}}
\sim \epsilon^{-2k}.
\end{equation}
The corresponding parameter update can be approximated as $\Delta\theta \approx \mathbf H^{-1}\nabla_\theta\mathcal L$, whose direction is dominated by singular regions. The alignment between this update direction and the gradient contributions from smooth regions satisfies $\cos\theta \sim O(\epsilon^{k})$, indicating an almost orthogonal relationship that severely suppresses global convergence.

\textbf{Imbalanced Residual Energy.}
Let the volume of the singular region scale as $V_s \sim \epsilon^{d}$, where $d$ denotes the spatial dimension, while the smooth region volume satisfies
$V_f \sim 1$.
The total residual energy can then be decomposed as
\begin{equation}
\mathcal L
= V_s \cdot O(\epsilon^{-2k}) + V_f \cdot O(1)
= O(\epsilon^{\,d-2k}) + O(1).
\end{equation}
When $d<2k$, as in convection-dominated problems, the first term becomes dominant. In this regime, the optimizer allocates approximately $1 - O(\epsilon^{2k-d})$ of the effective update steps to the singular region, which occupies only a negligible fraction of the domain. As a result, gradients in smooth regions are nearly vanishing, while those in singular regions are excessively large. Consequently, the adaptive learning rate of Adam is forced to decrease to $\eta \sim O(\epsilon^{2k-d})$, leading to an exponential slowdown in the global convergence rate.

\subsection{Torus Mapping}
\label{app:3.2}
\textbf{Definition and Periodicity.} From a mathematical perspective, the core advantage of the Torus Mapping lies in embedding the real line $\mathbb{R}$ into the compact manifold $S^1$. After normalizing the physical domain to \([0,1]\), we define
\begin{equation}
\phi_{\mathrm{tor}}(x)=\frac{1}{2\pi}\operatorname{atan2}\!\bigl(\sin(2\pi x),\,\cos(2\pi x)\bigr)\in(-\tfrac12,\tfrac12].
\end{equation}
The two-argument arctangent \(\operatorname{atan2}(y,x)=\arg(x+iy)\) remains continuous even at \(x=0\), and for any \(\theta\) satisfies
\begin{equation}
\operatorname{atan2}(\sin\theta,\cos\theta)=\theta\;(\operatorname{mod}2\pi).
\end{equation}

Consequently, \(\phi_{\mathrm{tor}}(x+1)=\phi_{\mathrm{tor}}(x)\), realizing a surjective mapping \(\mathbb{R}\to S^1\) with period 1. This construction intrinsically enforces periodic boundary coupling, without requiring additional constraints.

\textbf{Gradient Uniformization.} Regarding gradient propagation, taking derivatives with respect to a single component yields
\begin{equation}
\phi_{\mathrm{tor}}'(x)=1,\qquad \phi_{\mathrm{tor}}''(x)=0.
\end{equation}

The Jacobian is the identity mapping, and the second derivative vanishes. This implies that, regardless of the location in the physical coordinate, the backpropagated gradient
\begin{equation}
\nabla_{\!x}u=\nabla_{\!\xi}\tilde u\cdot\mathbf J_{\phi},\quad \mathbf J_{\phi}=\mathbf I,
\end{equation}
maintains a uniform scale across the domain, thereby avoiding the gradient non-uniformity that may arise in the original coordinate system.

\subsection{Radial Mapping}
\label{app3.3}
In unbounded domains or shock problems, the far-field region \(|x|\to\infty\) often only requires asymptotic resolution, while the near-field demands high resolution. The Radial Mapping compresses the far-field into a finite interval via a logarithmic rate, allowing the network to cover the entire domain with fixed capacity and automatically expanding the mesh near shocks. From Eq.~\eqref{eq:1}, as \(|x|\to\infty\), 
\begin{equation}
\phi_{\mathrm{rad}}(x)\sim \pm \frac{\log|x|}{\log(1+\alpha)}
\end{equation}
grows logarithmically, while any finite \(x\) is mapped approximately into \([-1,1]\), achieving geometric compactification.

\textbf{Derivative Decay.}  
The first derivative
\begin{equation}
\phi_{\mathrm{rad}}'(x)=\frac{\alpha}{(1+\alpha|x|)\log(1+\alpha)}
\end{equation}
decays hyperbolically as \(O(|x|^{-1})\) in the far field, consistent with asymptotic behavior of potential or diffusion solutions. The second derivative
\begin{equation}
\phi_{\mathrm{rad}}''(x)=-\frac{\alpha^{2}\operatorname{sgn}(x)}{(1+\alpha|x|)^2\log(1+\alpha)}
\end{equation}
decays as \(O(|x|^{-2})\), naturally reducing the diffusion term \(\nu u_{xx}\) in the transformed space without additional asymptotic constraints.

\textbf{Adaptive Resolution.}  
The parameter \(\alpha\) is adjusted in real time via backpropagation:
\begin{itemize}
\item Increasing \(\alpha\) smooths the mapping, further compressing the far-field and enhancing compactification if residuals remain large;
\item Decreasing \(\alpha\) approximates the identity, preserving local details.
\end{itemize}
For a shock solution \(u(x)=\tanh((x-x_0)/\delta)\), the Local Stretching factor
\begin{equation}
\frac{\mathrm{d}\phi}{\mathrm{d}x}\Big|_{x_0} \propto \frac{1}{1+\alpha|x_0|}
\end{equation}
amplifies the effective mesh spacing at the shock center, whereas conventional grids require manual refinement, despite \(\alpha\) needing manual initialization.

\textbf{Gradients and Condition Number.}  
The Jacobian norm \(\|\phi'\|\sim O(1)\) in the near-field and \(O(|x|^{-1})\) in the far-field ensures uniform gradient scales during backpropagation. Moreover, far-field compression reduces the operator eigenvalue ratio from \(\kappa \sim |x|_{\max}\) to \(\kappa \sim \log|x|_{\max}\), significantly improving the Hessian condition number and accelerating convergence.

\subsection{Local Stretching}
\label{app3.4}
The Local Stretching is implemented via a Gaussian gating function
\begin{equation}
w(x)=\exp[-\beta(x-c)^2],
\end{equation}
which provides a movable and scalable high-resolution window for the network without introducing discontinuous interfaces. Similarly, from Eq.~\eqref{eq:2}, when \(|\mathbf{x}|\ll 1/\sqrt{\beta}\), \(w\to 1\) and the mapping behaves like a \(\tanh\) stretch; when \(|\mathbf{x}|\gg 1/\sqrt{\beta}\), \(w\to 0\) and the mapping degenerates to the identity, achieving a \(C^{\infty}\) smooth transition.

\textbf{Local Stretching and Resolution.}  
The radial Jacobian at the center reads
\begin{equation}
\nabla\phi_{\mathrm{loc}}(\mathbf{0})=(1+\beta)\mathbf{I},
\end{equation}
amplifying physical coordinates by a factor of \(\beta\). For a local feature of thickness \(\delta\), choosing \(\beta\sim 1/\delta\) expands the feature to an \(O(1)\) width in the domain, making it resolvable by a standard grid. Second-order derivatives provide curvature correction, assisting the balance between convection and diffusion.

\textbf{Learnable Focusing.}  
The parameter \(\beta\) is the only trainable variable. The gradient of the loss with respect to \(\beta\) is
\begin{equation}
\frac{\partial \mathcal L}{\partial \beta} \propto \mathbb{E}\Bigl[\frac{\partial \mathcal L}{\partial \phi} \cdot w |\mathbf{x}|^2 \tanh(\beta \mathbf{x})\Bigr],
\end{equation}
allowing the network to adaptively enlarge the gate width or adjust its strength based on residuals, without the need for explicit remeshing.

\subsection{Computational Complexity}  
\label{app3.5}
Consider a training set with \(N\) sample points, where the geometric mapping is a pointwise differentiable function. The introduction of the geometric compactification mapping incurs a computational cost of \(O(N)\), while the backbone PINN retains the same complexity as a standard PINN. Consequently, the asymptotic computational complexity of the network remains unchanged, and the additional overhead from the mapping is limited to preprocessing, which is generally negligible.

\section{Neural Tangent Kernel Spectral Analysis}
\label{app:4}

\begin{figure*}[h]
\centering
\begin{subfigure}{0.48\textwidth}
    \centering
    \includegraphics[width=\textwidth]{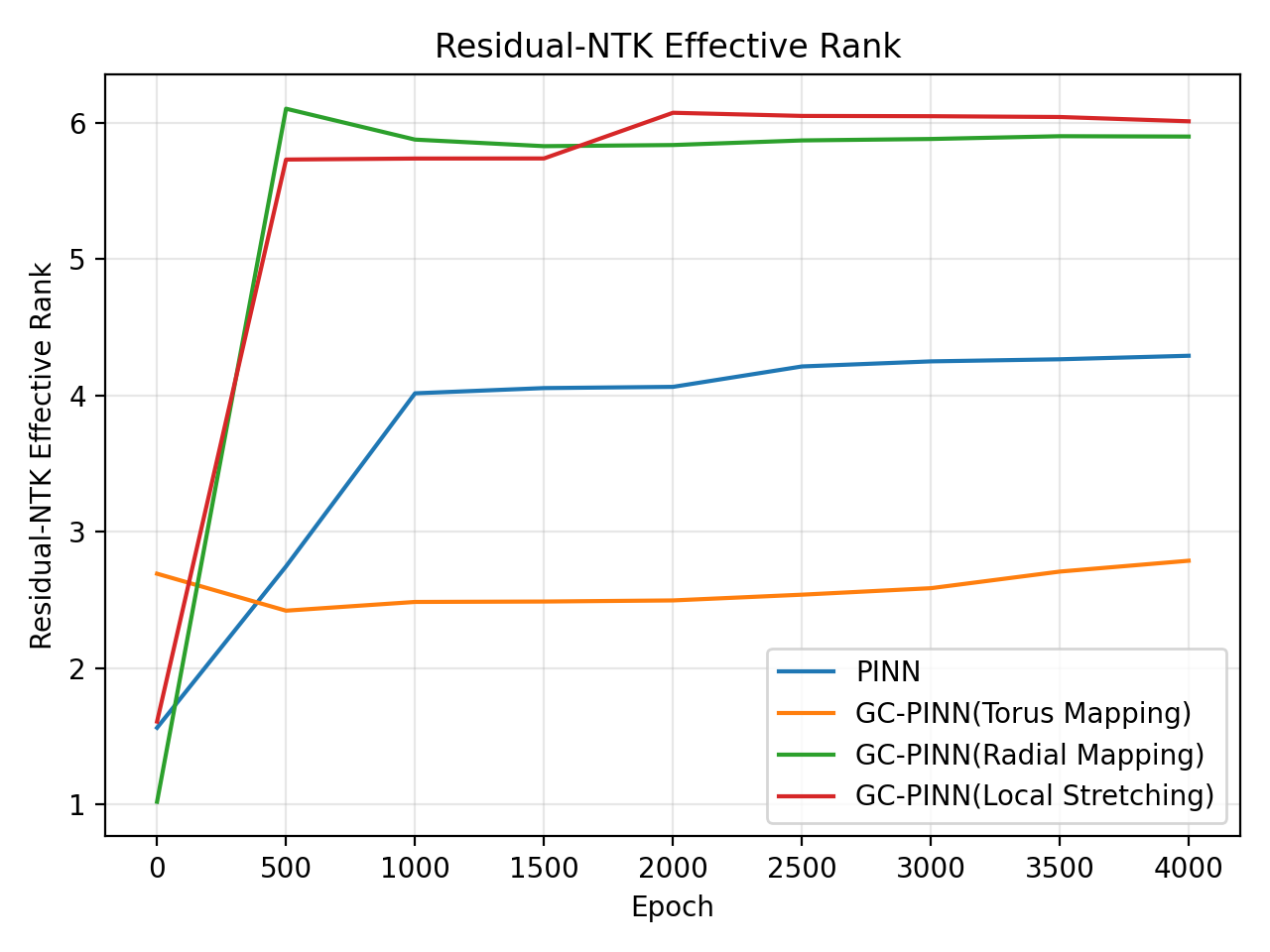}
    \caption{}
    \label{fig:a}
\end{subfigure}
\hfill
\begin{subfigure}{0.48\textwidth}
    \centering
    \includegraphics[width=\textwidth]{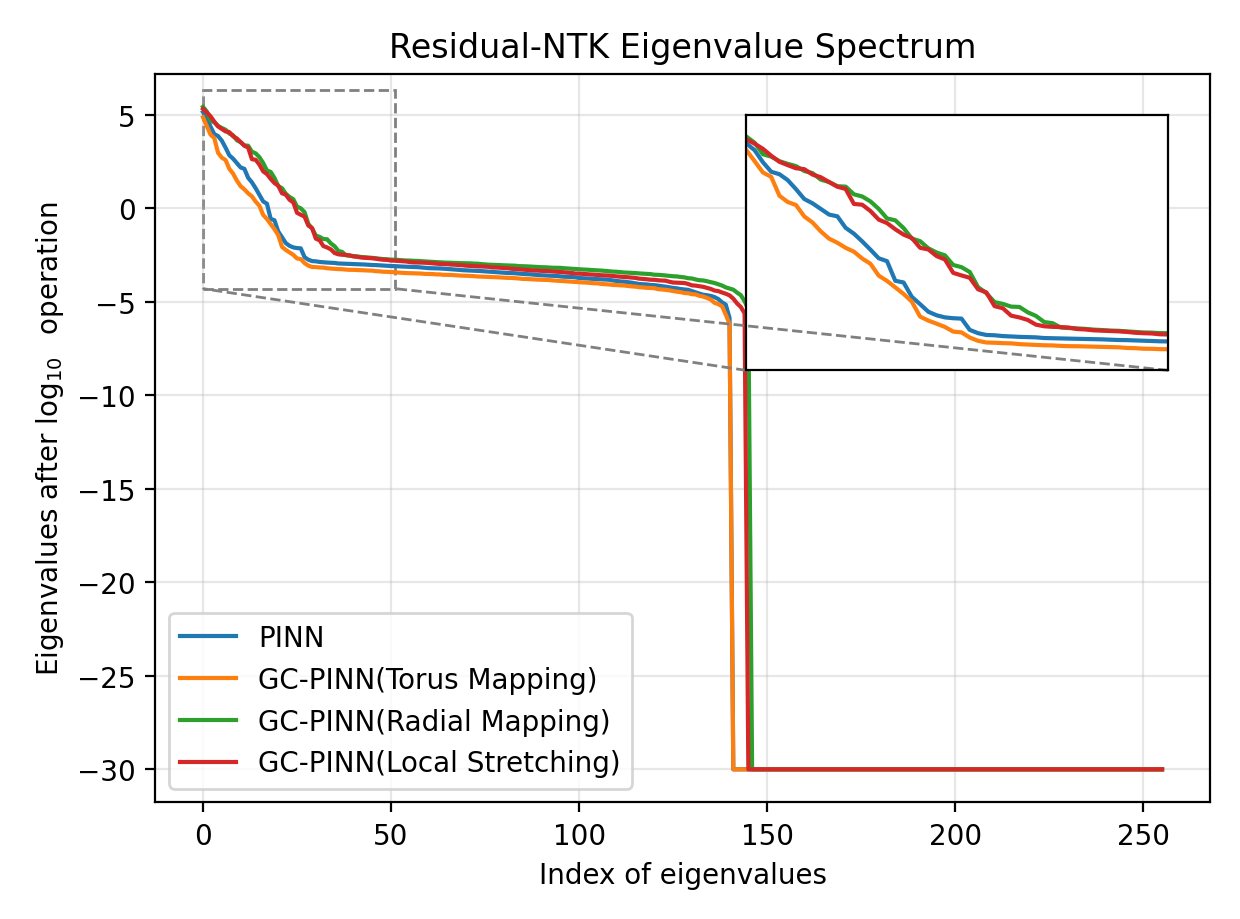}
    \caption{}
    \label{fig:b}
\end{subfigure}
\caption{Spectral analysis of residual neural tangent kernels under geometric compactification. (a) Eigenvalue distribution of the kernel matrix after training convergence. (b) Temporal evolution of the effective rank during training.}
\label{fig:ntk1}
\end{figure*}

\begin{figure*}[h]
\centering
\includegraphics[width=\textwidth]{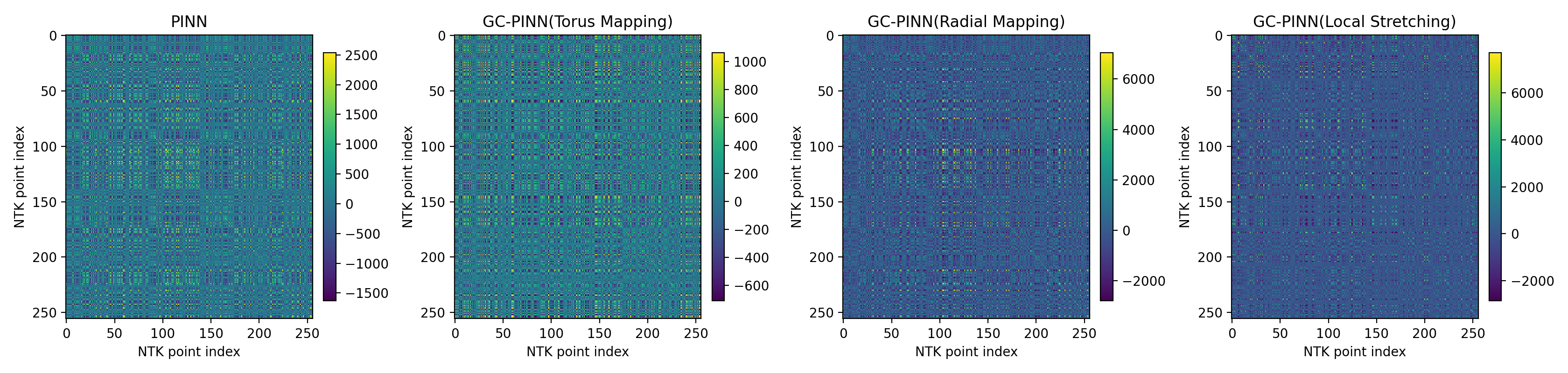}
\caption{Comparison of heatmaps of residual NTK kernel matrices.}
\label{fig:ntk2}
\end{figure*}

In this section, we analyze the spectral properties of the residual neural tangent kernel to characterize the proposed method and the limitations of fixed-coordinate PINNs. Figure~\ref{fig:ntk1} shows the eigenvalue distributions of the residual NTK and the temporal evolution of its effective rank. A higher effective rank indicates that residual gradients carry energy along a larger number of directions. The Local Stretching and Radial mappings increase the effective rank, which corresponds to weaker optimization stiffness and improved trainability. In contrast, the effective rank of Torus Mapping remains at a lower level, consistent with its suppression of high-frequency modes. 

The kernel matrix heatmaps in Fig.~\ref{fig:ntk2} show that the PINN under a fixed Euclidean coordinate system exhibits spectral imbalance and strong global coupling. This behavior is characterized by large off-diagonal responses and a highly dispersed eigenvalue distribution, which corresponds to a degraded optimization condition number and increased gradient stiffness. In contrast, the residual NTK of GC-PINN presents a more concentrated spectral energy distribution with clearer directional structure. The Torus mapping reduces the numerical range of the kernel and removes boundary discontinuities. The Radial mapping redistributes multi-scale energy at a global level, where large-magnitude components appear in sparse and structured patterns with approximate radial symmetry. The Local Stretching further reduces gradient coupling at local scales, leading the kernel matrix toward diagonal dominance. These results show that performance is stable over a moderate hyperparameter range, with limited sensitivity.

\section{Additional Experiments}
\label{app:5}

\subsection{Additional Parameter Sensitivity Analysis}
\begin{table*}[h]
\centering
\caption{Parameter sensitivity analysis for the 1D convection-diffusion and 2D Navier-Stokes equations}
\label{tab:param_sensitivity}
\small
\begin{tabular}{cccc|cccc}
\toprule
\multicolumn{8}{c}{\textbf{1D Convection-Diffusion Equation}} \\
\midrule
\multicolumn{4}{c}{Parameter $\alpha$} & \multicolumn{4}{c}{Parameter $\beta$} \\
\cmidrule(lr){1-4} \cmidrule(lr){5-8}
Param & MSE & Rel $L^2$ & Rel $H^1$ & Param & MSE & Rel $L^2$ & Rel $H^1$ \\
\midrule
5  & 8.22E-06 & 2.87E-03 & 2.99E-03 & 10 & 6.66E-06 & 2.58E-03 & 1.21E-03 \\
10 & 6.22E-06 & 2.50E-03 & 6.05E-03 & 20 & 5.66E-06 & 2.38E-03 & 2.02E-03 \\
15 & 6.07E-06 & 2.47E-03 & 1.34E-02 & 30 & 2.22E-06 & 1.49E-03 & 5.35E-04 \\
20 & \textbf{9.32E-08} & \textbf{3.06E-04} & 6.66E-04 & 40 & 2.01E-07 & 4.49E-04 & \textbf{2.07E-04} \\
25 & 1.93E-07 & 4.39E-04 & 8.97E-04 & 50 & \textbf{9.32E-08} & \textbf{3.06E-04} & 6.66E-04 \\
30 & 3.24E-06 & 1.80E-03 & 4.95E-03 & 60 & 1.82E-07 & 4.26E-04 & 6.10E-04 \\
35 & 3.50E-05 & 5.92E-03 & 5.80E-03 & 70 & 8.31E-07 & 9.12E-04 & 3.56E-04 \\
\midrule
\multicolumn{8}{c}{\textbf{2D Navier-Stokes Equation}} \\
\midrule
\multicolumn{4}{c}{Parameter $\alpha$} & \multicolumn{4}{c}{Parameter $\beta$} \\
\cmidrule(lr){1-4} \cmidrule(lr){5-8}
Param & MSE & Rel $L^2$ & Rel $H^1$ & Param & MSE & Rel $L^2$ & Rel $H^1$ \\
\midrule
5  & 1.73E-04 & 1.40E-02 & 6.30E-03 & 10 & 8.01E-03 & 9.54E-02 & 2.26E-02 \\
10 & 1.16E-04 & 1.15E-02 & 5.15E-03 & 20 & 1.15E-03 & 3.62E-02 & 1.27E-02 \\
15 & 9.33E-05 & 6.90E-03 & 2.09E-03 & 30 & 1.01E-04 & 1.28E-02 & 3.15E-03 \\
20 & \textbf{2.01E-05} & \textbf{4.87E-03} & \textbf{1.85E-03} & 40 & \textbf{9.33E-05} & \textbf{1.03E-02} & \textbf{2.09E-03} \\
25 & 8.92E-05 & 8.09E-03 & 2.40E-03 & 50 & 9.75E-05 & 1.07E-02 & 4.61E-03 \\
30 & 1.07E-04 & 1.10E-02 & 3.80E-03 & 60 & 8.92E-05 & 1.21E-02 & 2.40E-03 \\
35 & 1.24E-04 & 1.19E-02 & 9.19E-03 & 70 & 1.03E-04 & 2.29E-02 & 2.45E-03 \\
\bottomrule
\end{tabular}
\end{table*}

We extend the parameter sensitivity experiments in Section~\ref{sec:5.4} to provide a more comprehensive justification for the choice of hyperparameters, including the selected ranges and allowable variations. Table~\ref{tab:param_sensitivity} reports results on the 1D convection–diffusion and 2D Navier–Stokes equations. The optimal values remain within the ranges reported in the main text, and the resulting variations in error are limited to at most one order of magnitude.

\subsection{Additional Visualization Analysis}
\begin{figure}[h!]
    \centering
    \begin{subfigure}[t]{0.45\textwidth}
        \centering
        \includegraphics[width=\textwidth]{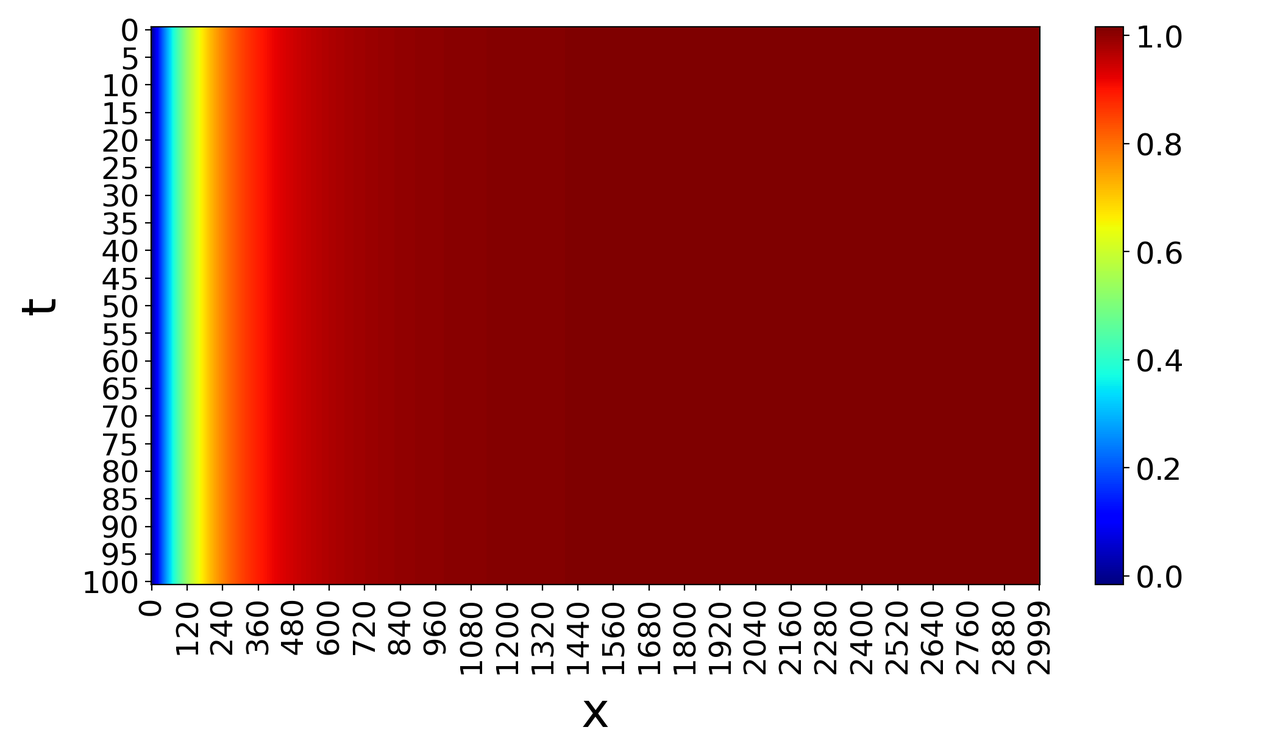}
        \caption{PINN prediction}
        \label{fig:loss1}
    \end{subfigure}
    \hfill
    \begin{subfigure}[t]{0.45\textwidth}
        \centering
        \includegraphics[width=\textwidth]{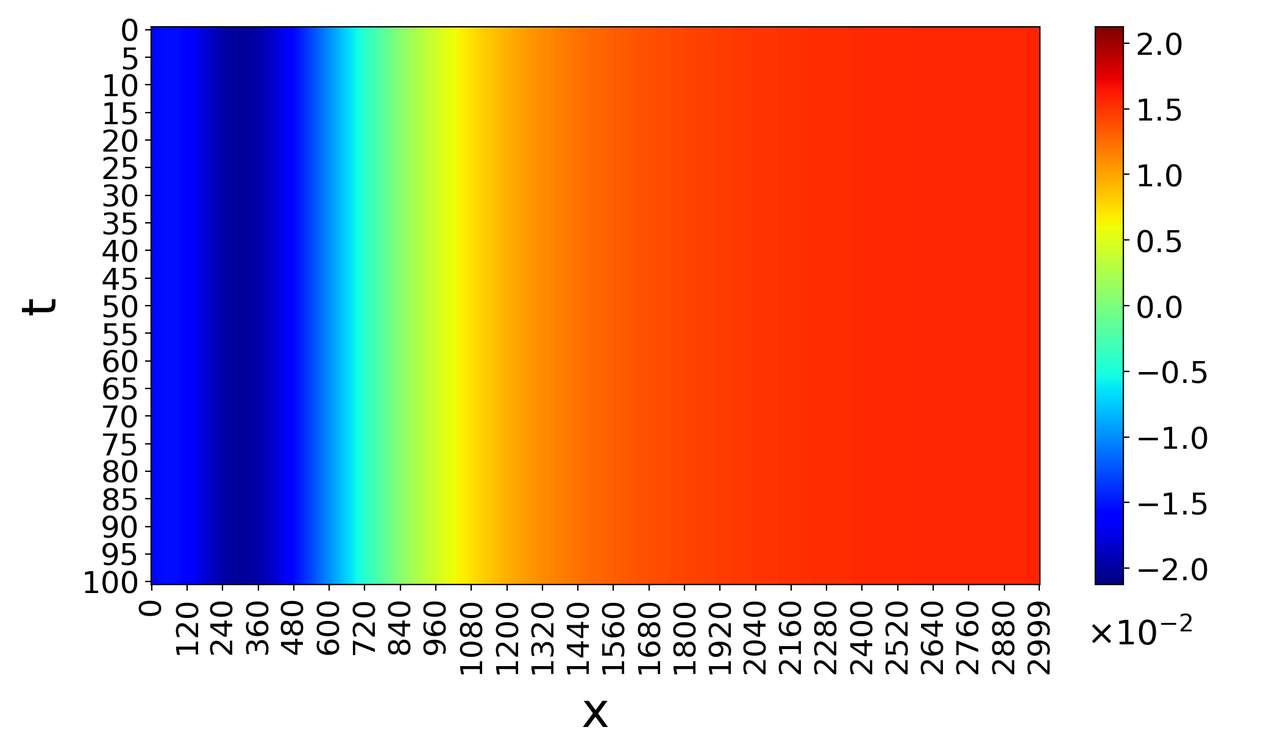}
        \caption{PINN error distribution}
        \label{fig:loss2}
    \end{subfigure}
  
    \begin{subfigure}[t]{0.45\textwidth}
        \centering
        \includegraphics[width=\textwidth]{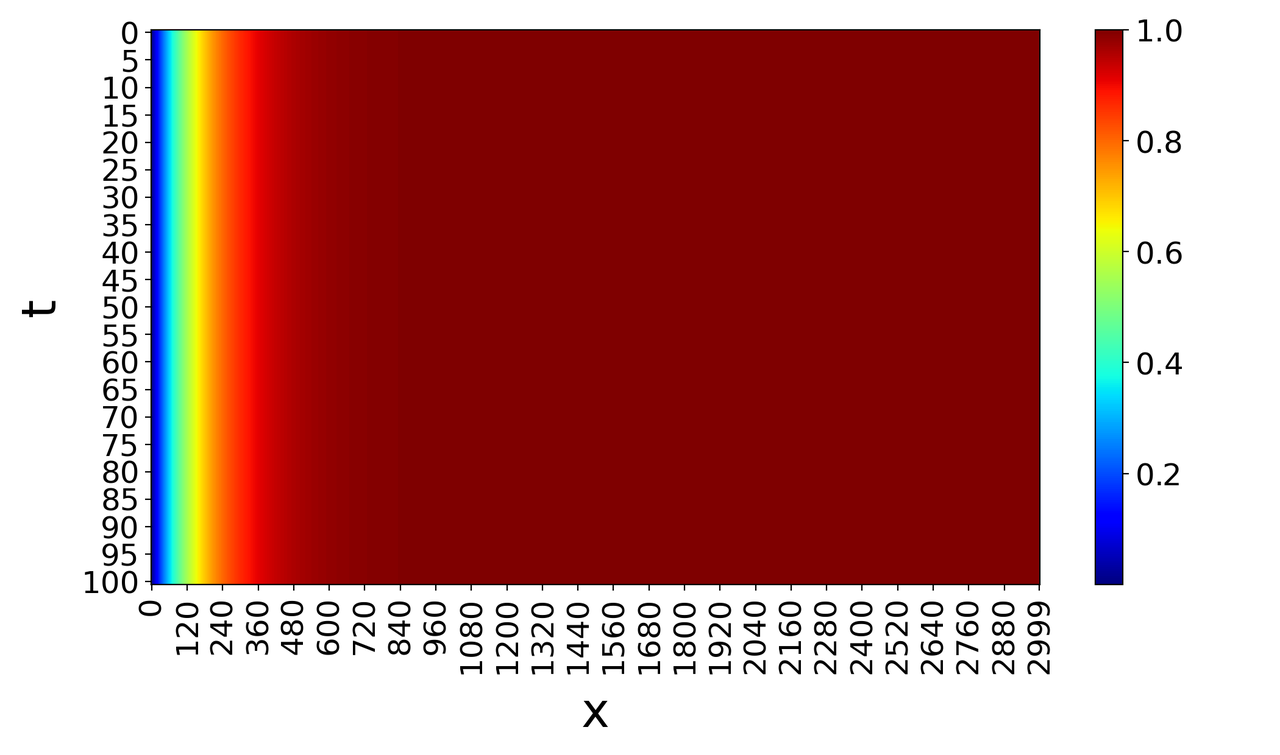}
        \caption{FF-PINN prediction}
        \label{fig:loss3}
    \end{subfigure}
      \hfill
    \begin{subfigure}[t]{0.45\textwidth}
        \centering
        \includegraphics[width=\textwidth]{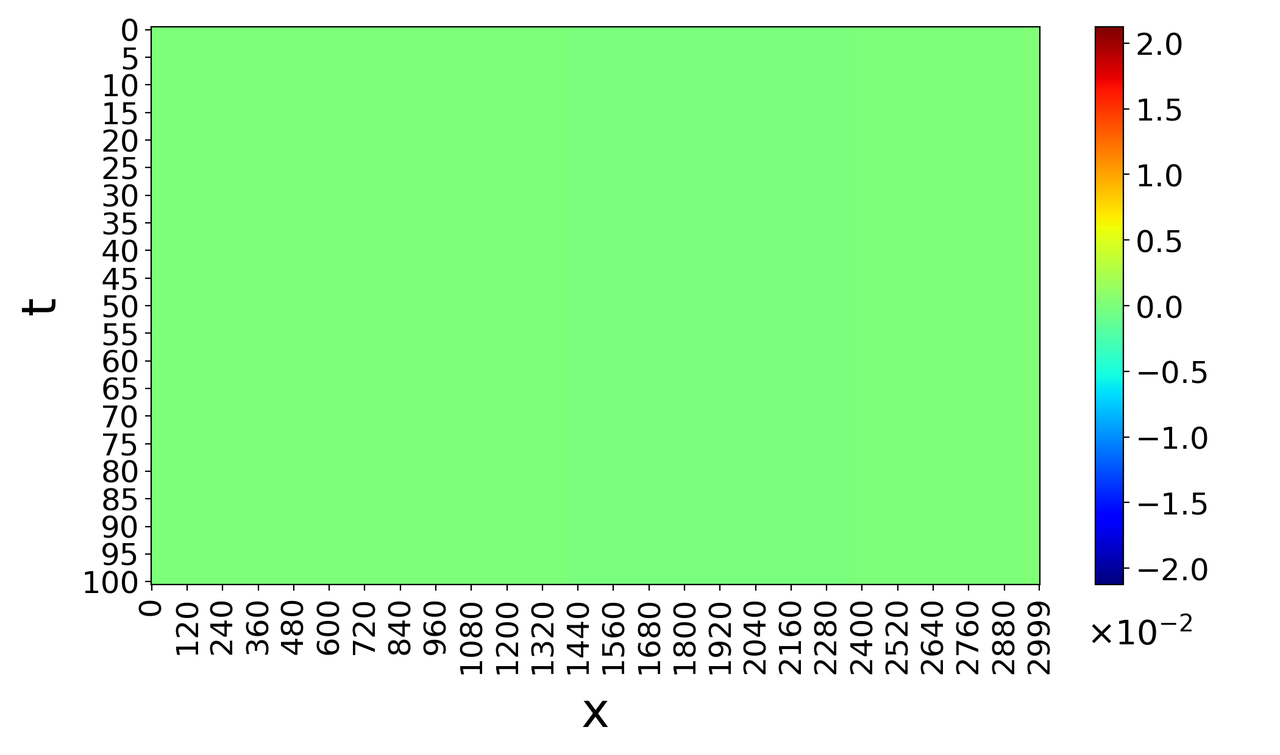}
        \caption{FF-PINN error distributionn}
        \label{fig:loss4}
    \end{subfigure}

    \begin{subfigure}[t]{0.45\textwidth}
        \centering
        \includegraphics[width=\textwidth]{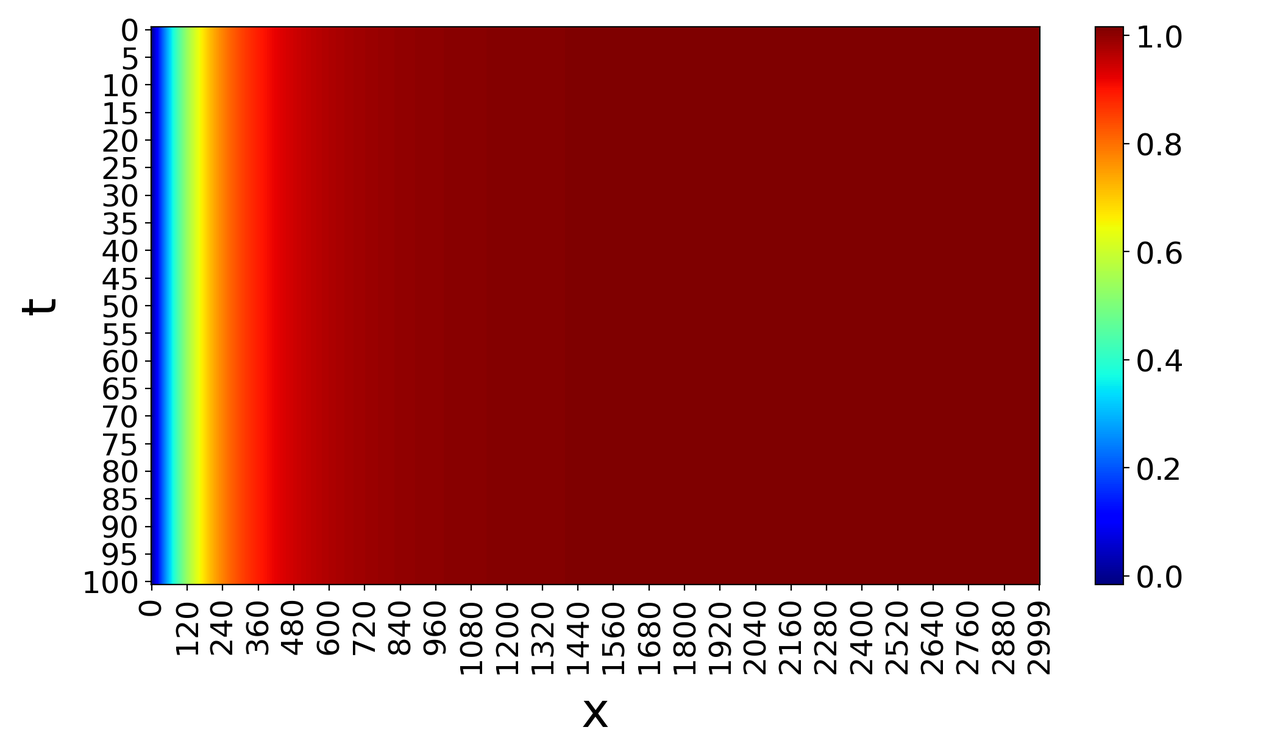}
        \caption{gPINN prediction}
        \label{fig:loss5}
    \end{subfigure}
    \hfill
    \begin{subfigure}[t]{0.45\textwidth}
        \centering
        \includegraphics[width=\textwidth]{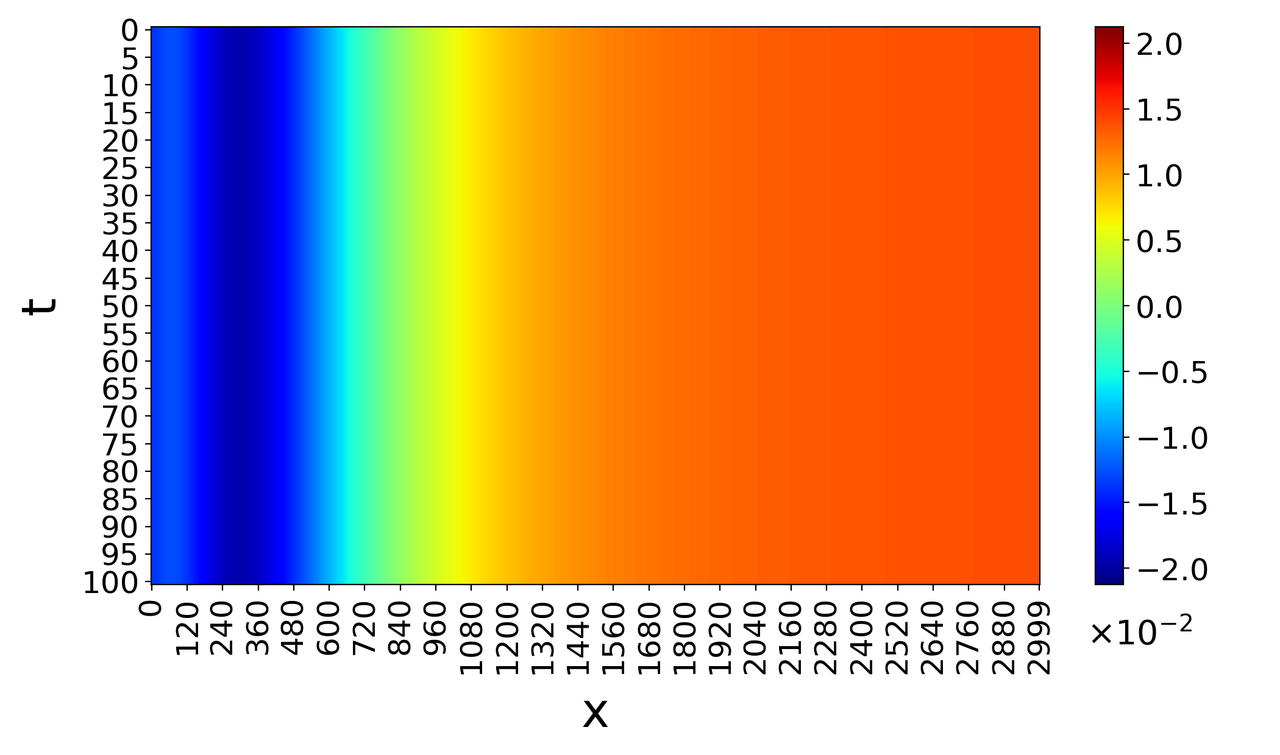}
        \caption{gPINN error distribution}
        \label{fig:loss6}
    \end{subfigure}
    \begin{subfigure}[t]{0.45\textwidth}
        \centering
        \includegraphics[width=\textwidth]{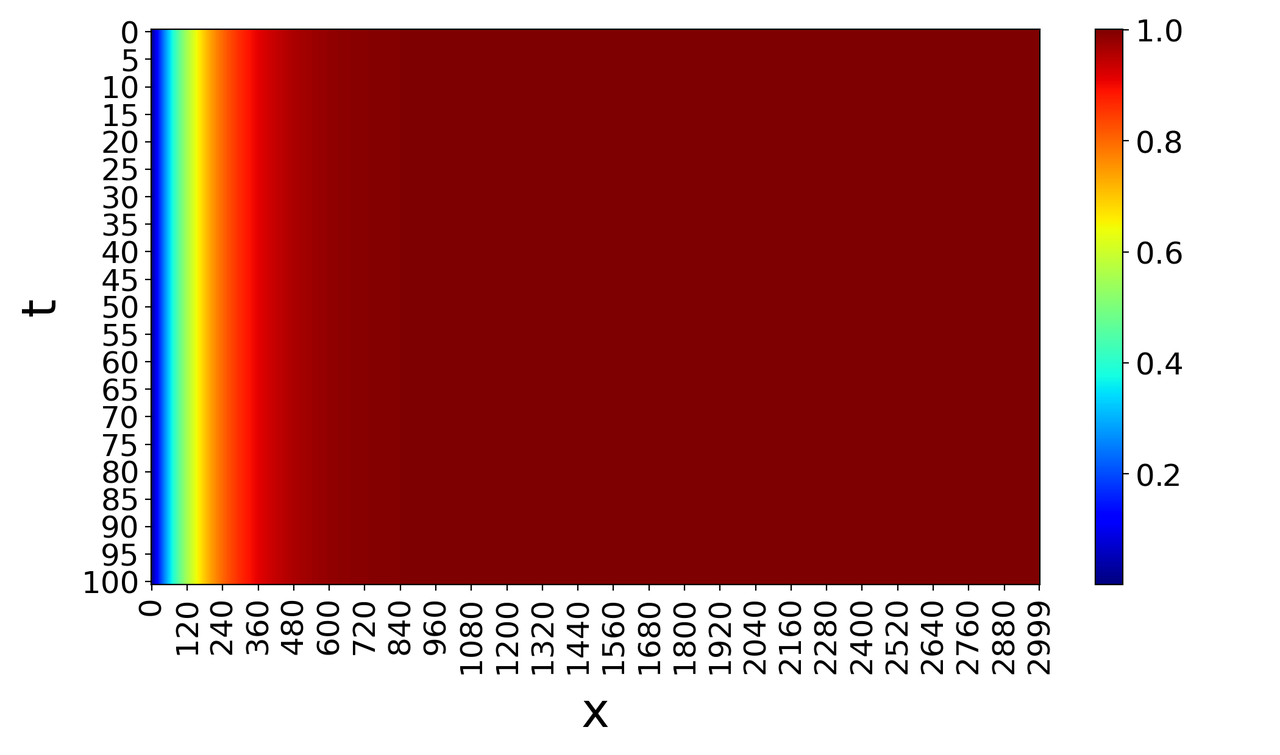}
        \caption{SA-PINN prediction}
        \label{fig:loss7}
    \end{subfigure}
    \hfill
    \begin{subfigure}[t]{0.45\textwidth}
        \centering
        \includegraphics[width=\textwidth]{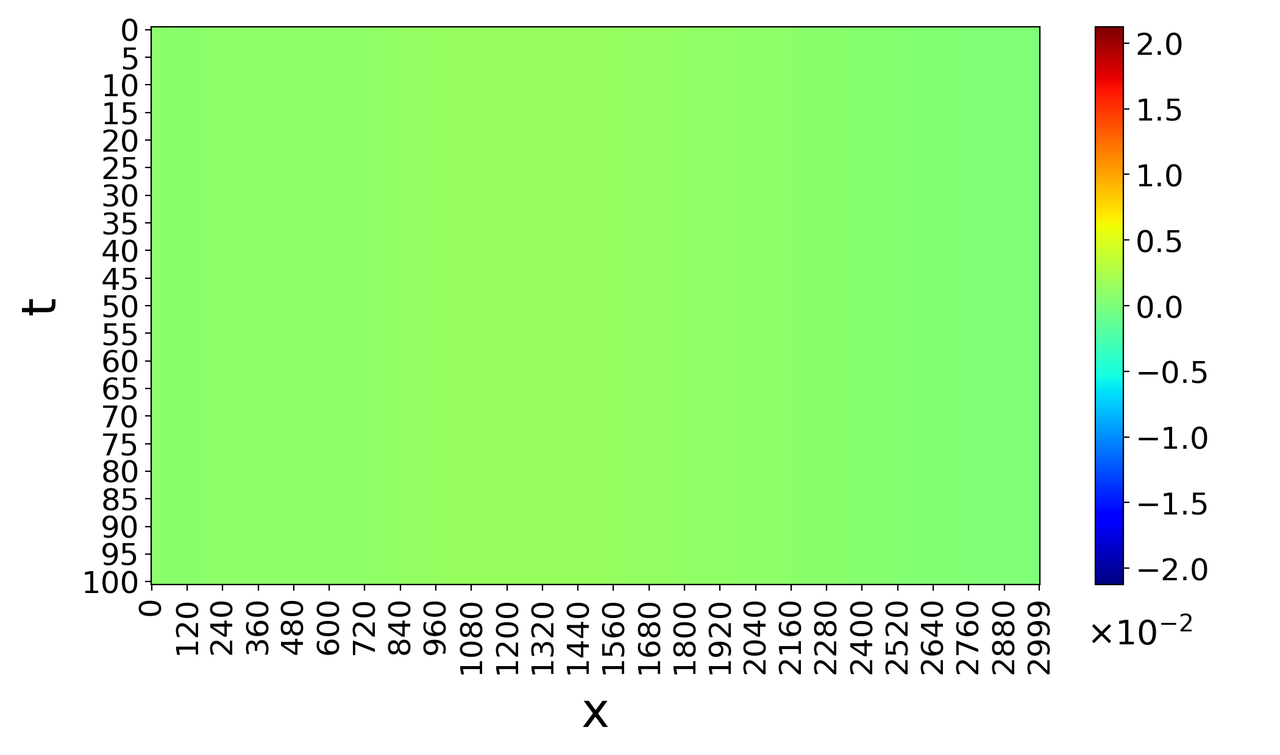}
        \caption{SA-PINN error distribution}
        \label{fig:loss8}
    \end{subfigure}
    \caption{Predicted solutions and error distributions across models for the 1D convection-diffusion equation (1).}
    \label{fig:v1}
\end{figure}

\begin{figure}[h!]
    \centering
    \begin{subfigure}[t]{0.45\textwidth}
        \centering
        \includegraphics[width=\textwidth]{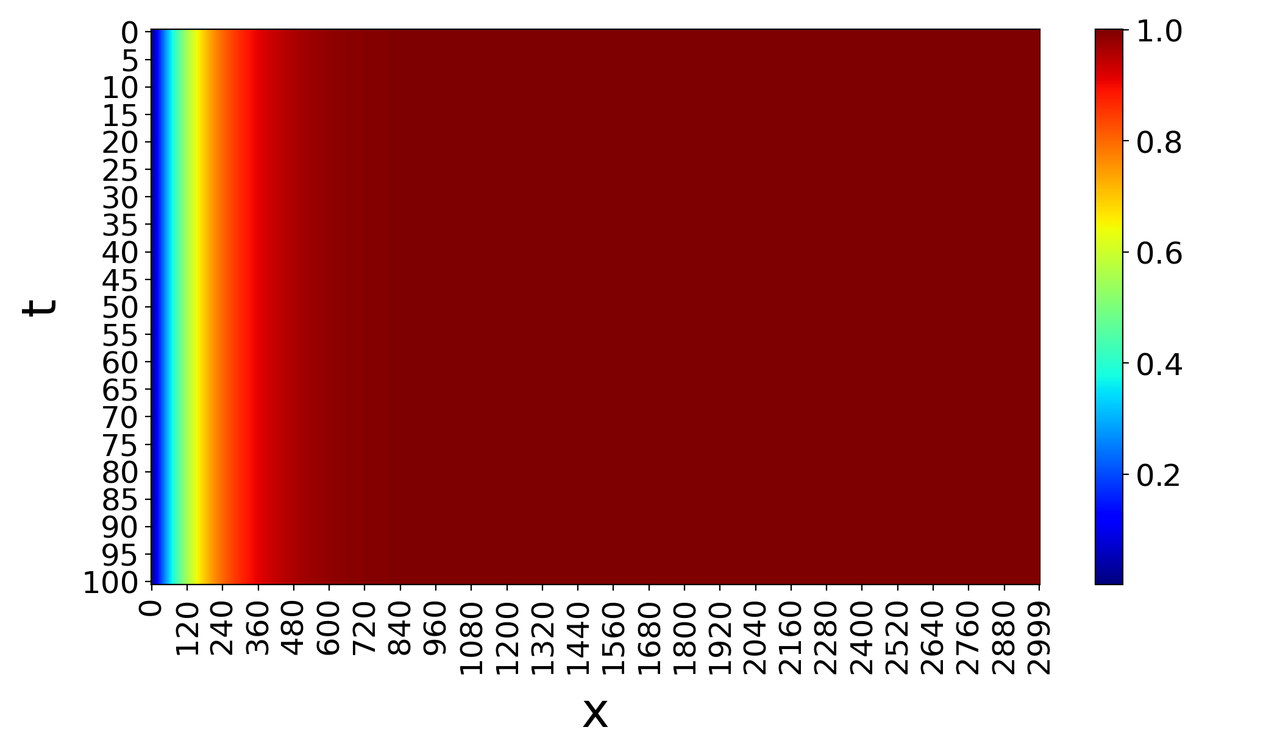}
        \caption{PINN+RAR prediction}
        \label{fig:loss9}
    \end{subfigure}
      \hfill
    \begin{subfigure}[t]{0.45\textwidth}
        \centering
        \includegraphics[width=\textwidth]{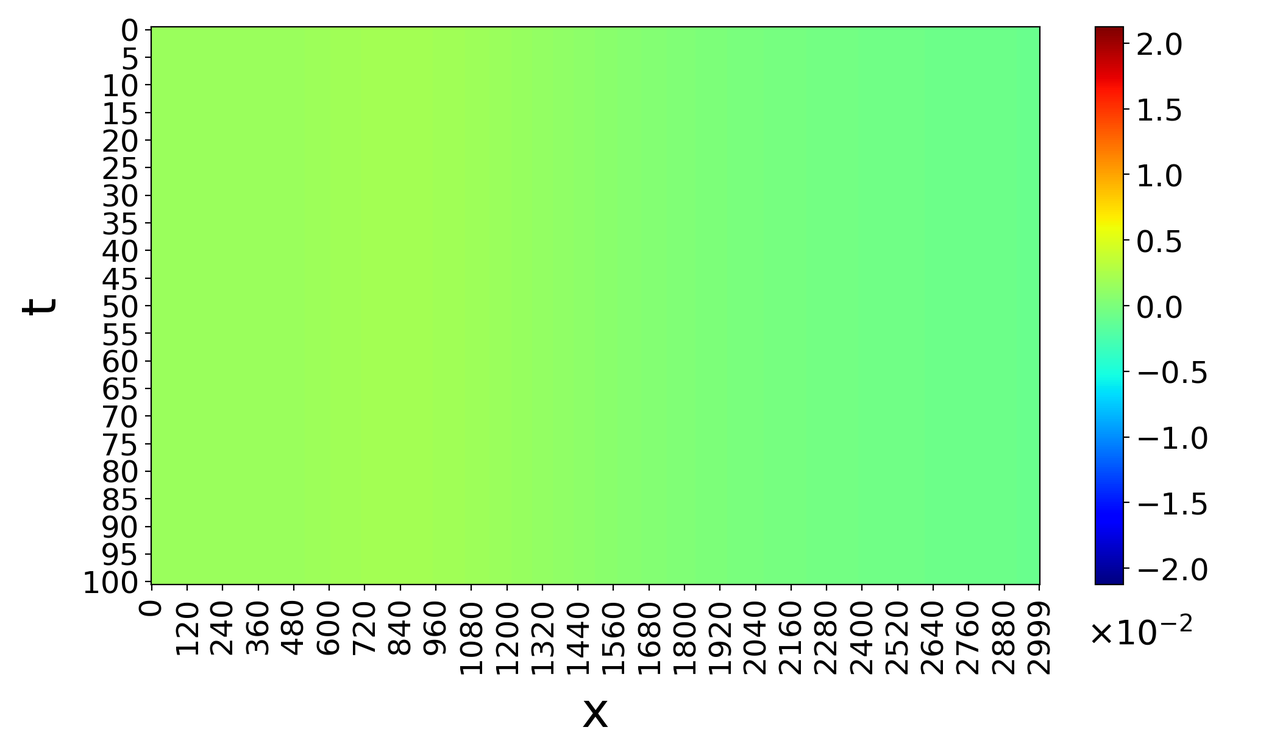}
        \caption{PINN+RAR error distribution}
        \label{fig:loss10}
    \end{subfigure}

    \begin{subfigure}[t]{0.45\textwidth}
        \centering
        \includegraphics[width=\textwidth]{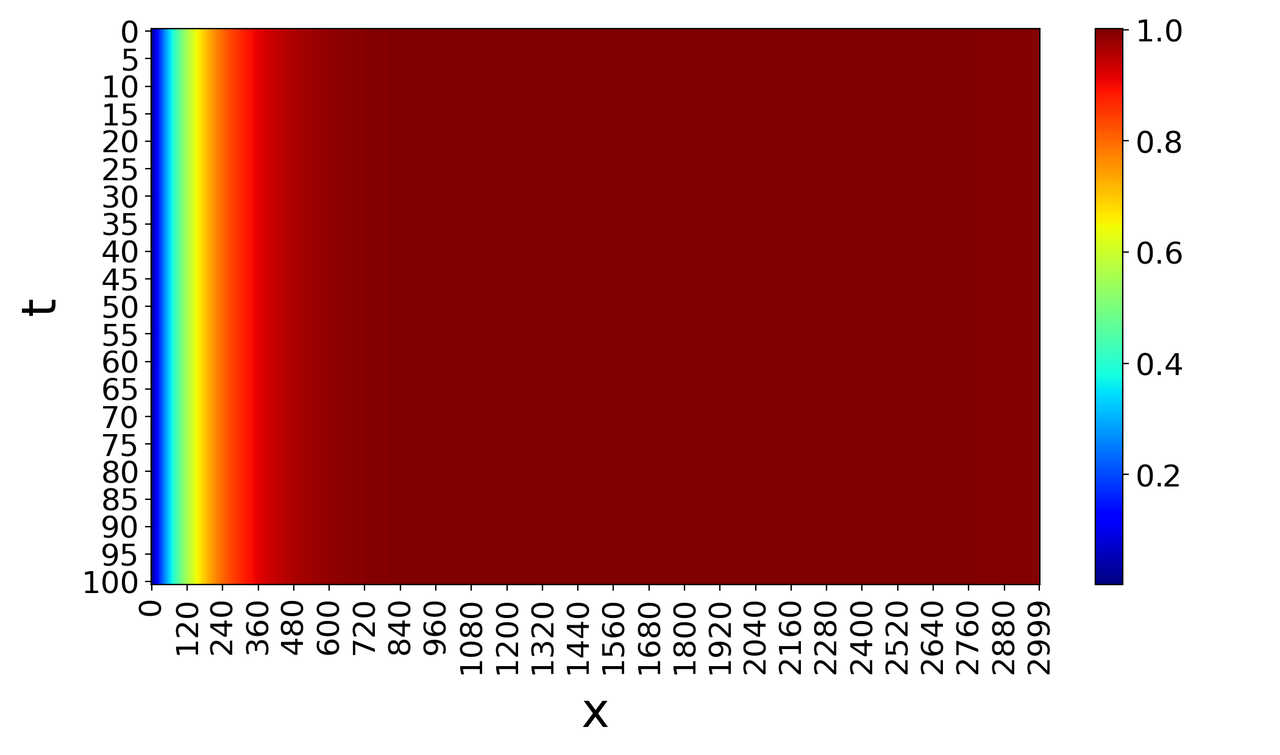}
        \caption{GC-PINN (Radial Mapping) prediction}
        \label{fig:loss11}
    \end{subfigure}
    \hfill
    \begin{subfigure}[t]{0.45\textwidth}
        \centering
        \includegraphics[width=\textwidth]{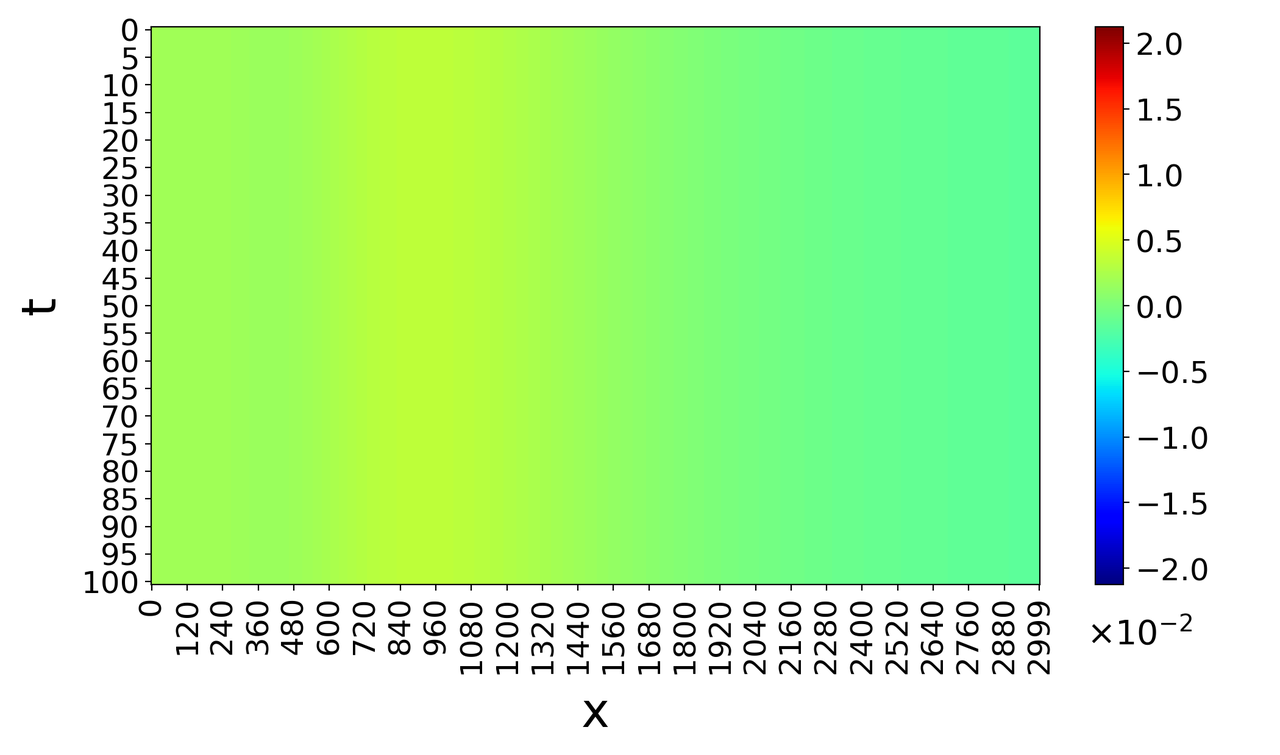}
        \caption{GC-PINN (Radial Mapping) error distribution}
        \label{fig:loss12}
    \end{subfigure}
    \begin{subfigure}[t]{0.45\textwidth}
        \centering
        \includegraphics[width=\textwidth]{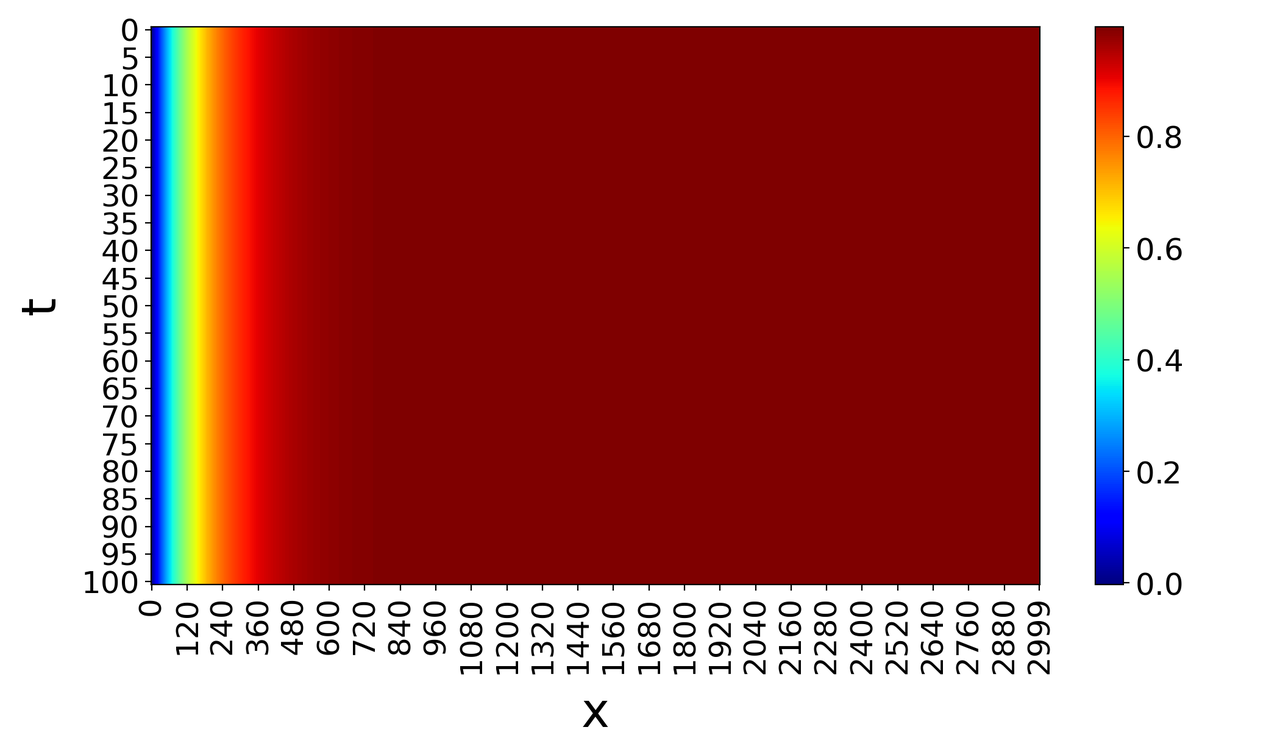}
        \caption{GC-PINN (Radial Mapping) prediction}
        \label{fig:loss13}
    \end{subfigure}
    \hfill
    \begin{subfigure}[t]{0.45\textwidth}
        \centering
        \includegraphics[width=\textwidth]{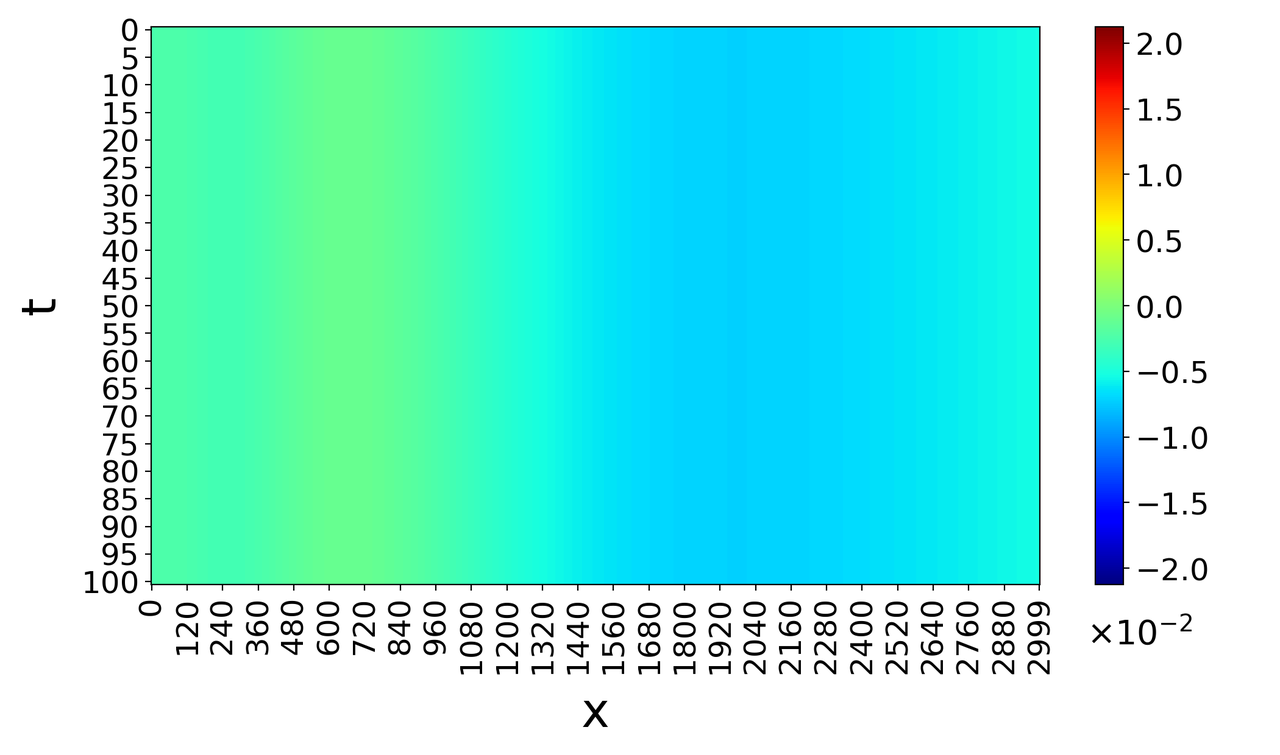}
        \caption{GC-PINN (Radial Mapping) error distribution}
        \label{fig:loss14}
    \end{subfigure}
  
    \begin{subfigure}[t]{0.45\textwidth}
        \centering
        \includegraphics[width=\textwidth]{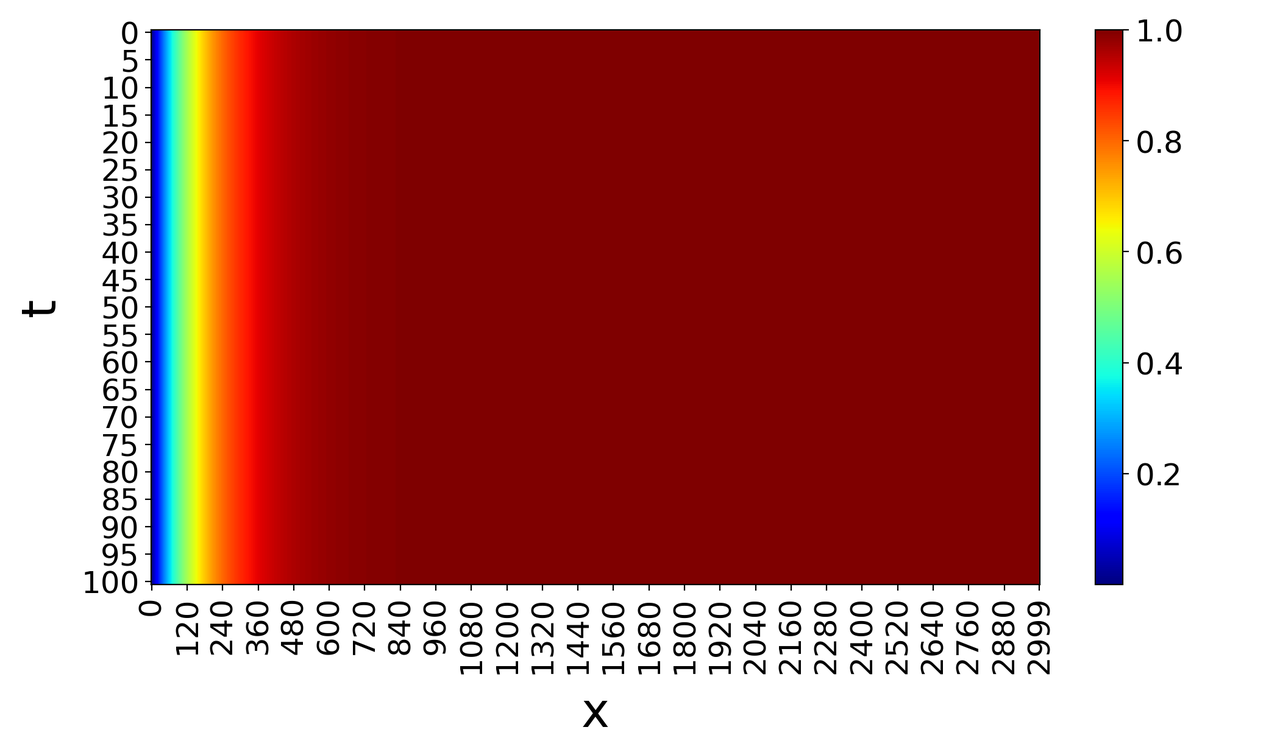}
        \caption{GC-PINN (Local Stretching) prediction}
        \label{fig:loss15}
    \end{subfigure}
      \hfill
    \begin{subfigure}[t]{0.45\textwidth}
        \centering
        \includegraphics[width=\textwidth]{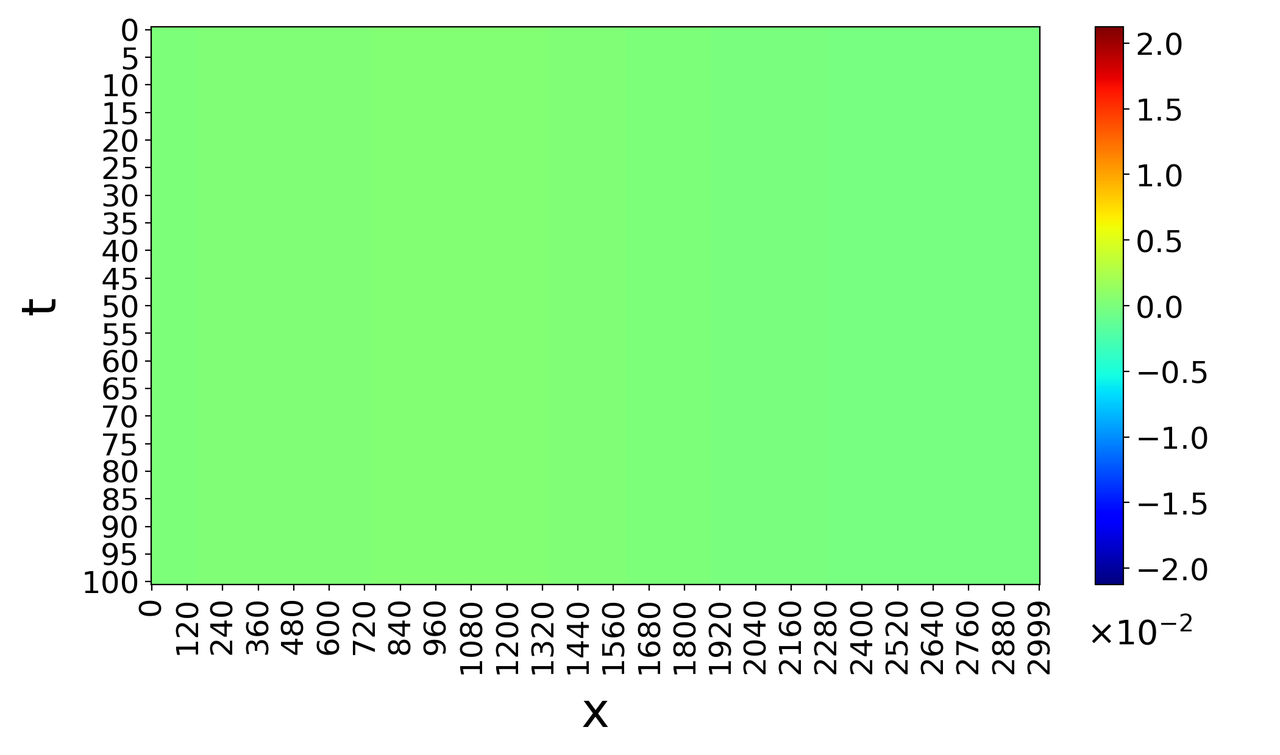}
        \caption{GC-PINN (Local Stretching) error distribution}
        \label{fig:loss16}
    \end{subfigure}

    \caption{Predicted solutions and error distributions across models for the 1D convection-diffusion equation (2).}
    \label{fig:v2}
\end{figure}

\begin{figure}[h!]
    \centering
    \begin{subfigure}[t]{0.45\textwidth}
        \centering
        \includegraphics[width=\textwidth]{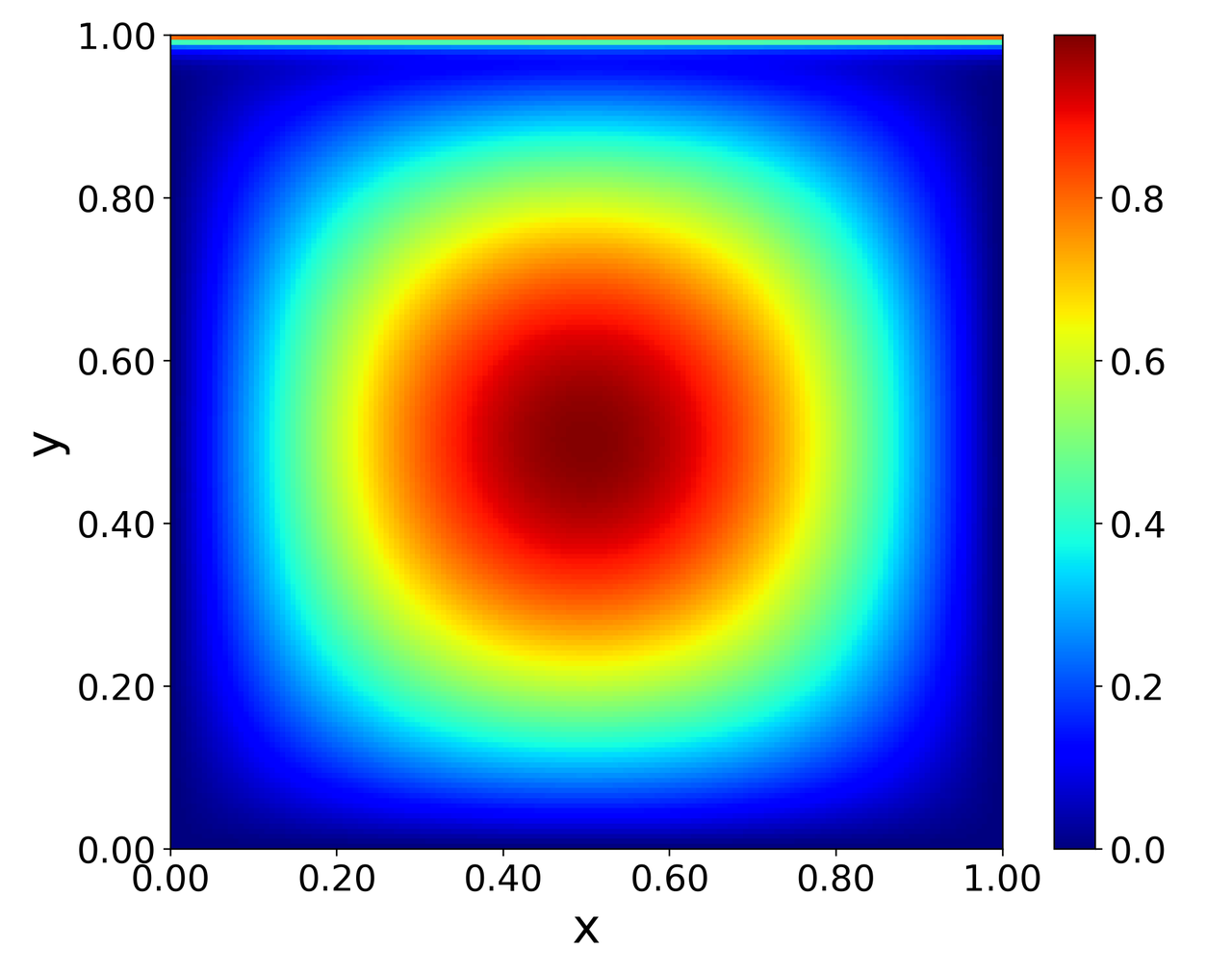}
        \caption{PINN prediction}
    \end{subfigure}
    \hfill
    \begin{subfigure}[t]{0.45\textwidth}
        \centering
        \includegraphics[width=\textwidth]{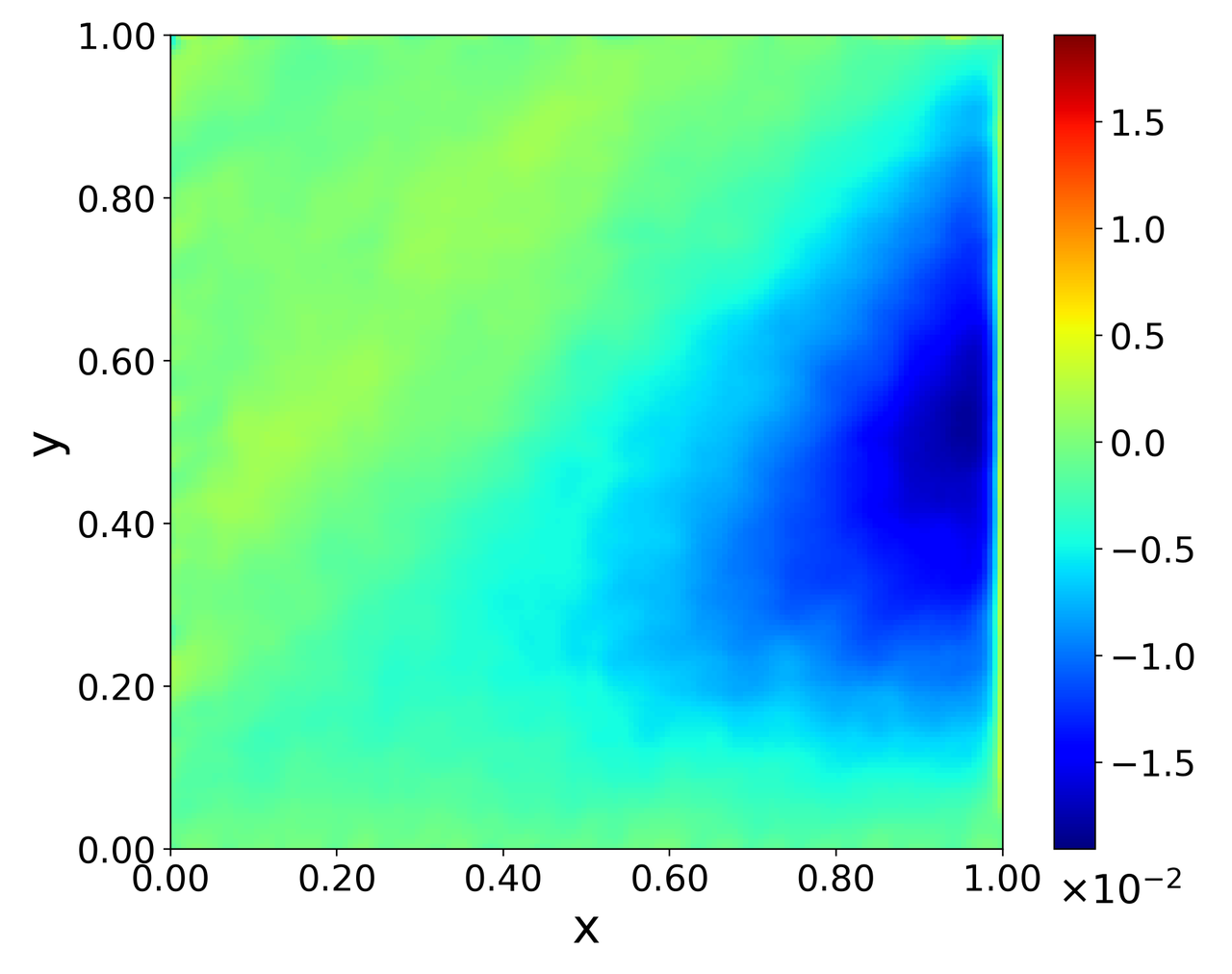}
        \caption{PINN error distribution}
    \end{subfigure}
  
    \begin{subfigure}[t]{0.45\textwidth}
        \centering
        \includegraphics[width=\textwidth]{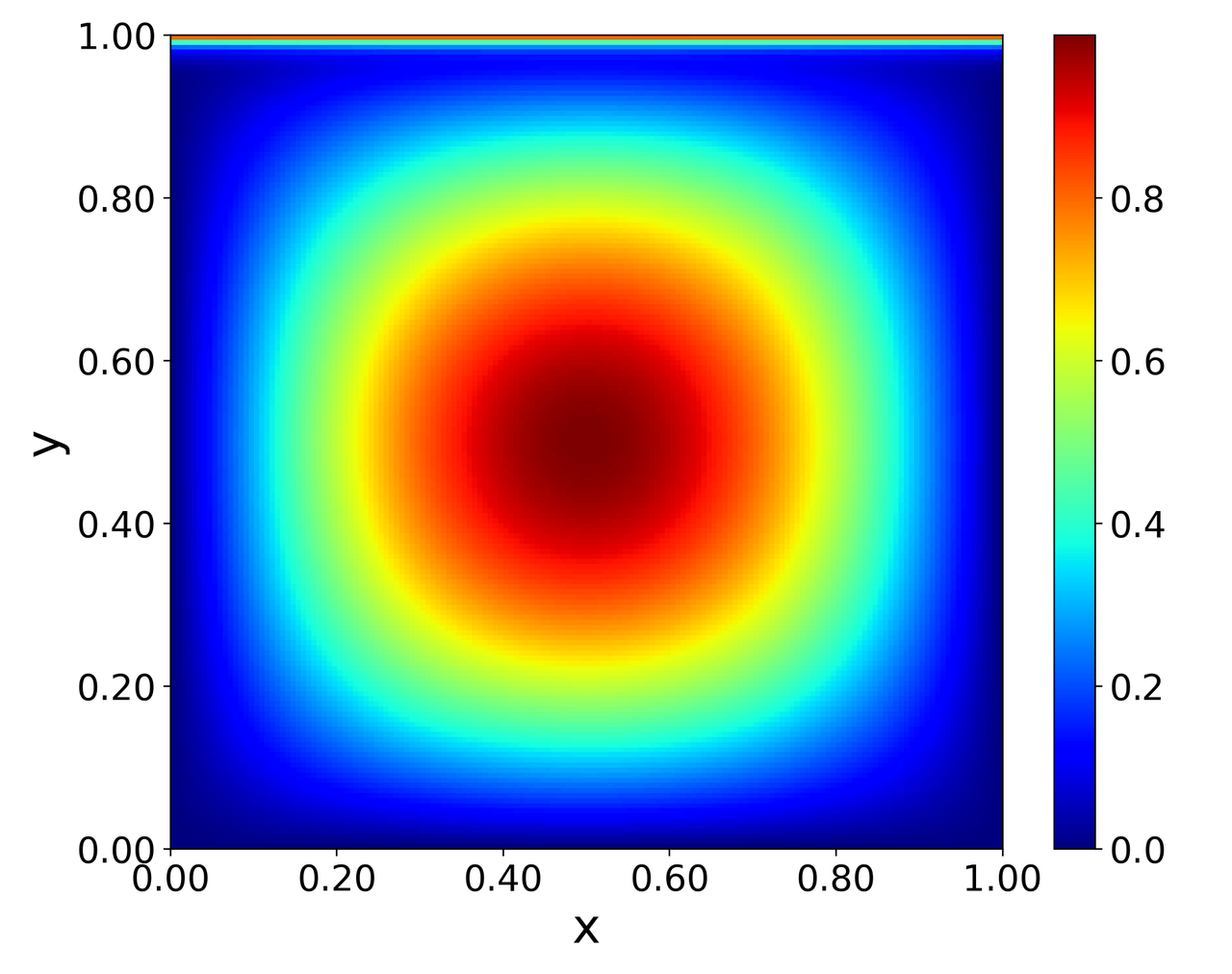}
        \caption{FF-PINN prediction}
    \end{subfigure}
      \hfill
    \begin{subfigure}[t]{0.45\textwidth}
        \centering
        \includegraphics[width=\textwidth]{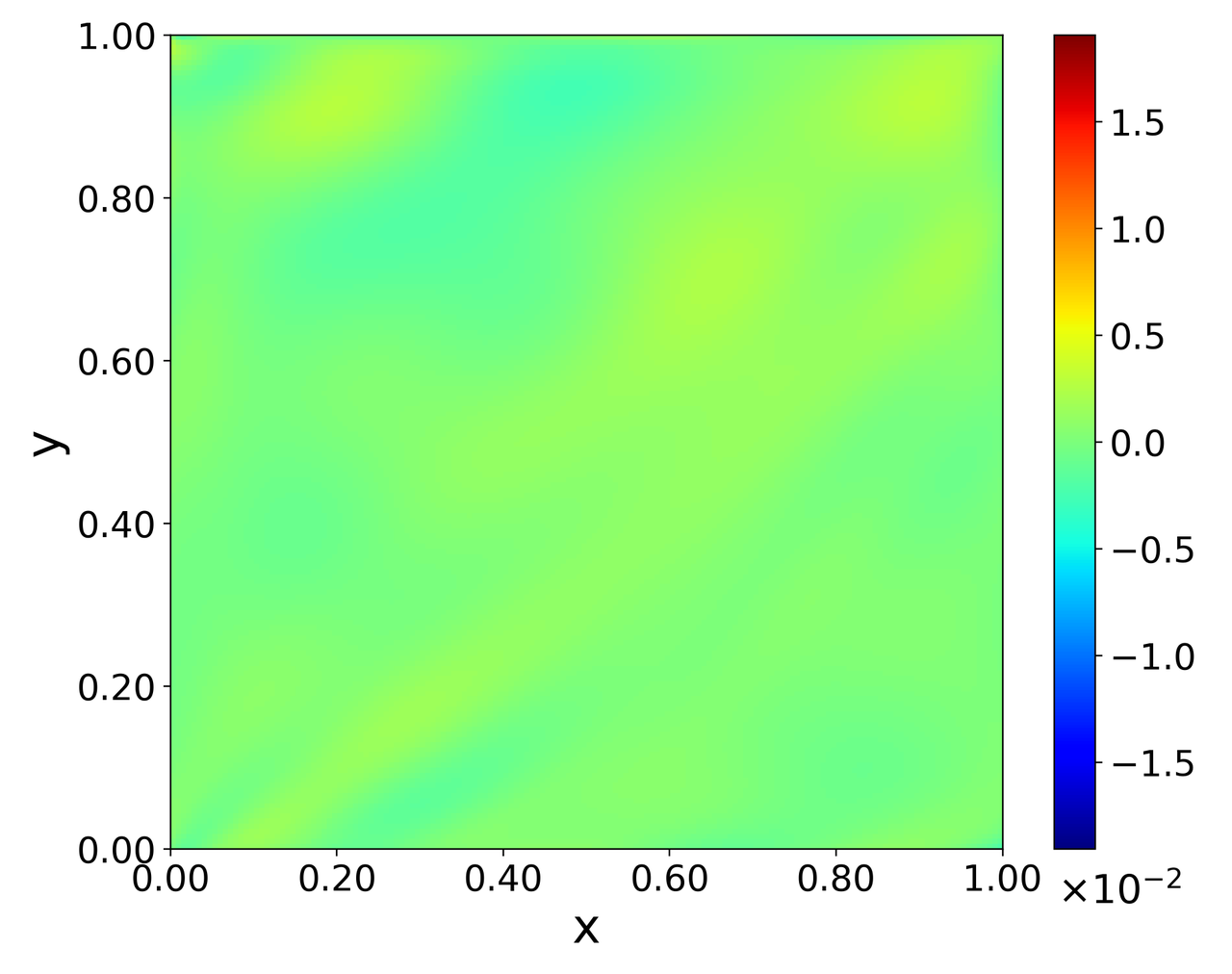}
        \caption{FF-PINN error distribution}
    \end{subfigure}

    \begin{subfigure}[t]{0.45\textwidth}
        \centering
        \includegraphics[width=\textwidth]{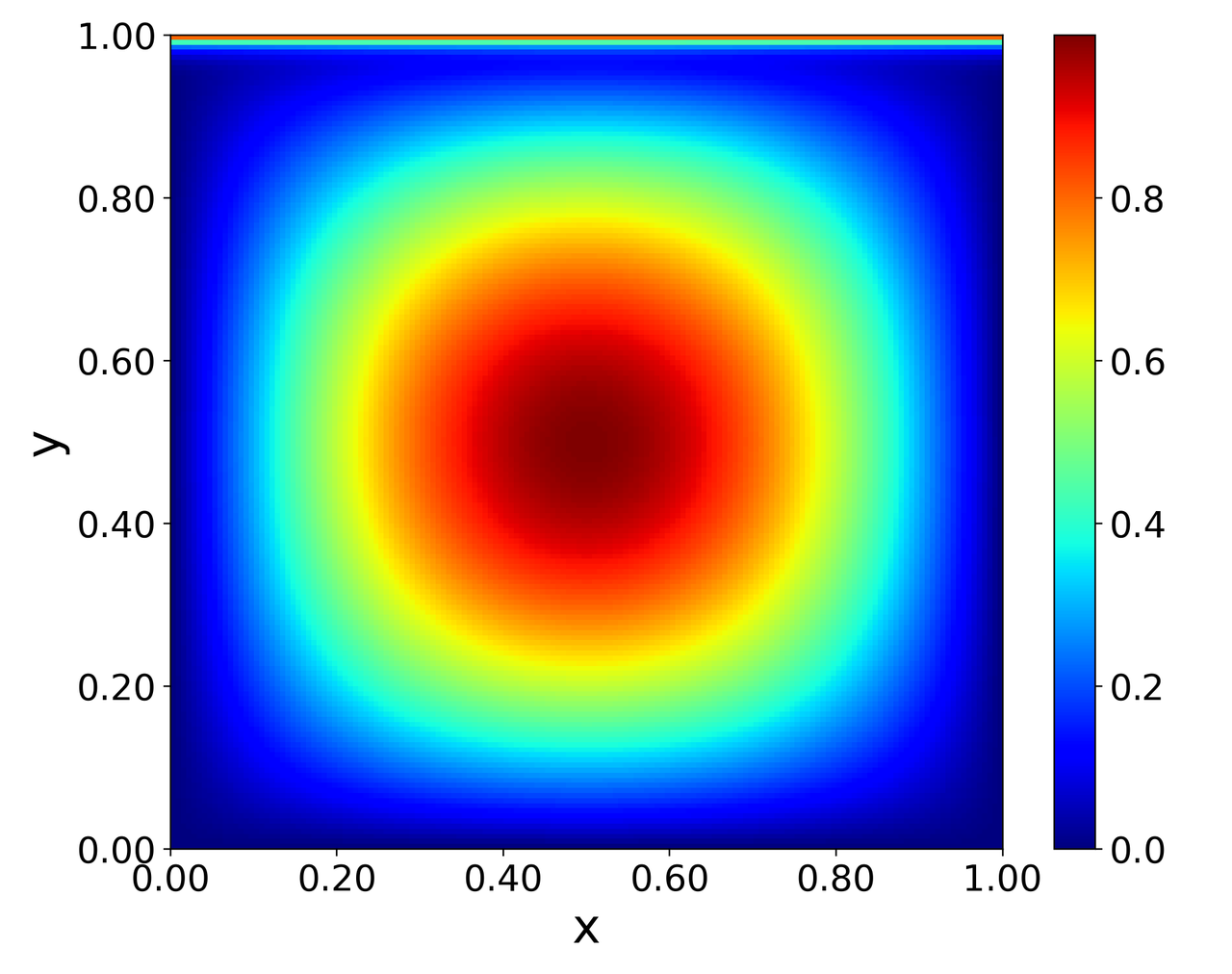}
        \caption{gPINN prediction}
    \end{subfigure}
    \hfill
    \begin{subfigure}[t]{0.45\textwidth}
        \centering
        \includegraphics[width=\textwidth]{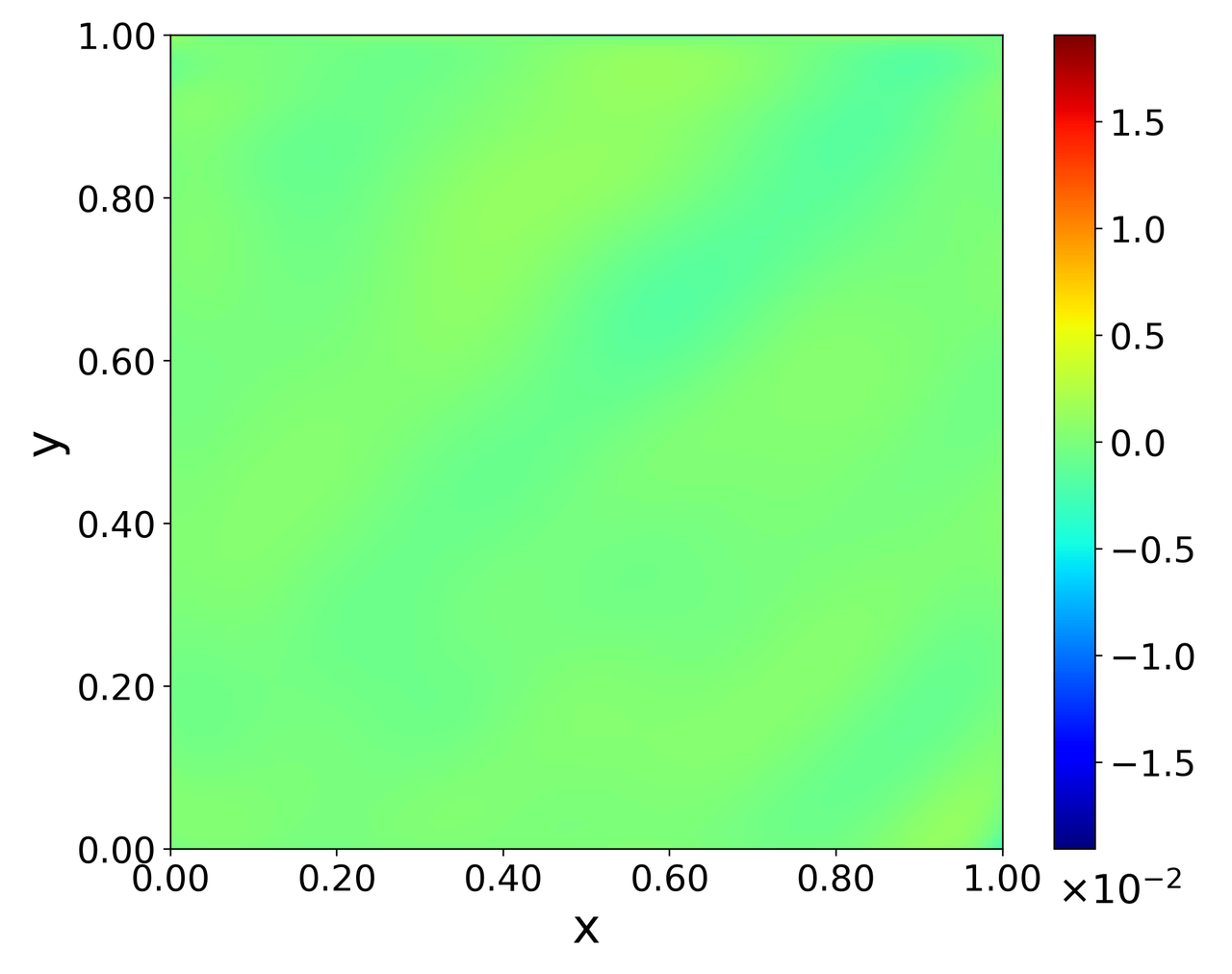}
        \caption{gPINN error distribution}
    \end{subfigure}

    \caption{Predicted solutions and error distributions across models for the 2D convection-diffusion equation (1).}
    \label{fig:v3}
\end{figure}

\begin{figure}[h!]
    \centering
        \begin{subfigure}[t]{0.45\textwidth}
        \centering
        \includegraphics[width=\textwidth]{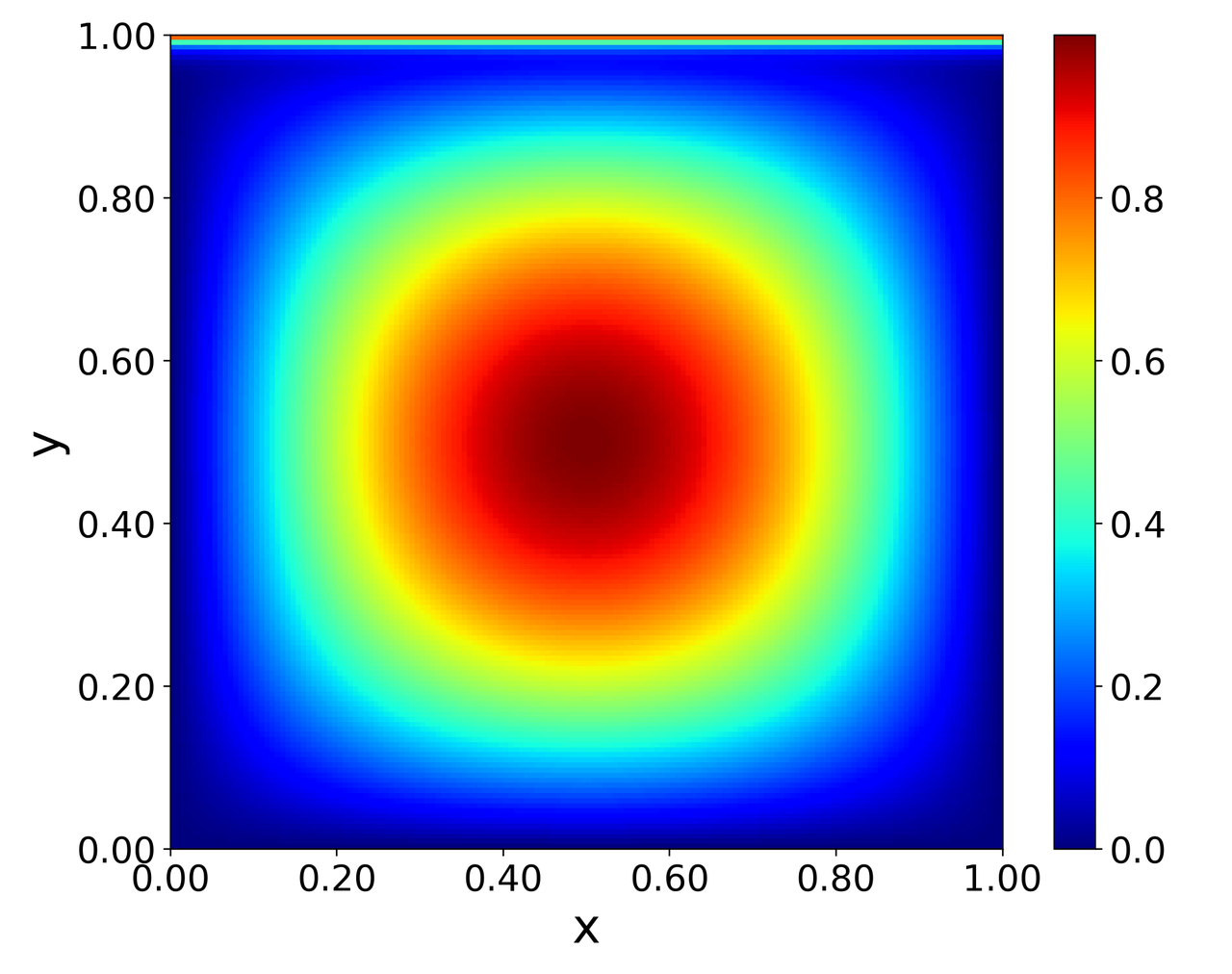}
        \caption{SA-PINN prediction}
    \end{subfigure}
    \hfill
    \begin{subfigure}[t]{0.45\textwidth}
        \centering
        \includegraphics[width=\textwidth]{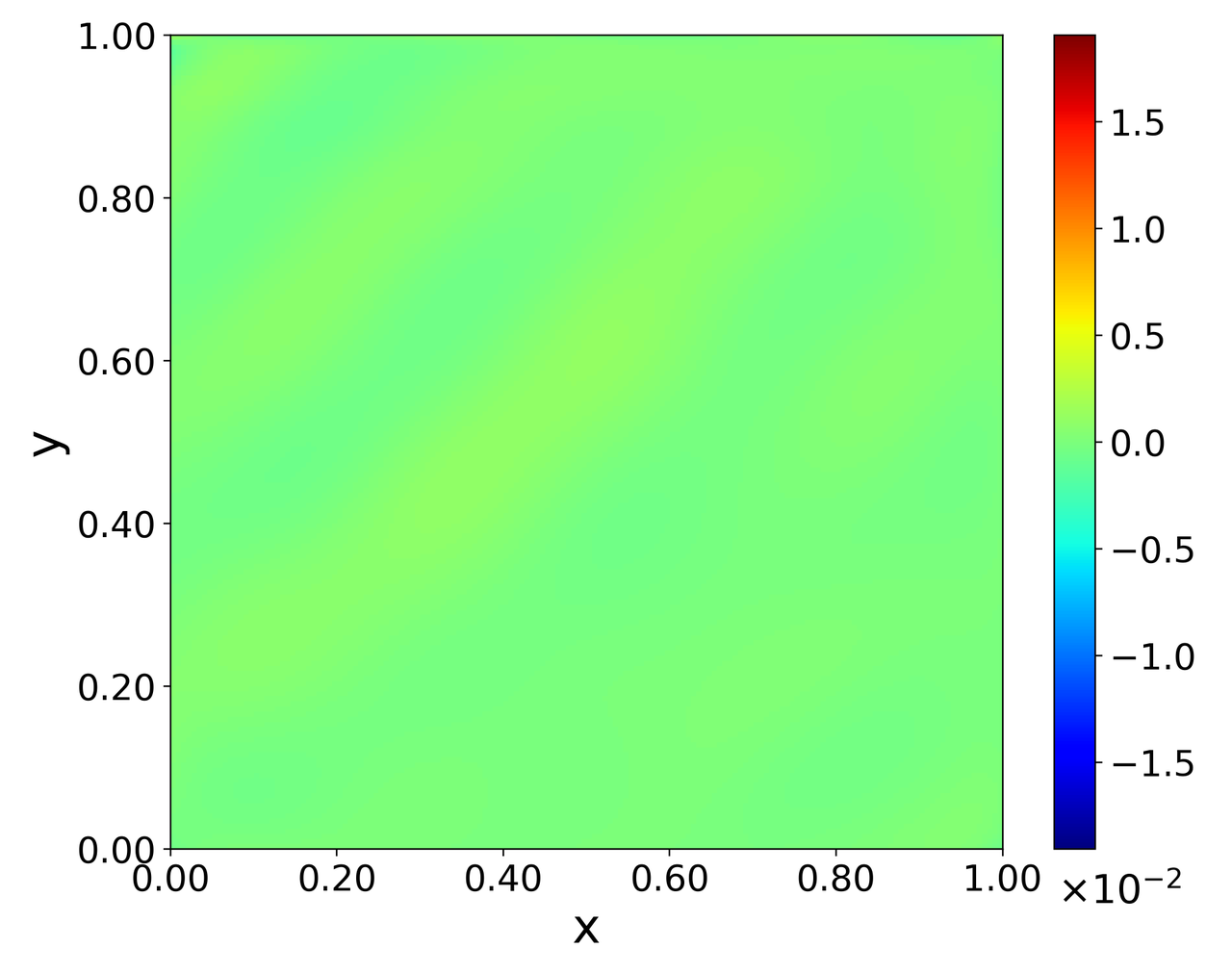}
        \caption{SA-PINN error distribution}
    \end{subfigure}
    \begin{subfigure}[t]{0.45\textwidth}
        \centering
        \includegraphics[width=\textwidth]{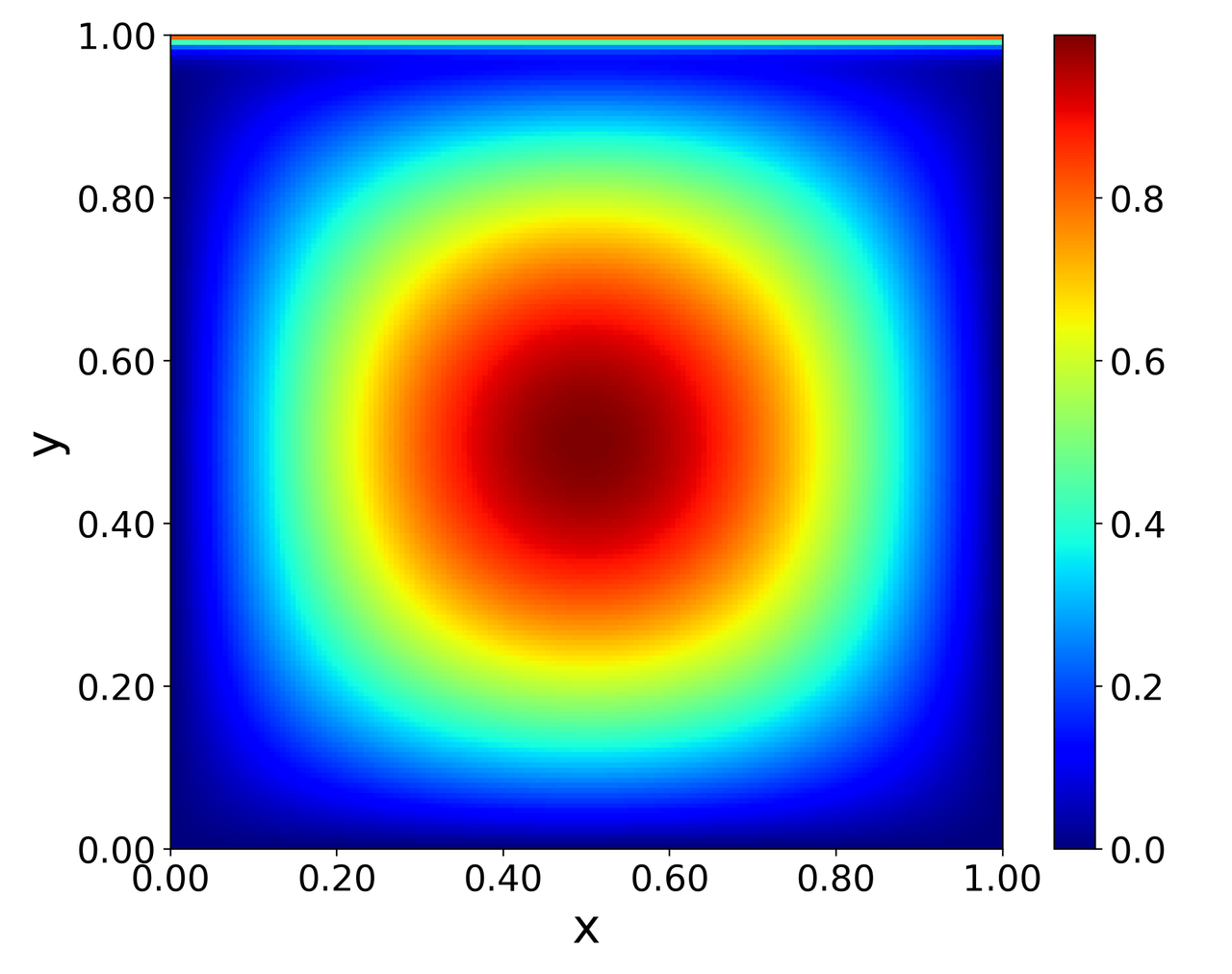}
        \caption{PINN+RAR prediction}
    \end{subfigure}
      \hfill
    \begin{subfigure}[t]{0.45\textwidth}
        \centering
        \includegraphics[width=\textwidth]{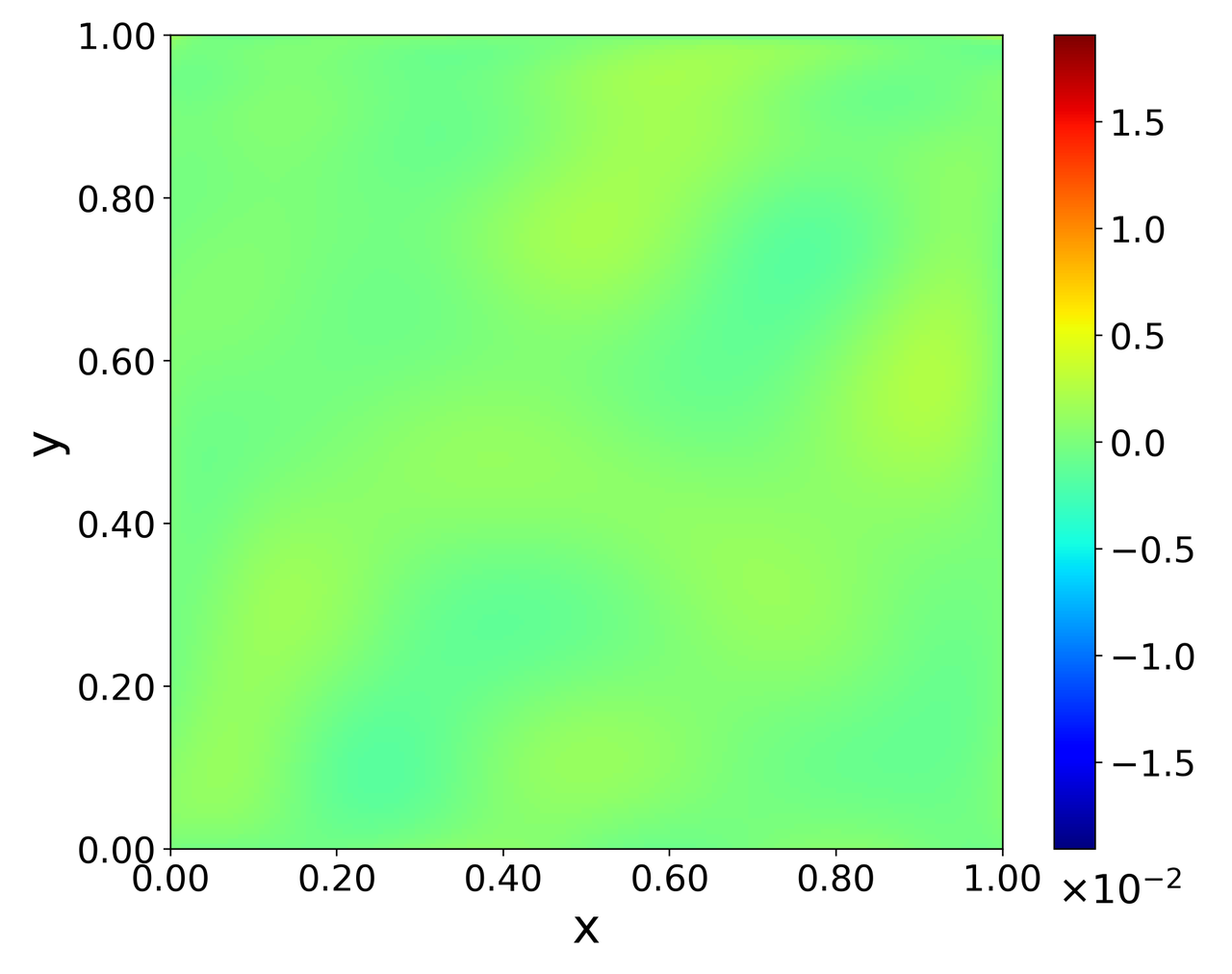}
        \caption{PINN+RAR error distribution}
    \end{subfigure}

    \begin{subfigure}[t]{0.45\textwidth}
        \centering
        \includegraphics[width=\textwidth]{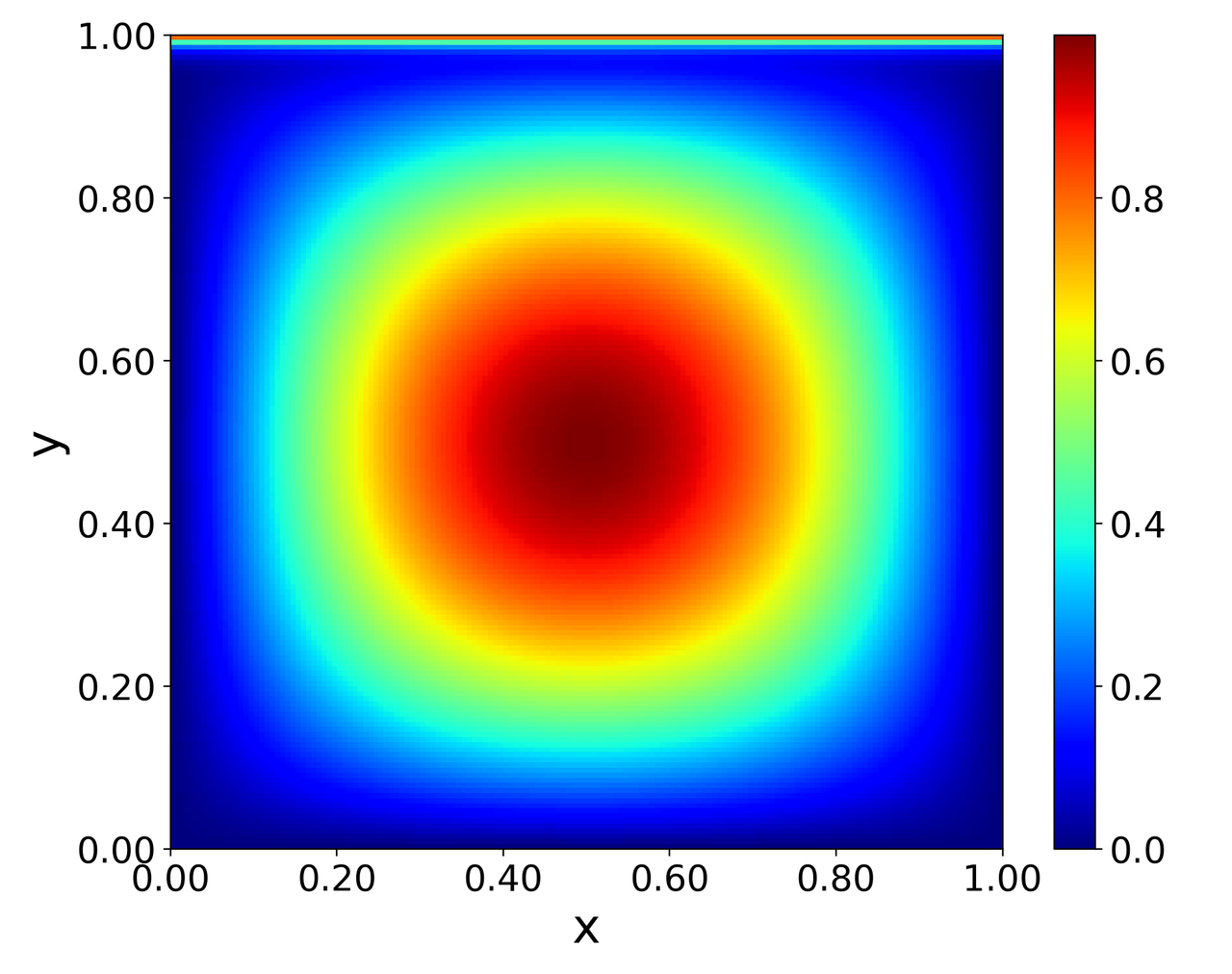}
        \caption{GC-PINN (Radial Mapping) prediction}
    \end{subfigure}
    \hfill
    \begin{subfigure}[t]{0.45\textwidth}
        \centering
        \includegraphics[width=\textwidth]{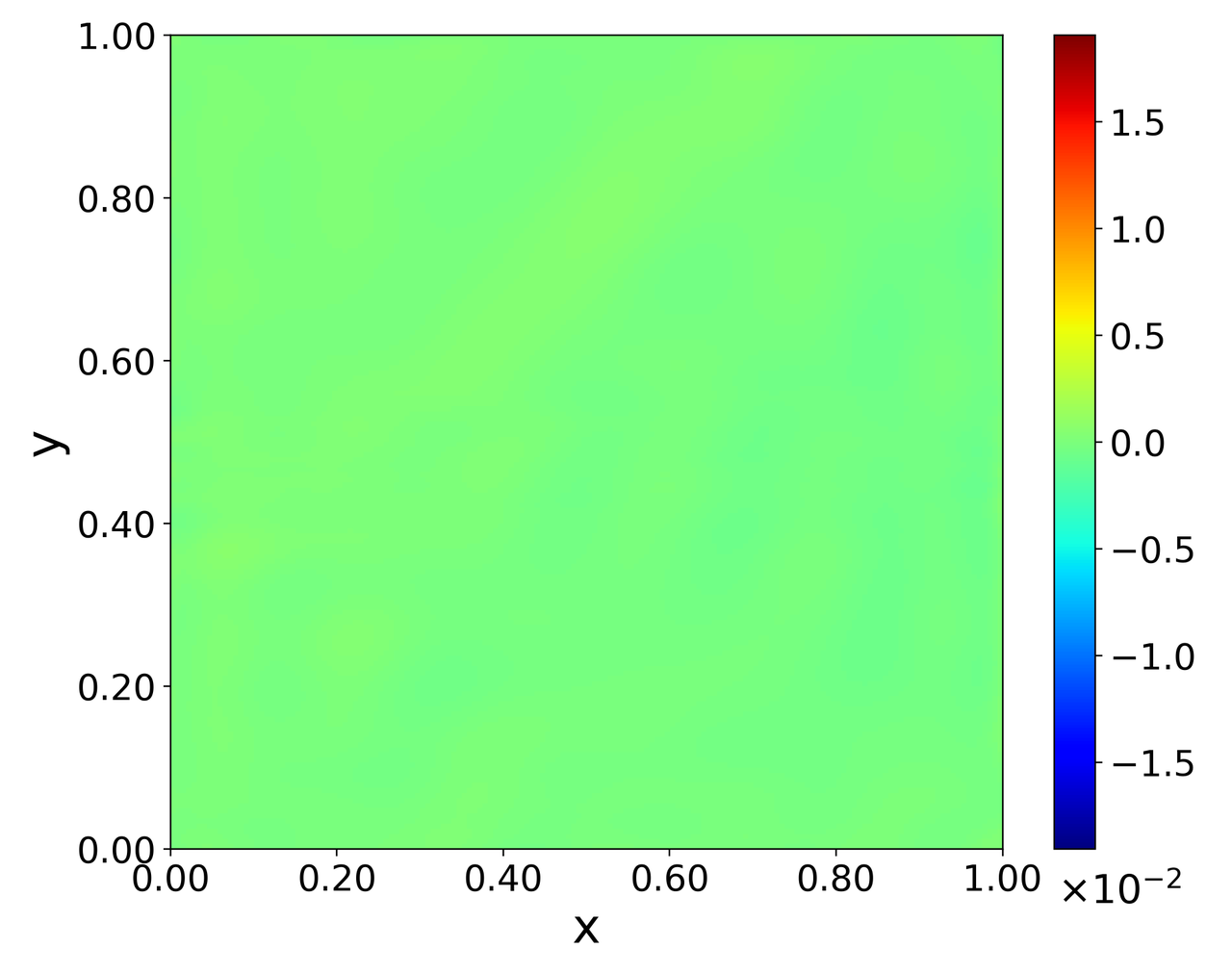}
        \caption{GC-PINN (Radial Mapping) error distribution}
    \end{subfigure}

    \caption{Predicted solutions and error distributions across models for the 2D convection-diffusion equation (2).}
    \label{fig:v4}
\end{figure}

\begin{figure}[h!]
    \centering
    \begin{subfigure}[t]{0.45\textwidth}
        \centering
        \includegraphics[width=\textwidth]{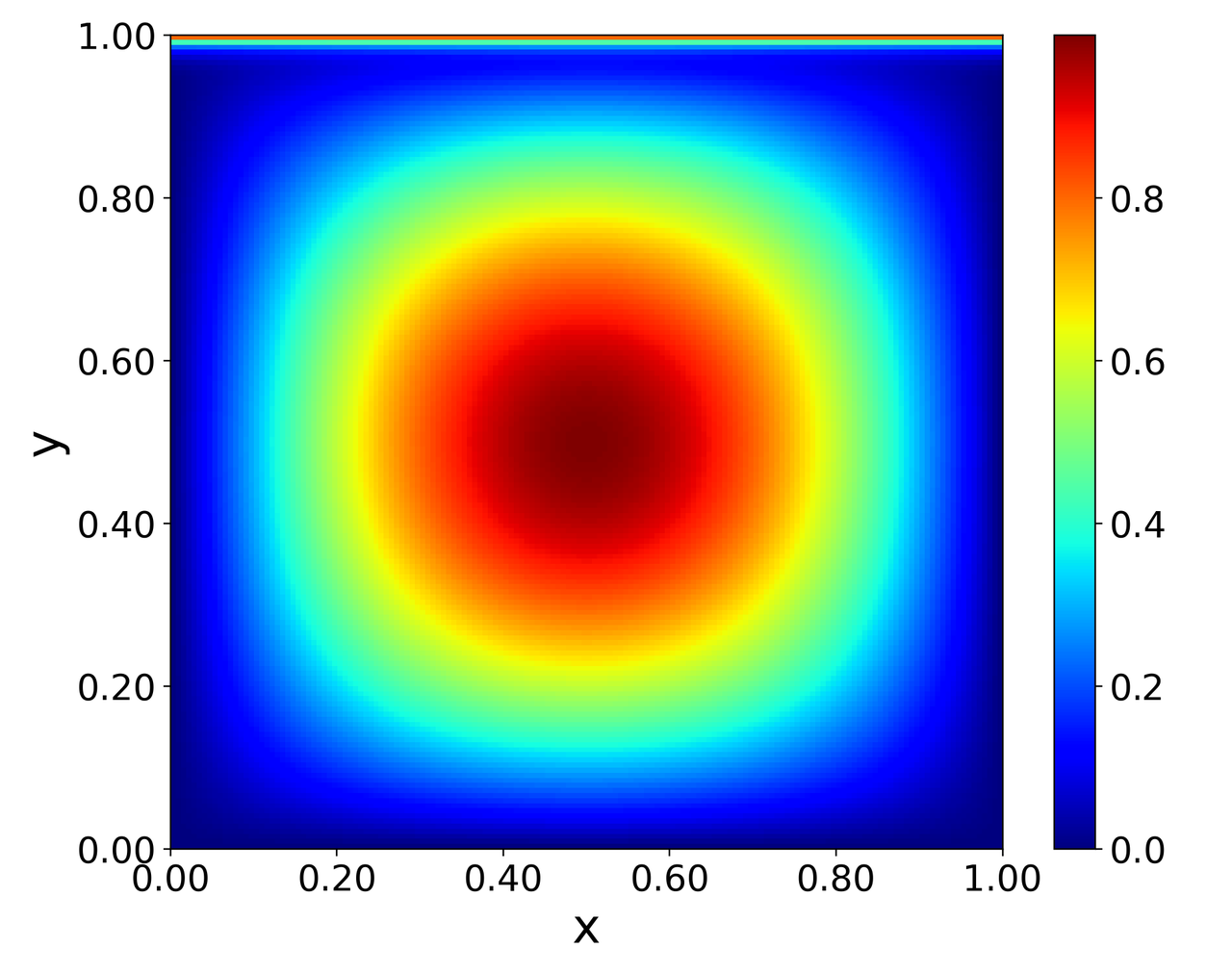}
        \caption{GC-PINN (Radial Mapping) prediction}
    \end{subfigure}
    \hfill
    \begin{subfigure}[t]{0.45\textwidth}
        \centering
        \includegraphics[width=\textwidth]{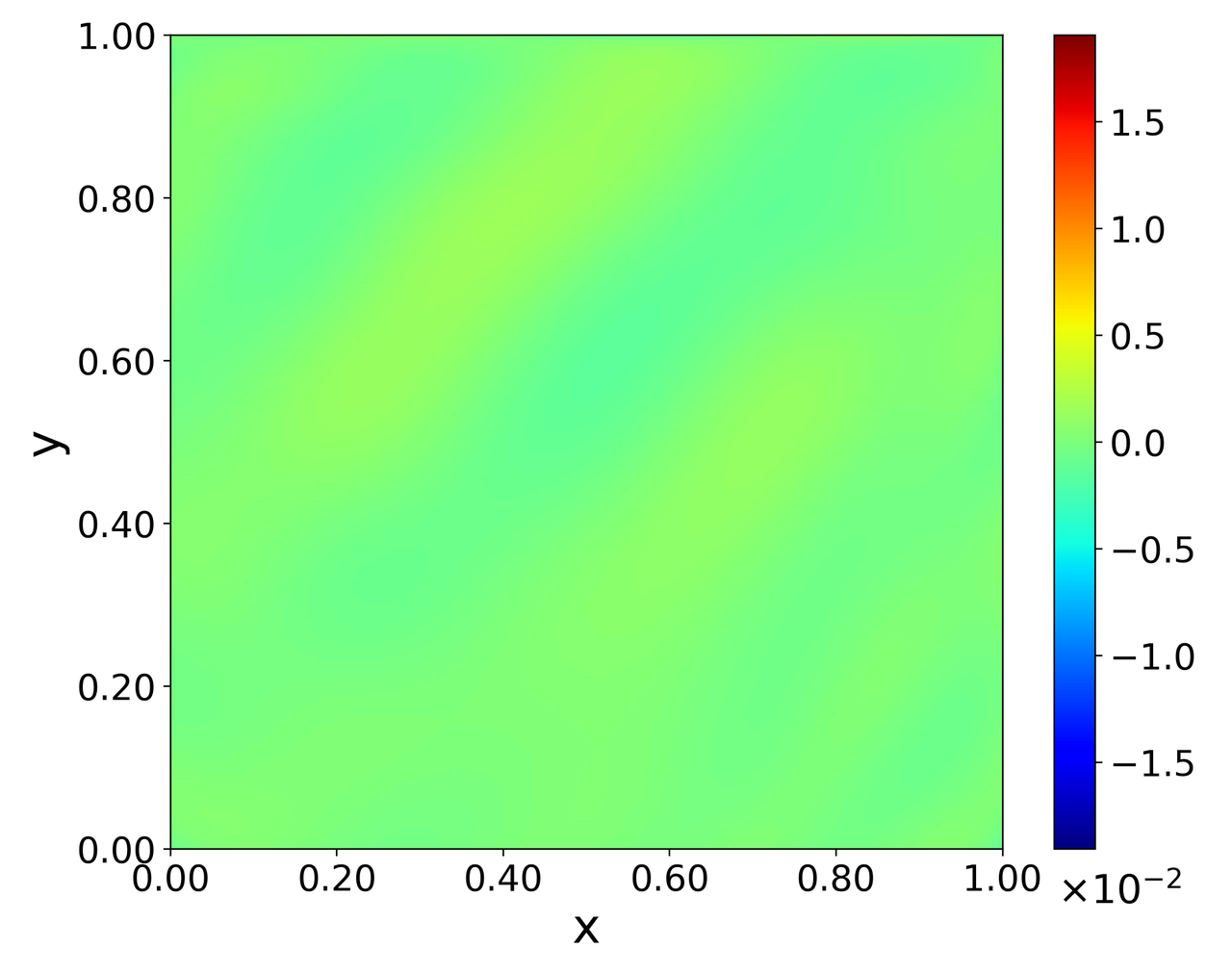}
        \caption{GC-PINN (Radial Mapping) error distribution}
    \end{subfigure}
  
    \begin{subfigure}[t]{0.45\textwidth}
        \centering
        \includegraphics[width=\textwidth]{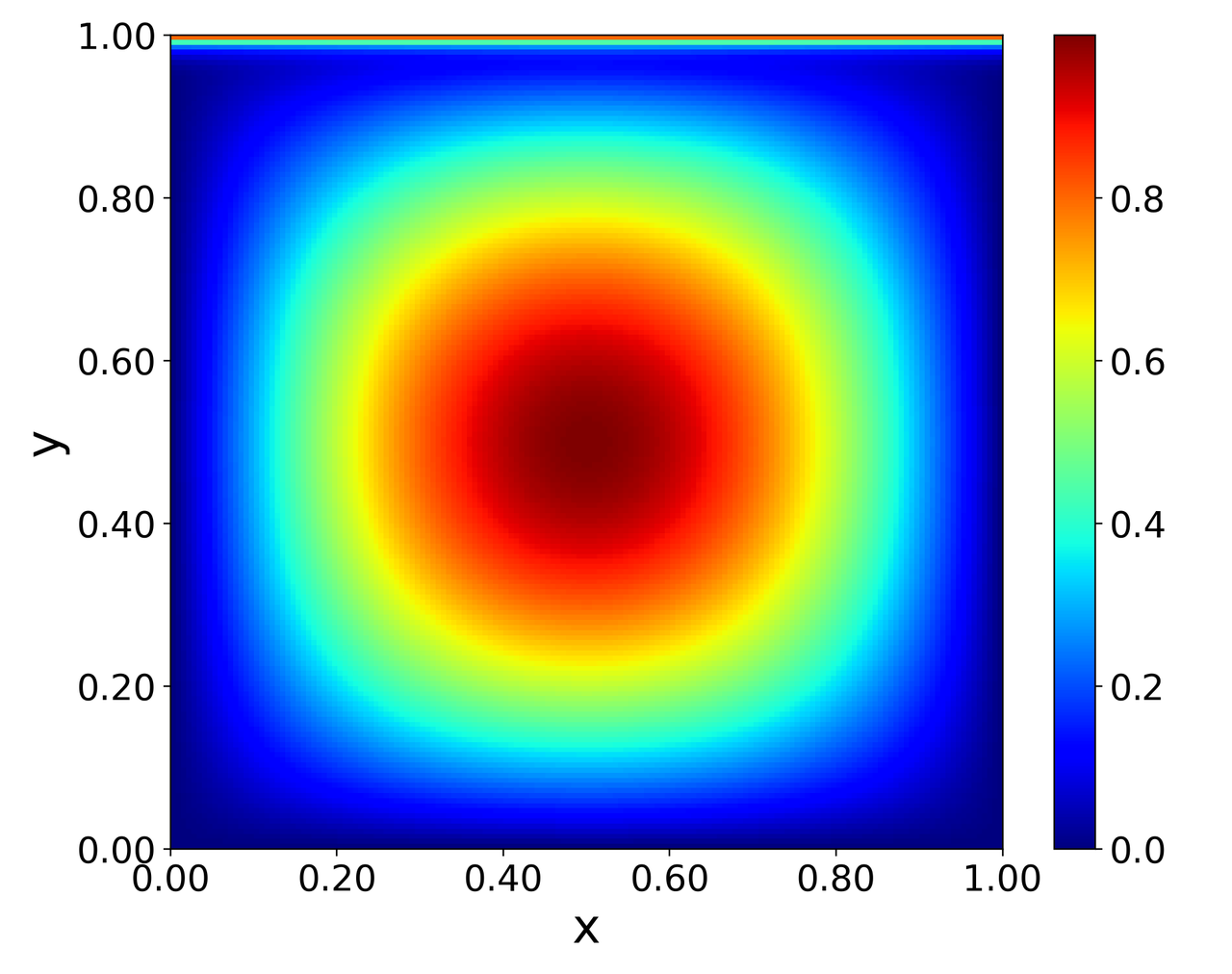}
        \caption{GC-PINN (Local Stretching) prediction}
    \end{subfigure}
      \hfill
    \begin{subfigure}[t]{0.45\textwidth}
        \centering
        \includegraphics[width=\textwidth]{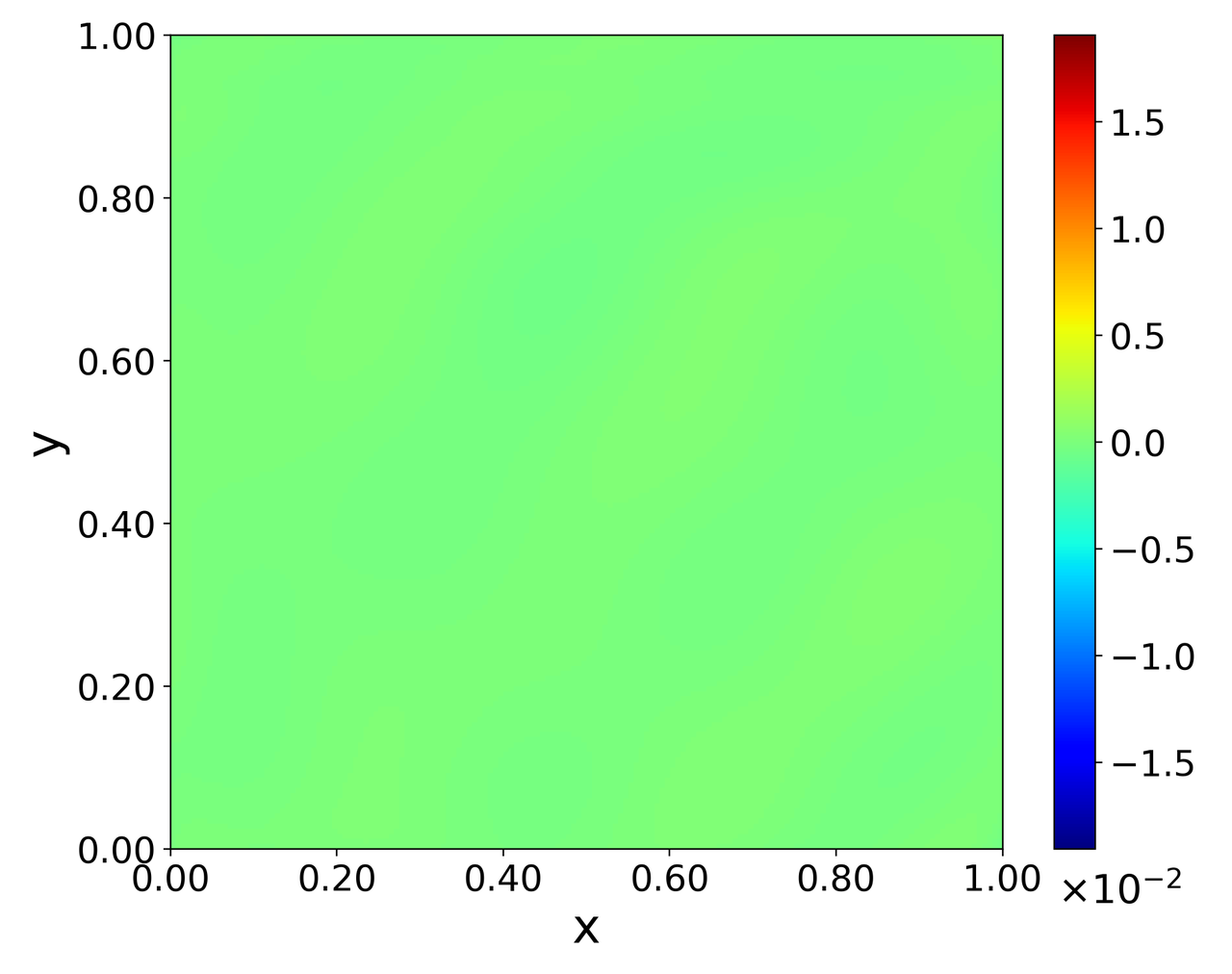}
        \caption{GC-PINN (Local Stretching) error distribution}
    \end{subfigure}

    \caption{Predicted solutions and error distributions across models for the 2D convection-diffusion equation (3).}
    \label{fig:v5}
\end{figure}

\begin{figure}[h!]

\begin{subfigure}[t]{\textwidth}
    \centering
    \includegraphics[width=0.45\textwidth]{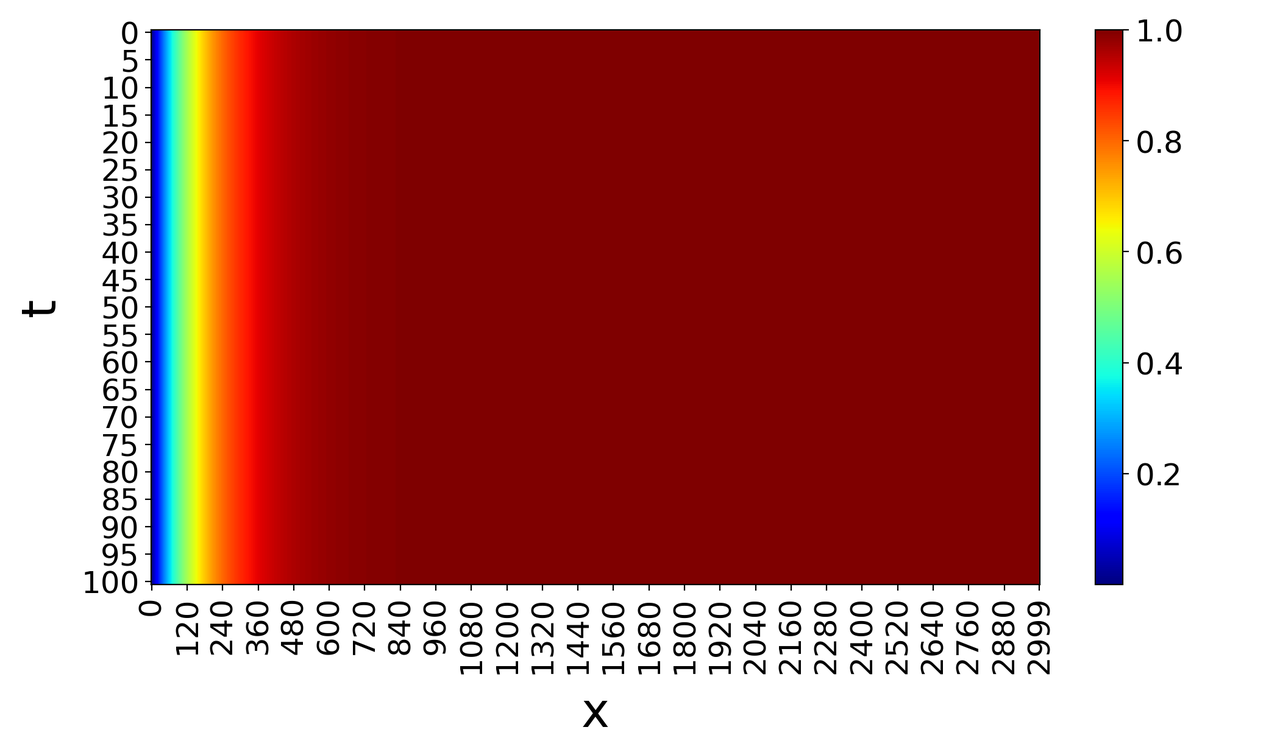}
    \caption{1D convection-diffusion equation}
\end{subfigure}
\begin{subfigure}[t]{\textwidth}
    \centering
    \includegraphics[width=0.45\textwidth]{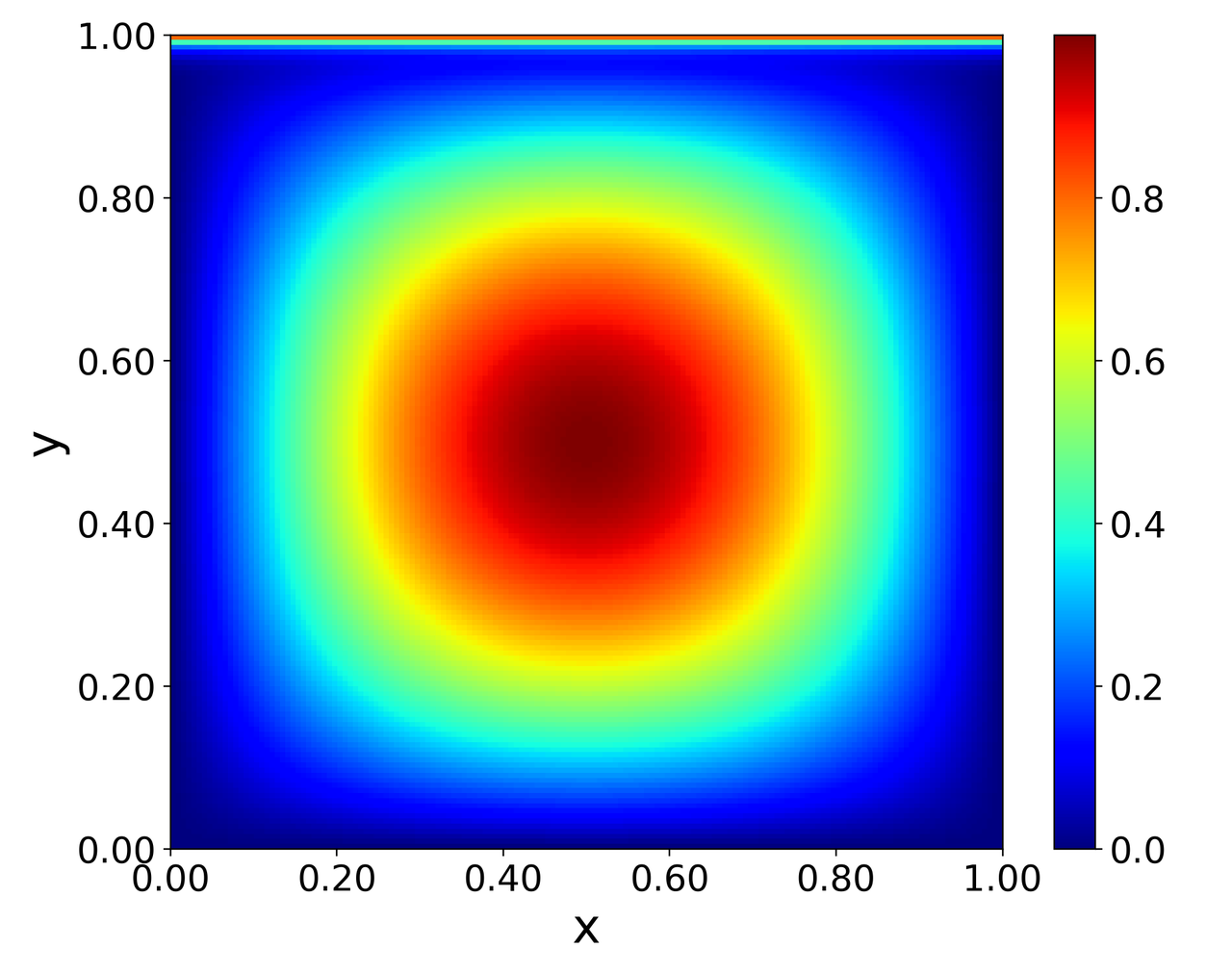}
    \caption{2D convection-diffusion equation}
\end{subfigure}
    \caption{Ground-truth solution for the 1/2D convection-diffusion equation.}
    \label{fig:v6}
\end{figure}

In this section, we present additional visualization results. These include the predicted solutions and error distributions of different models for the 1/2D convection–diffusion equations, as shown in Fig.~\cref{fig:v1,fig:v2,fig:v3,fig:v4,fig:v5}, together with the corresponding ground-truth solutions for reference in Fig.~\ref{fig:v6}. The results show that PINN and several enhanced baselines, including FF-PINN, gPINN, and SA-PINN, achieve reasonable accuracy in smooth regions, while some models exhibit pronounced error concentration near boundary layers. In contrast, GC-PINN produces predictions that closely match the ground-truth solutions, and its error distribution remains uniformly low over the entire space–time domain, with no evident error peaks near boundary layers. These observations provide further evidence of the effectiveness of the proposed method.

\subsection{Analysis of Other Mappings}
In this section, we additionally examine several alternative mapping strategies as representative failure cases for analysis. Specifically, we introduce piecewise linear (PWL) mappings and saturating mappings as comparative baselines with inferior performance. The corresponding designs and analyses are presented as follows.

\textbf{PWL Mapping.} This approach constructs a learnable monotonically increasing function from $[0,1]$ to $[0,1]$. Specifically, the input domain $[0,1]$ is uniformly divided into $K=16$ subintervals, where the $i$th interval is given by $[\frac{i}{K}, \frac{i+1}{K})$ for $i = 0, 1, \ldots, K-1$. On each subinterval, the mapping function $\phi(x)$ is defined as a linear segment with a learnable \emph{output increment} $s_i > 0$ (the actual slope with respect to $x$ is therefore $K s_i$).

To ensure that $\phi(0)=0$ and $\phi(1)=1$, we impose the integral constraint $\int_0^1 \phi'(x) \,dx = 1$. Since $\phi'(x) = K \cdot s_i$ on the $i$th subinterval (noting that $dx$ corresponds to an interval length of $\frac{1}{K}$), the total integral evaluates to $\frac{1}{K} \sum_{i=0}^{K-1} K s_i = \sum_{i=0}^{K-1} s_i$. Consequently, the constraint becomes $\sum_{i=0}^{K-1} s_i = 1$. We define the cumulative sum $S_i = \sum_{j=0}^{i-1} s_j$ (with $S_0 = 0$). For any input $x \in [0,1]$, we compute the subinterval index $i = \min(\lfloor Kx \rfloor, K-1)$ and the local coordinate $t = Kx - i \in [0,1)$. The mapping is obtained via linear interpolation:
\begin{equation}
    \phi(x) = S_i + s_i \cdot t.
\end{equation}
In practice, the parameters $\{s_i\}$ are typically generated via a softmax function to automatically satisfy the positivity and sum-to-one constraints.

\textbf{Saturating Mapping.} In contrast to the above, the saturating mapping serves as a deliberately impaired baseline in our comparative study to validate the necessity of well-designed mappings. The mathematical form employs a sigmoid function:
\begin{equation}
\phi(x) = \sigma(k(x - c)) = \frac{1}{1 + e^{-k(x-c)}},
\end{equation}
where the hyperparameters $k=50$ and $c=0.5$ control the steepness and the center position, respectively. While this formulation represents a smooth monotonic function from $[0,1]$ into $(0,1)$, the large value of $k$ induces extreme nonlinear characteristics. The derivative is computed via the chain rule:
\begin{equation}
\frac{d\phi}{dx} = k \cdot \sigma(k(x-c)) \cdot (1 - \sigma(k(x-c))).
\end{equation}

This expression yields a bell-shaped curve with respect to $(x-c)$, attaining its maximum at $x=c$ where $\sigma(0) = 0.5$ and the derivative reaches $k/4 = 12.5$. However, the derivative rapidly decays as $|x - c|$ deviates slightly from zero. We employ this mapping as a negative example to illustrate the pitfalls of poorly designed mappings: this construction causes the derivative to approach zero in most regions, leading to vanishing gradients during backpropagation through the chain rule.

\begin{table*}[h]
\centering
\caption{Performance comparison of different mapping methods across multiple benchmark problems.}
\label{tab:all_results}
\begin{tabular}{llccc}
\toprule
\textbf{Problem} & \textbf{Mapping Method} & \textbf{MSE} & \textbf{$\text{Rel}_{L^2}$} & \textbf{ $\text{Rel}_{H^1}$} \\
\midrule
\multirow{5}{*}{1D Burgers} 
                        & Torus Mapping           & 3.25E-10     & 2.44E-05         & 3.97E-05 \\
                        & Radial Mapping          & 8.69E-11     & 1.26E-05         & 1.18E-04 \\
                        & Local Stretching        & 1.69E-10     & 1.76E-05         & 9.15E-05 \\
                        & Saturating Mapping      & 2.75E-01     & 7.11E-01         & 8.33E-01 \\
                        & PWL Mapping             & 2.56E-06     & 2.17E-03         & 1.57E-03 \\
\midrule
\multirow{5}{*}{\parbox{3cm}{1D Convection-\\Diffusion}} 
                        & Torus Mapping           & 3.99E-06     & 2.00E-03         & 2.16E-03 \\
                        & Radial Mapping          & 6.00E-06     & 2.45E-03         & 3.55E-03 \\
                        & Local Stretching        & 9.32E-08     & 3.06E-04         & 6.66E-04 \\
                        & Saturating Mapping      & 3.98E-01     & 6.31E-01         & 1.00E-01 \\
                        & PWL Mapping             & 3.92E-05     & 6.27E-03         & 5.85E-03 \\
\midrule
\multirow{5}{*}{1D Helmholtz} 
                        & Torus Mapping           & 1.52E-08     & 1.74E-04         & 5.79E-05 \\
                        & Radial Mapping          & 1.42E-05     & 5.33E-03         & 1.87E-03 \\
                        & Local Stretching        & 1.42E-05     & 5.33E-03         & 1.77E-03 \\
                        & Saturating Mapping      & 2.57E+00     & 2.27E+00         & 8.62E-01 \\
                        & PWL Mapping             & 1.00E-04     & 1.42E-02         & 3.92E-03 \\
\midrule
\multirow{5}{*}{\parbox{3cm}{2D Convection-\\Diffusion}} 
                        & Torus Mapping           & 7.19E-08     & 5.33E-04         & 1.09E-03 \\
                        & Radial Mapping          & 3.88E-07     & 1.24E-03         & 1.76E-03 \\
                        & Local Stretching        & 3.26E-08     & 3.59E-04         & 5.17E-04 \\
                        & Saturating Mapping      & 9.04E-02     & 5.98E-01         & 1.03E+00 \\
                        & PWL Mapping             & 7.07E-05     & 1.67E-02         & 3.51E-02 \\
\bottomrule
\end{tabular}
\end{table*}

Table~\ref{tab:all_results} reports a quantitative comparison of five geometric mapping strategies across all benchmark problems. The three mappings based on geometric compactification consistently achieve relative errors below $10^{-4}$ in all test cases, whereas the Saturating Mapping and PWL Mapping exhibit substantially inferior performance. As a static ill-conditioned baseline, the Saturating Mapping yields errors three to seven orders of magnitude higher than the GC-based methods, reaching an MSE of $2.57$ on the 1D Helmholtz problem. This result indicates that enforcing the differentiability alone lead to a severe vanishing of the gradient. Although the PWL Mapping introduces learnable parameters via piecewise linear interpolation, its accuracy remains two to three orders of magnitude lower than that of the GC-based mappings. These observations suggest that learnability alone is insufficient to ensure reliable convergence; instead, effective performance requires mappings that explicitly encode problem-specific geometric structure.

\subsection{Model Complexity Analysis}
Table \ref{tab:param_counts} reports the parameter counts of all compared PINNs. The results show that GC-PINN introduces only a negligible increase in model complexity relative to the PINN, with parameter counts remaining within the same order of magnitude across all configurations. The Torus, Radial, and Local Stretching mappings introduce a small number of additional parameters, resulting in parameter counts that remain within the same order of magnitude as other PINN-based methods. In comparison, FF-PINN has a higher parameter count due to the inclusion of Fourier feature embeddings. Overall, the parameter statistics indicate that GC-PINN does not significantly alter model complexity in terms of parameter count.

\begin{table}[h]
\centering
\caption{Parameter counts of compared PINNs.}
\label{tab:param_counts}
\begin{tabular}{lr}
\toprule
\textbf{Model} & \textbf{Parameters} \\
\midrule
GC-PINN(Torus Mapping)             & 19,841 \\
GC-PINN(Radial Mapping)      & 19,682 \\
GC-PINN(Local Stretching)       & 19,684 \\
PINN                       & 19,681 \\
FF-PINN                    & 20,641 \\
PINN+RAR                   & 19,681 \\
SA-PINN                    & 19,683 \\
gPINN                      & 19,681 \\
\bottomrule
\end{tabular}
\end{table}
\section{Limitations}  
The proposed geometric compactification mapping paradigm and GC-PINN exhibit clear advantages in addressing multiscale partial differential equations. Nonetheless, several limitations remain. Only three types of mappings have been identified and studied, leaving open the possibility of additional mappings that could further extend the paradigm. Moreover, the successful application of these mappings requires substantial mathematical understanding, and the introduction of new mappings may entail additional hyperparameters, potentially increasing the burden for practitioners. Finally, due to the lack of adaptive mapping selection, practitioners must rely on prior analytical insight to choose an appropriate mapping for a specific PDE, or otherwise resort to simultaneously deploying multiple mappings at the expense of increased computational and time overhead.

\newpage

\end{document}